\documentclass[review]{elsarticle}

\usepackage{algorithm}  
\usepackage{algpseudocode}  
\usepackage{amsmath}  
\usepackage{xcolor}
\usepackage{graphicx}
\usepackage{subfigure}
\usepackage{underscore}
\usepackage{booktabs}
\usepackage{array}
\usepackage{multirow}
\makeatletter
\newcommand{\thickhline}{%
	\noalign {\ifnum 0=`}\fi \hrule height 1pt
	\futurelet \reserved@a \@xhline
}
\journal{arxiv}

\begin{document}

\begin{frontmatter}

\title{Nondiscriminatory Treatment: a straightforward framework for multi-human parsing}

\author[mymainaddress]{Min Yan}
\author[mymainaddress]{Guoshan Zhang\corref{mycorrespondingauthor}}
\cortext[mycorrespondingauthor]{Corresponding author}
\ead{zhanggs@tju.edu.cn}

\author[mymainaddress]{Tong Zhang}
\author[mymainaddress]{Yuemin Zhang}

\address[mymainaddress]{School of Electrical and Information Engineering, Tianjin University, 300072 Tianjin, China}

\begin{abstract}
Multi-human parsing aims to segment every body part of every human instance. Nearly all state-of-the-art methods follow the ``detection first'' or ``segmentation first'' pipelines.
Different from them, we present an end-to-end and box-free pipeline from a new and more human-intuitive perspective. In training time, we directly do instance segmentation on humans and parts. More specifically, we introduce a notion of ``indiscriminate objects with categories'' which treats humans and parts without distinction and regards them both as instances with categories. In the mask prediction, each binary mask is obtained by a combination of prototypes shared among all human and part categories. In inference time, we design a brand-new grouping post-processing method that relates each part instance with one single human instance and groups them together to obtain the final human-level parsing result. We name our method as Nondiscriminatory Treatment between Humans and Parts for Human Parsing (NTHP). Experiments show that our network performs superiorly against state-of-the-art methods by a large margin on the MHP v2.0 and PASCAL-Person-Part datasets.  
\end{abstract}


\end{frontmatter}

\section{Introduction}

Human parsing has become a significant aspect of human-centric analysis in recent years, which requires fine-grained semantic segmentation on pixel level. Extensive studies have been explored on parsing a single human in an image and obtained remarkable progress \cite{Yamaguchi2012}\cite{Dong2013}\cite{Liang2017}. But in real cases, various numbers of persons are present simultaneously with interaction and occlusion, which heighten the need for better instance-level multi-human parsing methods. Multi-human parsing has many real-world applications, such as virtual reality \cite{Lin2016}, video surveillance \cite{Liu2017}, and human behavior analysis \cite{Fan2019}\cite{Zhou2019}\cite{Wang2020b}. In this work, we aim at solving the task of multi-human parsing.

\begin{figure}[t]
	\centering
	\subfigure[]
	{
		\begin{minipage}[b]{0.22\linewidth}
			\centering
			\includegraphics[height=3cm]{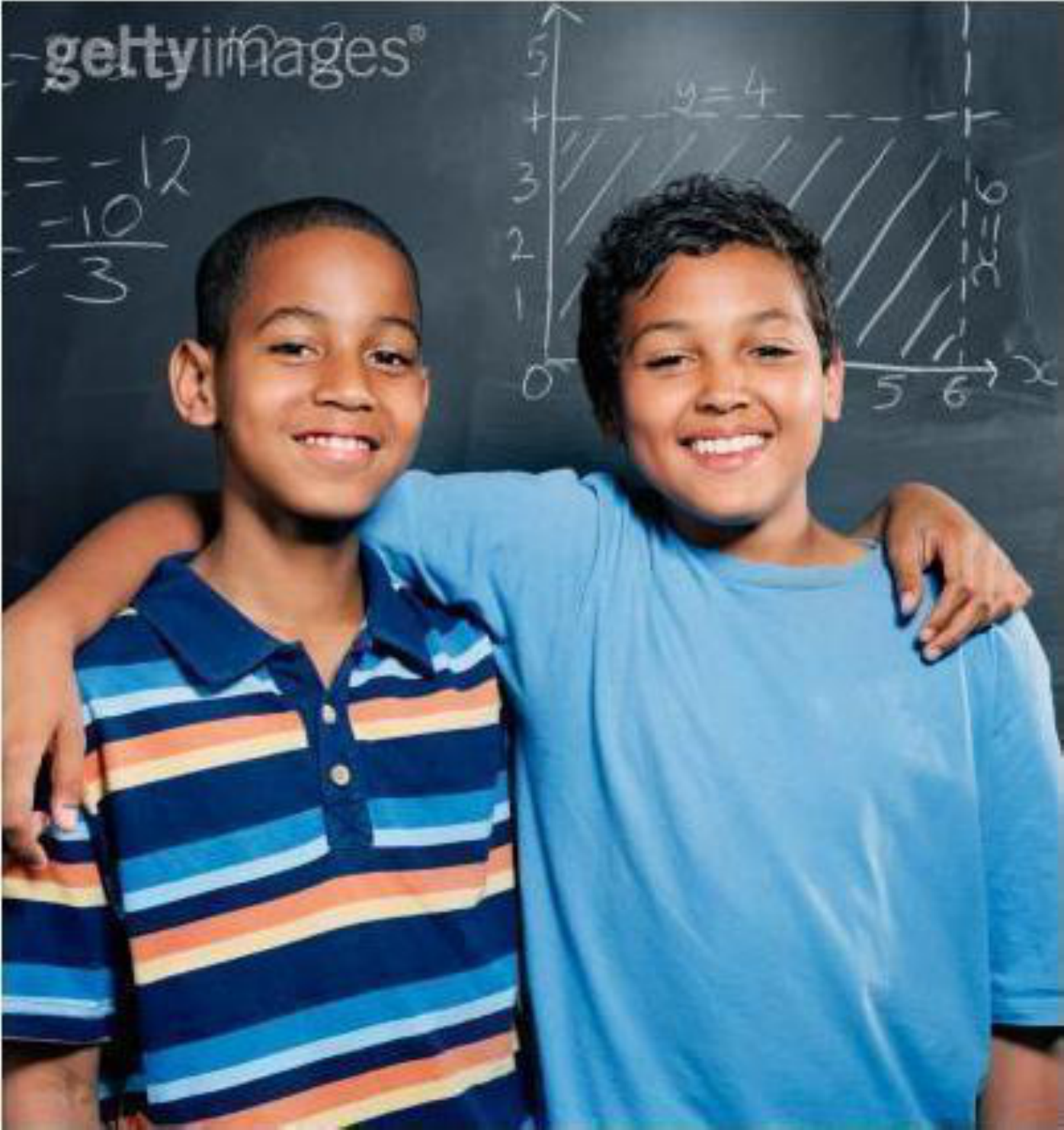}
		\end{minipage}
	}
	\subfigure[]
	{
		\begin{minipage}[b]{0.22\linewidth}
			\centering
			\includegraphics[height=3cm]{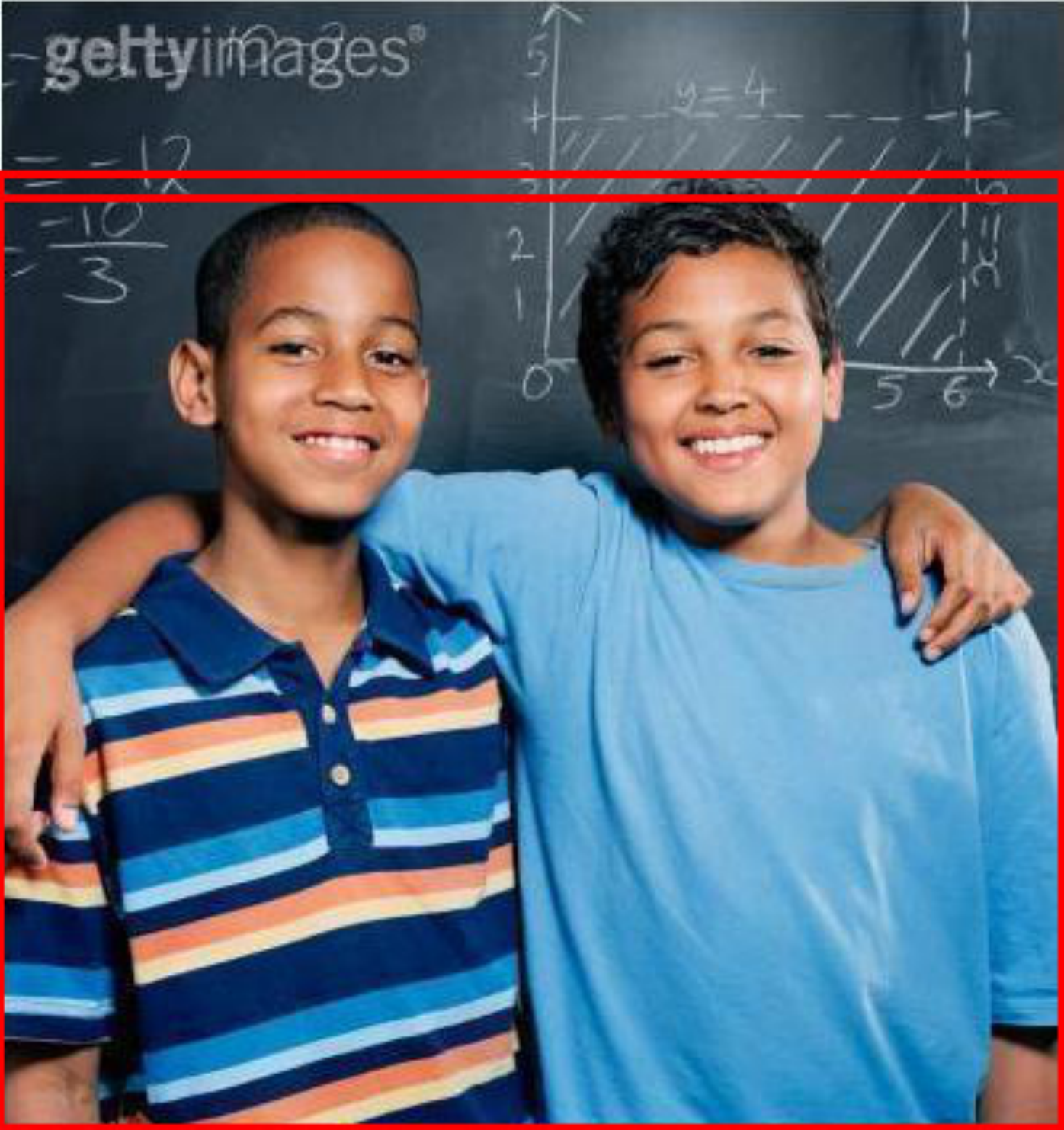}
		\end{minipage}
	}
	\subfigure[]
	{
		\begin{minipage}[b]{.22\linewidth}
			\centering
			\includegraphics[height=3cm]{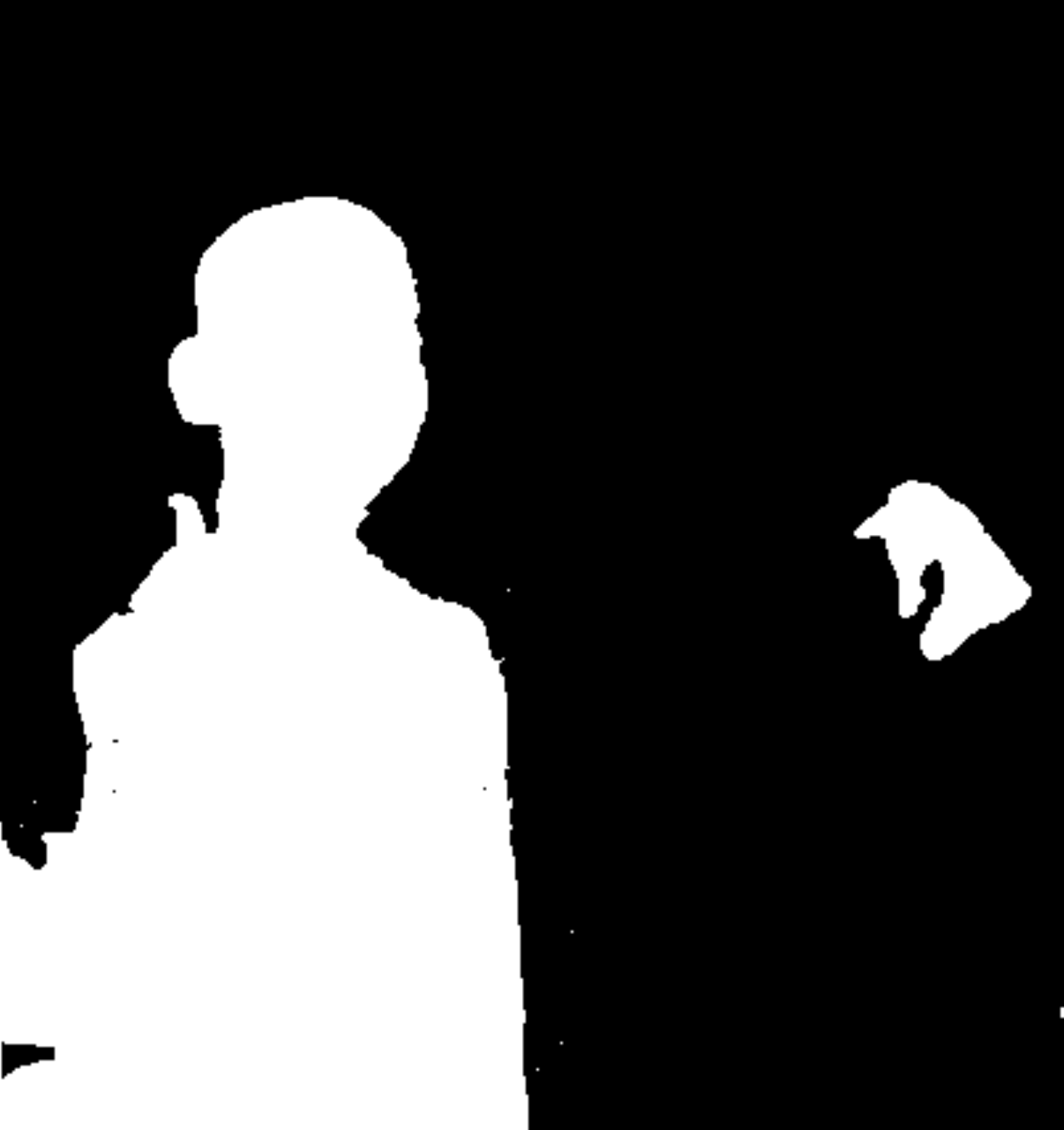}
		\end{minipage}
	}
	\subfigure[]
	{
		\begin{minipage}[b]{.22\linewidth}
			\centering
			\includegraphics[height=3cm]{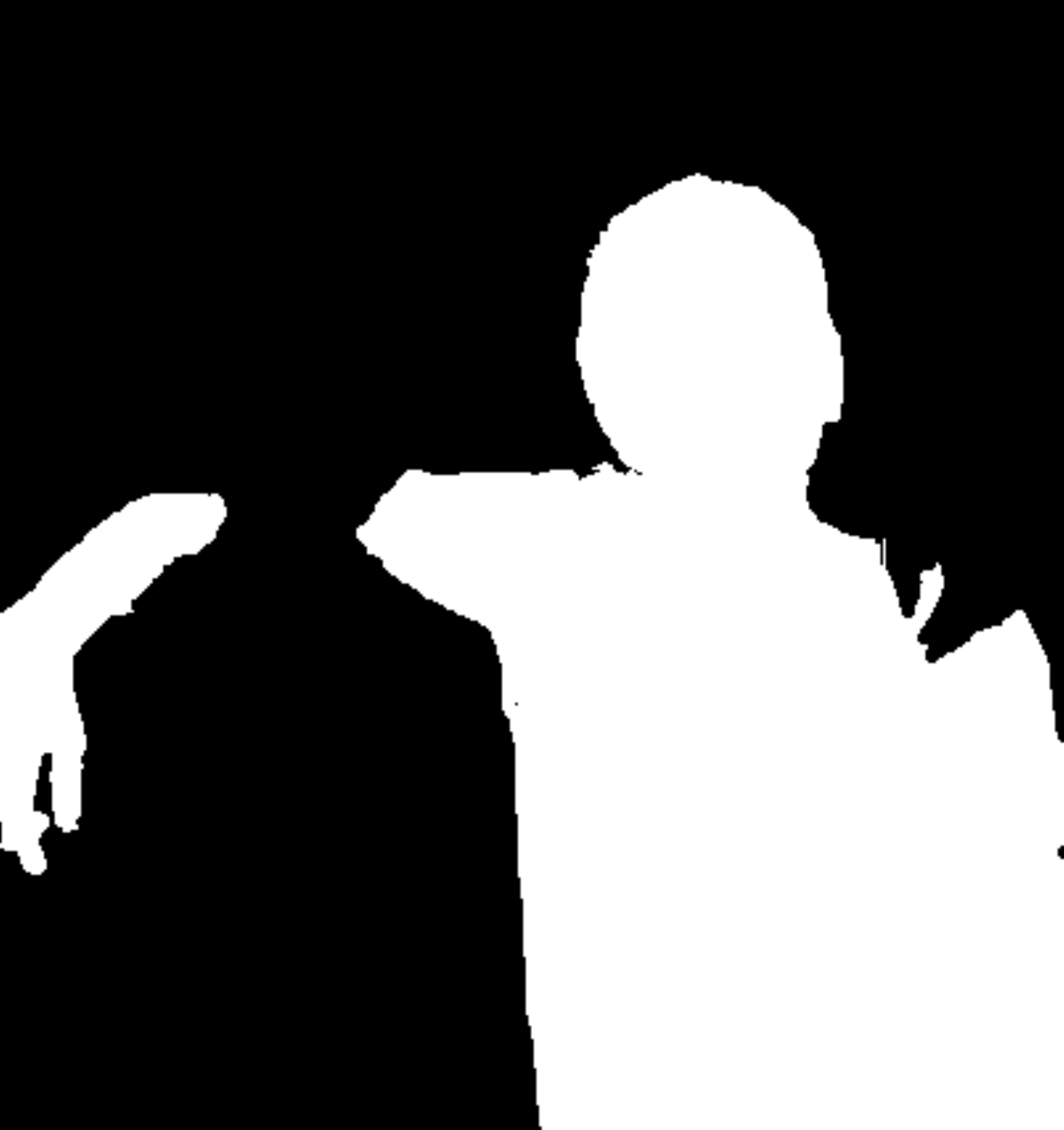}
		\end{minipage}
	}
	\caption{An example. (a) is the original picture, (b) is the bounding-box ground truth and (c)(d) are human-level mask labels. In (b), the Intersection over Union (IOU) between the two bounding boxes is 97.5$\%$ through calculation. }
	\label{figure1}
\end{figure}

Due to the successful development of fully convolutional neural networks \cite{Girshick2014}\cite{Krizhevsky2012} \cite{Shelhamer2017}\cite{Szegedy2016}\cite{Wu2018}, multi-human parsing has achieved great progress\cite{Qin2020}\cite{Li2017}\cite{Yang2019a}\cite{Gong2018}. Existing works dealing with multi-human parsing can be divided into two categories, “detection first” and “segmentation first” paradigms. 

“Detection first” methods consist of two stages. They detect human instances in the first stage then utilize regions-of-interest (ROIs) to parse the detected human instances in the second stage \cite{Qin2020}\cite{Yang2019a}\cite{Yang2020a}. However, these methods may have the following drawbacks. 1)Human instances are often with irregular shapes, while ROIs are axis-aligned bounding-boxes, so the cropped features used to parse a single human can be excessive. We show an example in Figure \ref{figure1} (b). 2)These methods strongly rely on the quality of the bounding boxes predicted. A Little deviation can result in huge faults. 3)Human parsing requires more detailed information in the second stage so that cropping the region to a 14×14 resolution of the conventional ROI align operation is not enough. 4)In these methods, the second stage must wait for the first stage to get accurate ROIs so that the processing procedure is slow. 

On the other hand, “segmentation first” methods also have two stages. They apply a fine-grained semantic segmentation to obtain a pixel-level classification in the first stage then group pixels that belong to the same human instance in the second stage \cite{Gong2018}\cite{Li2017a}. In the first stage, different parts demand different receptive fields due to their various sizes. Many approaches are committed to solving this problem, such as ASPP \cite{Chen2018} , PSP \cite{Zhao2017a}, but lead to great computational complexity. In the second stage, some works separate different human instances via edges \cite{Gong2018}. There is more than one boundary if a human is blocked. Figure \ref{figure1} (b)(c) shows examples, both persons are in two parts, and previous approaches tend to group the apart hand by mistake.

\begin{figure}[t]
	\centering
	\subfigure
	{
		\begin{minipage}[b]{.3\linewidth}
			\centering
			\includegraphics[height=4cm]{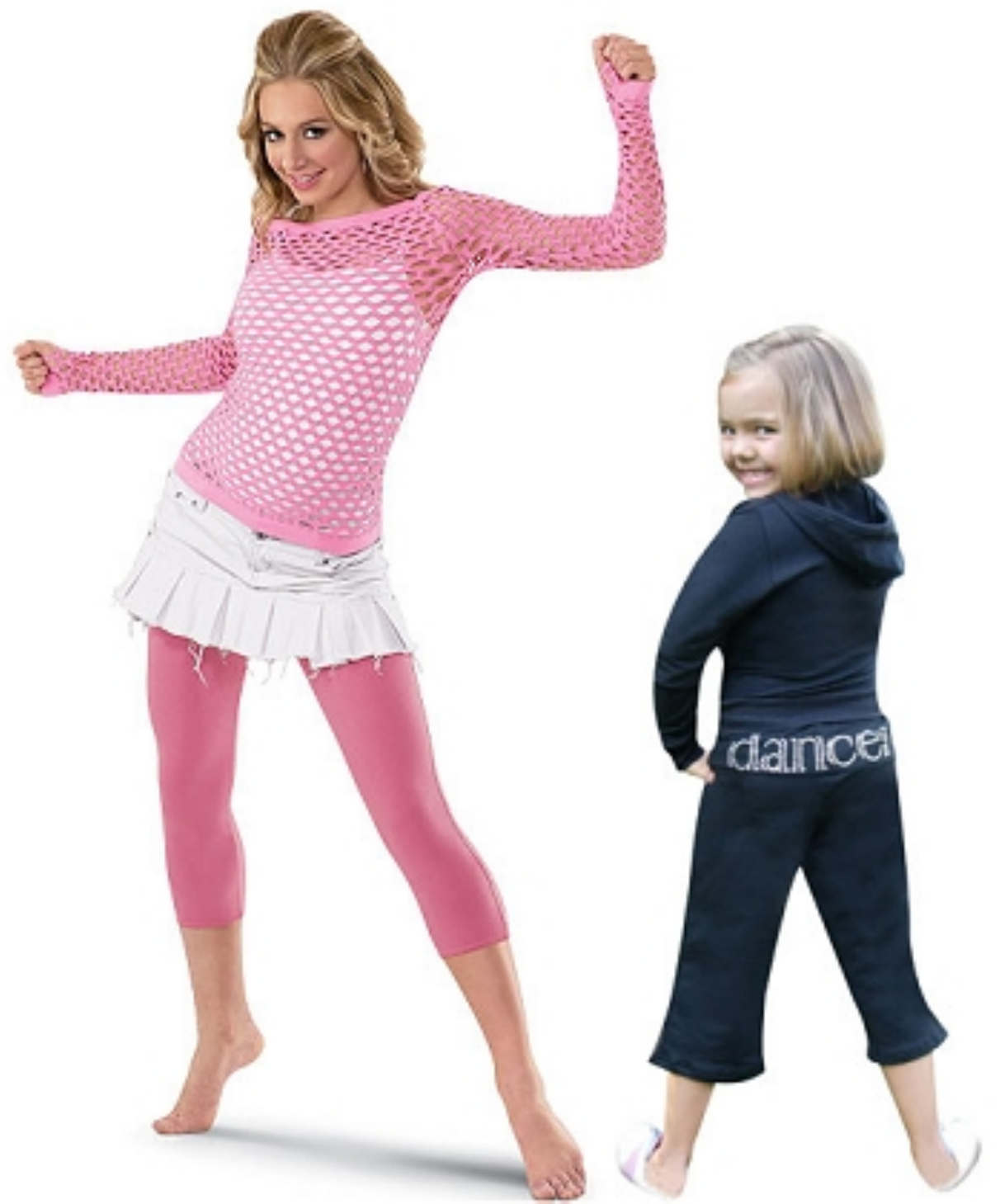}
		\end{minipage}
	}
	\subfigure
	{
		\begin{minipage}[b]{.3\linewidth}
			\centering
			\includegraphics[height=4cm]{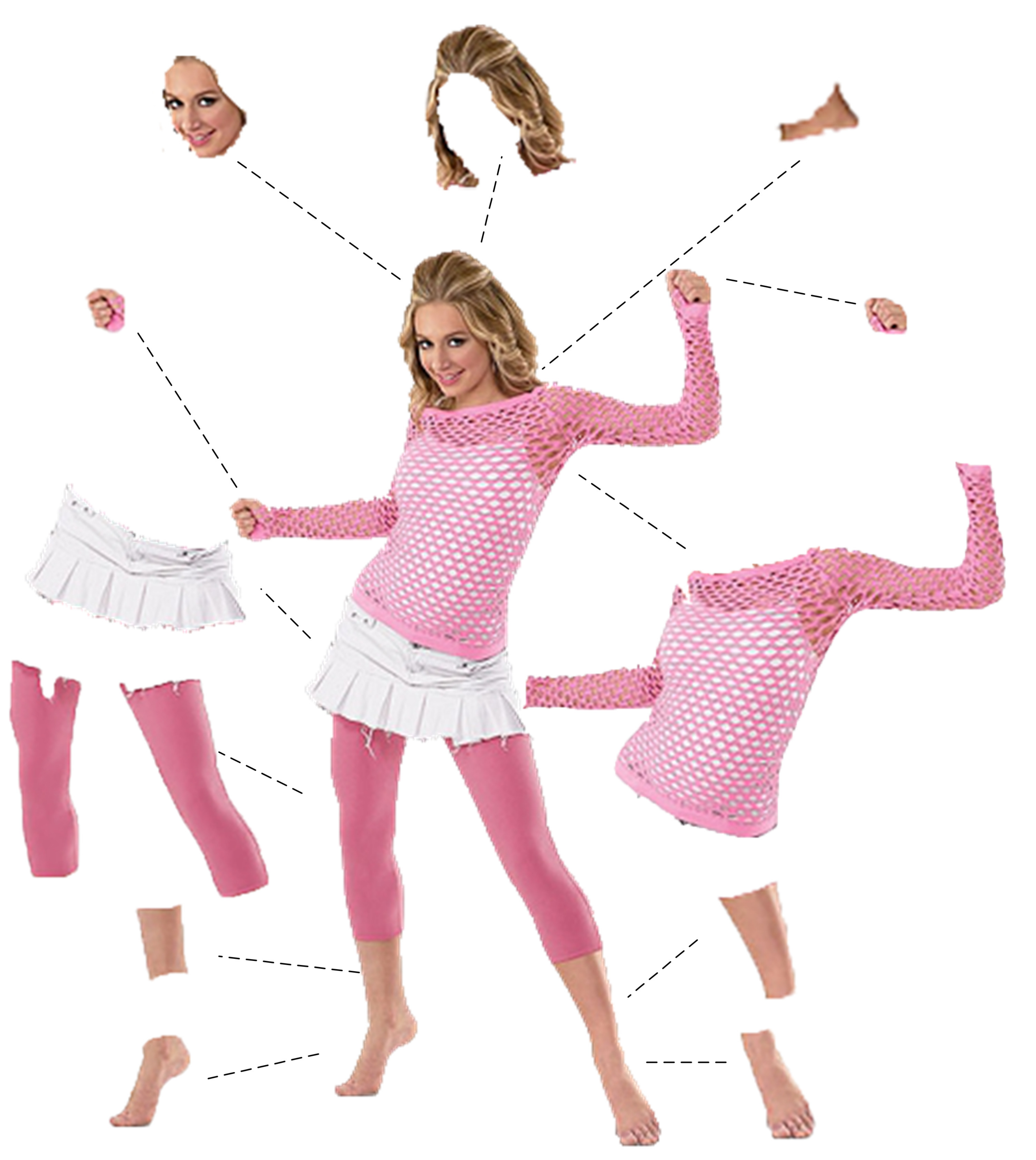}
		\end{minipage}
	}
	\subfigure
	{
		\begin{minipage}[b]{.3\linewidth}
			\centering
			\includegraphics[height=4cm]{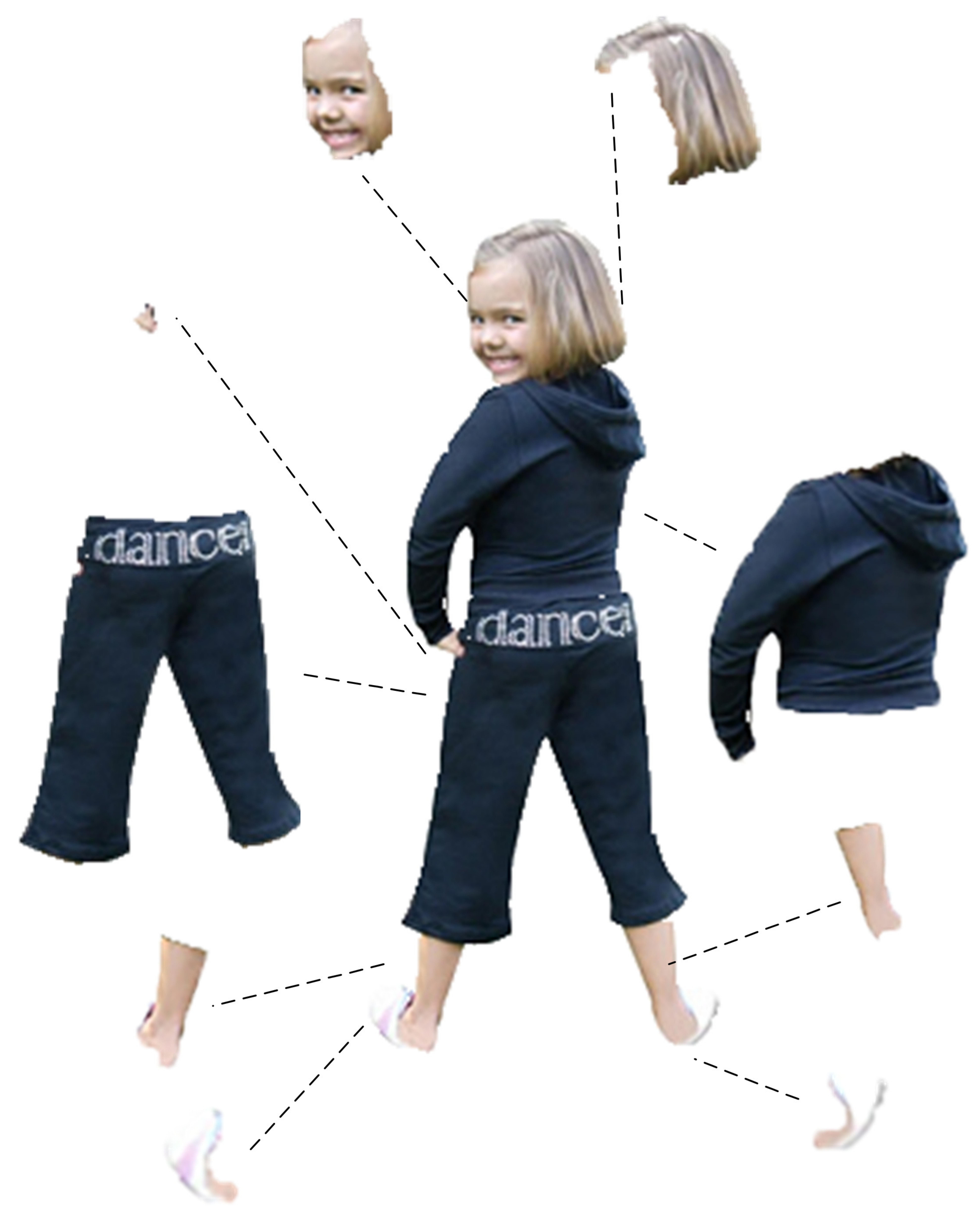}
		\end{minipage}
	}
	\caption{Solution formulation. According to human intuition, a human in the picture is composed of several parts with categories rather than dense pixels with semantic labels.}
	\label{figure2}
\end{figure}

Our research explores multi-human parsing from a brand-new point of view. We imitate the thinking process of human beings as shown in Figure \ref{figure2}, which views a human as a collection of parts and regards each part or human as an instance with category rather than pixels with categories. We simultaneously execute instance segmentation from two aspects, part and human, with nondiscriminatory treatment and predict their class-agnostic masks and instance categories. We named our notion “indiscriminate objects with categories”. To better implement our notion, we propose a unified mask prediction module named Unified Mask Prediction Based on Prototypes (UMPP) which uses a unified prototype generation for both aspects. Finally, we design a simple grouping strategy that combines the separate parts belonging to the same human. Note that in our method, two aspects (human and part) are not completely unrelated. Both features are extracted from the same FPN structure but different levels, and share the prototypes, which can benefit from each other.

To evaluate our proposed framework, we conduct extensive experiments on the MHP v2.0 \cite{Zhao2018} and PASCAL-Person-Part \cite{Chen2014} datasets. We achieve state-of-the-art performance with 51.1 $AP^p_{50}$, 49.5 $AP^p_{vol}$, and 49.9 $PCP_{50}$, with a margin of 5.8 points $AP^p_{50}$, 2.7 points $AP^p_{vol}$, and 6.1 points $PCP_{50}$ over the best previous entry on the MHP v2.0 dataset \cite{Zhao2018}. As for the PASCAL-Person-Part dataset, we also achieve state-of-the-art performance with 47.1 $AP^r_{vol}$ and 53.9, 44.7, 31.9 $AP^r$ with IoU thresholds of 0.5, 0.6, 0.7, separately with a margin of 4, 5.8, 6.4, 6.2 points over the best previous entry.

The main contributions of our work are concluded as follows:

$\bullet$ We design an end-to-end and box-free framework named NTHP for multi-human parsing keeping in line with our new notion of “indiscriminate objects with categories”, which views both the humans and parts as object instances with categories rather than pixels with semantic labels.

$\bullet$ We propose a unified mask prediction module named Unified Mask Prediction Based on Prototypes (UMPP) formed by a linear combination of prototypes shared among humans and parts.

$\bullet$ We design a new grouping strategy in inference.

$\bullet$ We outperform all state-of-the-art methods on the MHP v2.0 \cite{Zhao2018} and PASCAL-Person-Part \cite{Chen2014} datasets.

\section{Related work}

\subsection{Human parsing} 

Human parsing has received a lot of attention in recent years. The hardest thing lies in obtaining the structure information within the human body. PCNet \cite{Zhang2020a} designs a relational aggregation module and a dispersion module to deliver human structure information between different parts. Hierarchical Human Parsing \cite{Wang2020a} utilizes graph convolutional networks to understand hierarchical human layouts better. CorrPM \cite{Zhang2020} puts forward a heterogeneous non-local block to fully take advantage of the correlation between parsing, pose, and edge. All these works are within the scope of the single person parsing which can be viewed as a dense per-pixel classification problem. 

Human parsing is elevated to a new level which has an unfixed number of persons in an image with the representation of the MHP dataset\cite{Li2017}. Nearly all methods follow the ``detection first'' or ``segmentation first'' pipeline in instance segmentation. PGN \cite{Gong2018} adopts the ``segmentation first'' pipeline which appends a human-level instance-aware edge detection branch parallel with semantic segmentation and connects segments via the predicted boundary. Parsing R-CNN \cite{Yang2019a} and Unified Framework \cite{Qin2020} adopt the ``detection first'' pipeline to parse distinct parts within the predicted human instances. CE2P \cite{Ruan2018} further appends a global parsing branch in parallel with the ``detection first'' pipeline to improve the performance. In contrast, we deal with humans and parts at the same time using the same structure.

\subsection{Instance segmentation} 

Multi-human parsing can be viewed as a more complicated instance segmentation to some extent. Most of the methods tackling multi-human parsing are evolved from instance segmentation. Instance segmentation is one of the most common tasks in computer vision.  
The methods mainly follow the ``top-down'' or ``bottom-up'' pipeline. ``Top-down'' methods follow the principle of Mask R-CNN \cite{He2017}. They first employ a detector to extract human-level features within a bounding-box. Then for each human instance, regions-of-interest (ROIs) are cropped from the original picture-level feature maps. Finally, these ROIs are used to obtain the detailed segmentation results. Follow-up works are dedicated to improving the accuracy. PANet \cite{Liu2018} adds another ``bottom-up'' path in FPN \cite{Lin2017a} to reinforce the feature representation and uses the “adaptive feature pooling” strategy to fuse the features from different levels to get better representation. Mask Scoring R-CNN \cite{Huang2019} realizes the misalignment between the masks and the classification scores so that a MaskIoU head is appended to predict the quality of the masks predicted. Other approaches adopt the ``bottom-up'' strategy. For instance, Panoptic Deeplab \cite{Cheng2020} aims at panoptic segmentation but achieves good performance in instance segmentation too.
 
Recently, some methods that follow another pipeline named ``one-stage'' have obtained more interest due to their simplicity and easy-understanding nature. YOLACT \cite{Bolya2019} makes use of prototypes to generate instance masks and coefficients to get a linear combination of all predicted masks, after which a cropping operation is used to localize the objects. PolarMask \cite{Xie2020}handles the task from a new perspective, which directly predicts the contours of instances in the polar coordinate. SOLO \cite{Wang2019} directly predicts binary masks and mask categories by building two branches following the same backbone and FPN \cite{Lin2017a}, one is the category branch, and the other is the mask branch. SOLOv2 \cite{Wang2020} further splits the mask branch into mask feature and convolutional kernel paths to implement dynamic convolution. One-stage methods have difficulty localizing objects since it’s commonly believed that convolutional operations are translation-invariant. YOLACT \cite{Bolya2019} obtains translation-variance by cropping the final mask with the predicted bounding box and SOLO \cite{Wang2019}\cite{Wang2020} utilizes  CoordConv \cite{RosanneLiuJoelLehmanPieroMolino2018} to get translation-variance. In this work, we introduce one-stage method to multi-human parsing and tackle the translation-invariant problem with CoordConv.

\subsection{Prototypes} 
Learning prototypes (aka vocabulary or codebook) has been extensively explored in object detection\cite{Ren2013}\cite{Agarwal2002}. But the prototypes in these works are used to represent features. YOLACT \cite{Bolya2019} learns prototypes specific to each image rather than global prototypes shared across the entire dataset. 
We obtain mask predictions using a linear combination of prototypes shared among humans and parts in this work.

\begin{figure}[t]
	\centering
	\includegraphics[width=12cm]{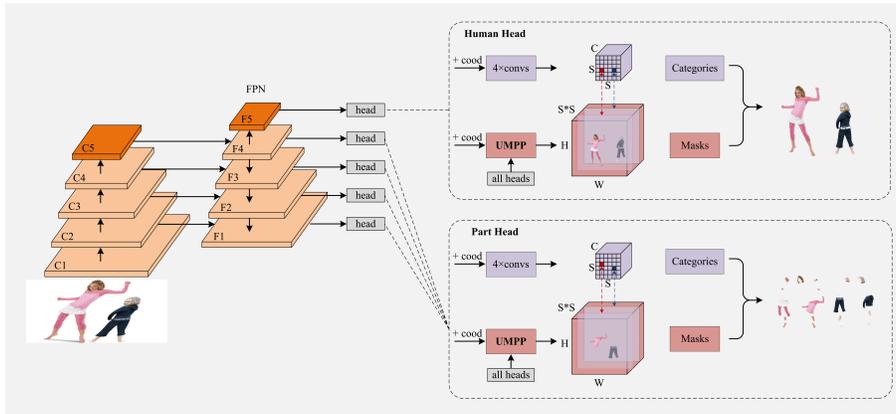}
	\caption{NTHP structure. The topmost level of FPN is assigned to humans, and the rest are to parts. UMPP represents our proposed module (Unified Mask Prediction Based on Prototypes). UMPP has two types of inputs: one is the corresponding FPN output, and the other is the combination of all FPN outputs.}
	\label{figure3}
\end{figure}

\section{Proposed method}

 In this section, we propose a straightforward approach extending one-stage methods original designed for instance segmentation to multi-human parsing \cite{Bolya2019}\cite{Wang2019}\cite{Wang2020}. We will introduce problem formulation, NTHP architecture, learning details, and inference procedure in detail.

\subsection{Solution formulation}

We consider the task from a human visual perspective. As shown in Figure \ref{figure2}, when a person looks at an image and does the same parsing job, he will first notice a human instance, then consider which parts belong to that human. There are three discoveries, 1) people will see an object as an instance with a category rather than a collection of pixels with categories, 2) parts and humans are both objects with little distinction from human perspective, 3) human instances are made up of different part instances. Based on the first two discoveries, we introduce the notion of “indiscriminate objects with categories”, which views humans and parts as indiscriminate objects and completes instance predictions with one structure. Based on the last discovery, we introduce our grouping strategy in inference time.

\subsection{Overview}

In this work, we directly predict the masks and their corresponding categories for both humans and parts with the same structure based on our notion of “indiscriminate objects with categories”.

We show our structure in Figure \ref{figure3}. Similar to \cite{Wang2019}\cite{Wang2020}, we divide the input images into several $S \times S$ grids aligned to different levels of the feature pyramid network (FPN) \cite{Lin2017a}. One of the grids ($\textit{i, j}$) is activated if it falls into the center region of any ground-truth mask. There are two branches following FPN, category branch ($C\times S \times S$) and mask branch ($S^2\times H \times W$), where $C$ equals the number of classes and $H$, $W$ respectively represent feature height and width. There is a one-to-one relationship between masks and categories. If a grid ($\textit{i, j}$) is activated, its category prediction is at ($\textit{i, j}$) of the category branch with $C$ channels, and the class-agnostic mask is at the $(\textit{i} \cdot \textit{S} + \textit{j})^{th}$ channel of the mask branch. We elaborate on the category branch in section 3.3. We name our structure in mask branch Unified Mask Prediction Based on Prototypes (UMPP) and tell more detail in 3.4.

There is a principle that, in most cases, two instances in an image either have different center locations or have different object sizes \cite{Wang2019}\cite{Redmon2016}. For the location issue, it is generally accepted that the original convolutions are translation-invariant to some degree. The solution is that we add Coordconv \cite{RosanneLiuJoelLehmanPieroMolino2018} in the mask branch, the same as SOLO \cite{Wang2019}\cite{Wang2020} and YOLACT \cite{Bolya2019}. And for the size issue, we assign objects of different sizes to different levels of FPN (five levels in total). Human instances always have bigger sizes than part instances, so we assign the topmost level to human instances and the rest to the parts. Detailed parameters for each level are shown in Table \ref{tab:table1}.

With the obtained instances with categories, we group part instances concerning human instances to form the human-level parsing result. We elaborate on the grouping process in section 3.6.

\subsection{Category branch}
There are five levels in total with output space $C\times S \times S$ in the category branch. $C$ equals 1 for the human category prediction branch and the number of classes excluded background for parts. There are $4$$\times$$convs$ ($3$$\times$$3$) for feature extraction and one for prediction in each level. We share weights for those levels assigned to parts. Note that the classes predicted here are for instances rather than pixels, thus worsening the difficulty.

One puzzling question is that without the structure based on the single human, the network may have difficulty differentiating confusing categories like left-right hands and left-right arms. But our network can successfully accomplish that. The reasons are: 1) Left and right have different directional properties, and the network can learn that information if the feature is fine enough. Thanks to the specific attribute of FPN, our network can satisfy the receptive fields required for objects of various sizes thus obtaining sufficient context information. 2)  We use only one FPN so that parts can also get the human-level structure information through information flow. 

\begin{figure}[t]
	\centering
	\includegraphics[width=12cm]{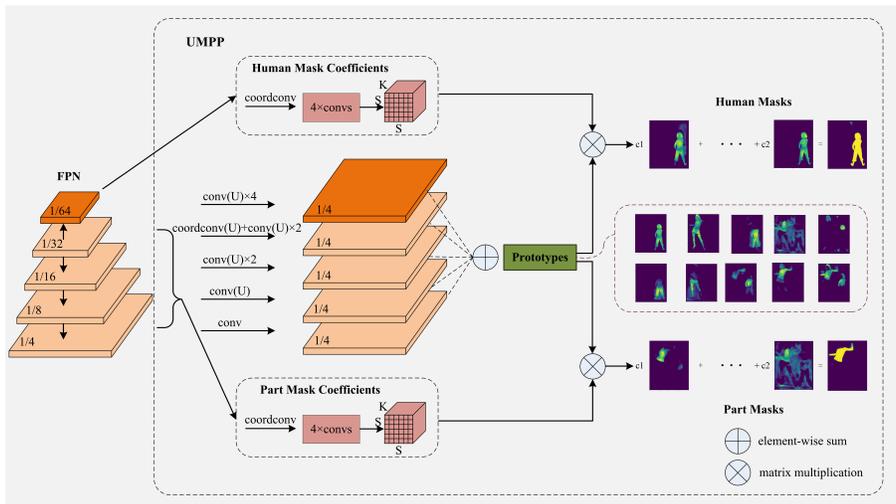}
	\caption{Unified Mask Prediction Based on Prototypes (UMPP) Blue/yellow indicates low/high values in the prototypes. We use all levels of FPN to form prototypes. ``U'' in brackets represents the upsampling operation}
	\label{figure4}
\end{figure}

\subsection{Unified Mask Prediction Based on Prototypes (UMPP)}

In UMPP, as shown in Figure \ref{figure4}, we learn mask coefficients separately for humans and parts and a unified collection of prototypes to form the class-agnostic masks.

Conventional convolutional operations are adept in taking advantage of the spatial coherence to obtain context information while $fc$ operations are good at producing semantic vectors. In our structure, the branches for mask coefficients are parallel with the category branches, also with $4$$\times$$convs$ ($3$$\times$$3$) for feature extraction and one $conv$ ($1$$\times$$1$) for prediction. The former $4$$\times$$convs$ can learn the context information for each grid due to the attribute of convolutions and the last $1$$\times$$1$ $conv$ is the re-implementation for $S$$\times$$S$ $fc$ operations to produce semantic vectors. For each grid, we predict $K$ coefficients in the channel direction. The output space of mask coefficients $F$ is $K$$\times$$S$$\times$$S$. Just like the category branch, the highest level is for humans. We use $K=256$ for both humans and parts in experiments.

We learn a unified collection of prototypes $P$ for humans and parts by applying feature pyramid fusion proposed in \cite{Kirillov2019} but adding Coordconv \cite{RosanneLiuJoelLehmanPieroMolino2018} to the highest part level. We use all levels of FPN. Each level consists of a series of convolutions, and all levels are upsampled to 1/4 scale of the input image and summed together. There is a $1$$\times$$1$ $conv$ with ReLU at the end. The output space is $K$$\times$$H$$\times$$W$. In the center-of-right of Figure \ref{figure4}, we show some of the visualization results of prototypes. The first two examples are prototypes of humans while the others are parts. We can see that our network can successfully learn the location information, thus obtaining translation-variance, i.e., two individual persons. Such phenomenon largely owes to Coordconv \cite{RosanneLiuJoelLehmanPieroMolino2018} or the padding operation according to YOLACT\cite{Bolya2019}, which demonstrates that padding gives the network the ability to tell how far away from the image’s edge a pixel is. We also consider that using a unified expression brings mutual-benefits. In FPN, higher-level features are convoluted using lower-level features and lower-level features combine higher-level features. That is to say, part features can get human-level information and learn more accurate context information, and in the meantime, human features can obtain the fusion of the part features to form the overall information. 

The final mask predictions are obtained by a linear combination of prototypes, which is implemented as a sigmoid of a single matrix multiplication:
\begin{equation}
	M=\sigma(PF^T)
\end{equation}	
where $P$ is the matrix of prototypes, $F$ is the matrix of mask coefficients, and $\sigma$ is the sigmoid operation. The output space of $M$ is $S^2 \times H \times W $.

\subsection{NTHP learning}
\subsubsection{Label assignment}
There are five levels of outputs with various resolutions that concentrate on objects of different sizes in the FPN of our structure. We assign the highest level to humans and the rest to parts. Besides, low levels have high resolutions and are responsible for small objects, thus requiring more grids. We show more details in Table \ref{tab:table1}. “Scale” means root mean square of the area of the minimum bounding box of the object. 
 
In the head after each FPN level, there are S×S grids. A grid $(i, j)$ is activated if it falls into the center region of any ground-truth mask, and 1) its category label is the class of the corresponding ground truth, 2) its mask label is the binary mask of the corresponding ground truth. Given a ground-truth mask, we calculate its mass center (\textit{$c_x$},\textit{$c_y$}), width \textit{w}, height \textit{h}, then the center region is controlled by constant scale factors
\textit{$\varepsilon$}:(\textit{$c_x$},\textit{$c_y$},\textit{$\varepsilon$w},\textit{$\varepsilon$h}), we set \textit{$\varepsilon$}=0.2 following SOLO \cite{Wang2019}. For each ground truth, there are no more than 9 grids activated. If the number of grids exceeds, the nine closest to the center point are used.

\begin{table}[h]
	\caption{Label assignment}
	\begin{center}
		\begin{tabular}{c c c c c c}
			\hline
			Pyramid & F1 & F2 & F3 & F4 & F5 \\\hline
			Object & Part & Part & Part & Part &Human \\ 
			Grids (S) & 40 & 36 & 24 & 16 & 20 \\ 
			Scale & $\textless$96& 48$\sim$192 & 96$\sim$384 & $\geq$192& -\\\hline
		\end{tabular}			
		\label{tab:table1}
	\end{center}
\end{table}

\subsubsection{Loss function}
We use the following training loss function:

\begin{equation}
	L=L_{cp}+\lambda L_{mp}+L_{ch}+\lambda L_{mh}
\end{equation}
where $L_{cp}$ is Focal loss \cite{Lin2017} for the category classification for parts, $L_{mp}$ is Dice Loss \cite{Milletari2016} for the mask prediction for parts, $L_{ch}$ is Focal loss \cite{Lin2017} for the category classification for humans and $L_{mh}$ is Dice Loss \cite{Milletari2016} for the mask prediction for humans. $\lambda$ is set to 3 in experiments. Note that we calculate classification loss for each grid but mask loss only for grids that have instance labels. 

\subsection{Grouping strategy in inference time}

Our structure obtains four types of information, categories of part instances, masks of part instances, categories of human instances, and masks of human instances. In inference time, we need to group the parts based on humans. We propose the following steps:

\begin{figure}[h]
	\centering
	\includegraphics[width=12cm]{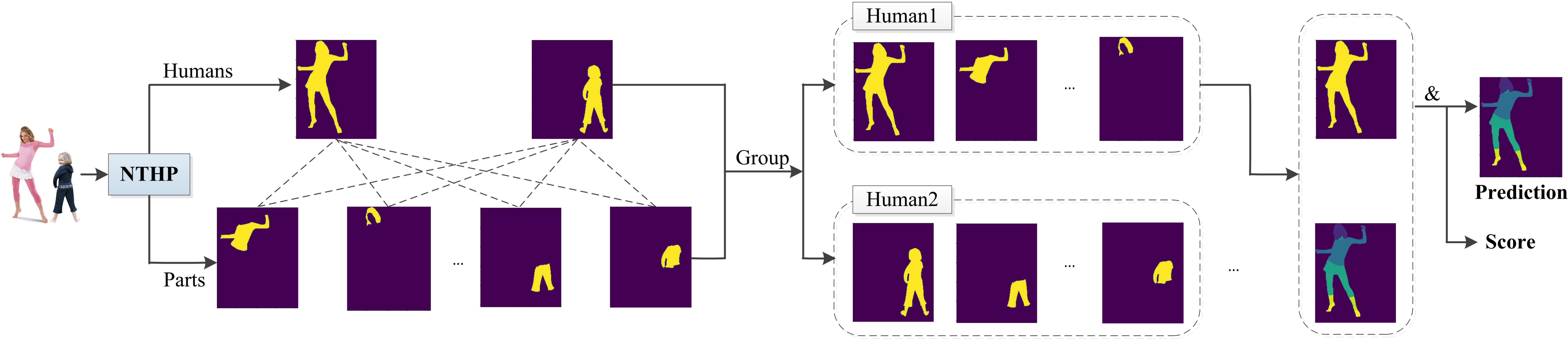}
	\caption{Grouping strategy in inference time. \& means the “and” operation.}
	\label{figure5}
\end{figure}

\begin{enumerate}[(1)]
	\item Assign the category with the highest classification score to the corresponding part mask. Pick $k_{part}$ part masks with the highest category scores. We predefine two thresholds, $n_{part}$ for the max number of instances in an image (to decrease the memory cost), $s_{part}$ for the minimum score value. $k_{part}$ can be calculated by:
	\begin{equation}
		k_{part}=\rm min(\textit{num}(\textit{$score_{part}$}> \textit{$s_{part}$}),\textit{$n_{part}$})
	\end{equation}	
	where \textit{num} means the number of masks that meets the criteria, and \textit{score} means the mask score, who is the product of the category score and the segmentation score. We use \textit{$n_{part}$}=200, \textit{$s_{part}$}=1/3. 
	\item Pick \textit{$k_{human}$} human masks whose scores are bigger than \textit{$s_{human}$} and apply matrix NMS \cite{Wang2020} to select the best ones. We use \textit{$s_{human}$}=0.1 in experiments.
	\item For each selected part instance, calculate its overlapping ratio $r_{part}$ towards every human instance:
	\begin{equation}
		r_{part}=\frac{area(intersection(part,human))}{area(part)}
	\end{equation}
	We show its pseudocode in Algorithm \ref{code:ratio}
	\item For each human instance, first pick the part instances whose overlapping ratio $r_{part}$ is bigger than $r_{human}$, then assign every pixel with the category label using the selected part instances sorted by scores, finally use the ‘and’ operation between the human and the combination of the selected part instance masks to get the final result.  $r_{human}$ =2/3 in experiments. We use the product of the class-agnostic human-level mask score and the mean of category scores of pixels within the human mask as our final score.
	\begin{equation}
		score_{parsing}=score_{human}*\rm mean(\textit{$score_{pixel}$})
	\end{equation}
	where $score_{parsing}$ is the final score for human parsing, $score_{human}$ is the human-level instance score, and ${score_{pixel}}$ is the category scores of pixels within the human mask.
\end{enumerate}

The process is illustrated in Figure \ref{figure5}

\begin{algorithm}[h]
	\caption{Pseudocode of computing overlapping ratio $r_{part}$} 
	\small
	\scriptsize 	 
	\begin{algorithmic}
		\State {\color{gray} \# mm: matrix multiplication}
		\State {\color{gray} \# num\_h: number of masks for humans}
		\State {\color{gray}\# num\_p: number of masks for parts}
		\State {\color{gray} \# h\_masks: binary masks for humans (num\_h$\times$h$\times$w)}
		\State {\color{gray}\# p\_masks: binary masks for parts (num\_p$\times$h$\times$w)}
		\State h\_masks = h\_masks.reshape(num\_h, h*w)
		\State p\_masks = p\_masks.reshape(num\_p, h*w)	 
		\State inter\_matrix = mm(h\_masks, p\_masks.permute(1,0))
		\State part\_matrix = p\_masks.sum((-2,-1)).unsqueeze(0).expand(num\_h,num\_p)
		\State ratio = inter\_matrix/part\_matrix
	\end{algorithmic} 
	\label{code:ratio} 
\end{algorithm}

\section{Experiments}

We conduct comprehensive experiments and compare our method with state-of-the-art methods on the MHP v2.0 \cite{Zhao2018} and PASCAL-Person-Part \cite{Chen2014} datasets.

\subsection{Datasets}
\subsubsection{MHP v2.0 dataset}
The MHP v2.0 \cite{Zhao2018} dataset is the most challenging dataset for multi-human parsing. It contains 15,403 training images, 5,000 validation images with 59 part classes. Each image contains 2-26 persons, with 3 on average. It has the maximum number of classes in multi-human parsing to the best of our knowledge.

\subsubsection{PASCAL-Person-Part dataset}
The PASCAL-Person-Part \cite{Chen2014} dataset contains 1,716 images for training and 1,817 for testing. The annotations include six human parts: Head, Torso, Upper arms, Lower arms, Upper legs, and Lower legs. Each image contains 2.2 persons on average. 

\subsection{Experimental settings}
\subsubsection{Implementation details}
We implement the NTHP based on Pytorch end-to-end on a server with 2 NVIDIA GeForce GTX 1080Ti GPUs. A mini-batch involves 6 images. We use Group Normalization (GN) \cite{Wu2018} with group size 32. The shorter side of the image scales randomly from [544, 864] pixels, and the longer side is set to 1333 pixels. The inference is on a single scale of 1333 pixels for the longer side and 800 pixels for the shorter side. All models are trained with ResNet50 \cite{He2016} as backbone. As for the MHP dataset \cite{Zhao2018}, we trained for 12 epochs with an initial learning rate of 0.001 per GPU per image, which is decreased by 10 at the $9^{th}$ and again at the $11^{th}$ epoch. Weight decay is 0.0001 and momentum is 0.9. Values of experiments with longer learning schedule are 36, 0.001, 10, 27, 33, 0.0001 and 0.9. As for the PASCAL-Person-Part \cite{Chen2014} dataset, we train for 54 epochs and decrease the learning rate at the $45^{th}$ and the $51^{th}$ epoch. Other settings are the same as the MHP dataset.

\subsubsection{Evaluation metric}
We use separate evaluation metrics for the MHP v2.0 dataset \cite{Zhao2018} and the PASCAL-Person-Part dataset \cite{Chen2014} for a fair comparison with other networks.

To evaluate our network on the MHP dataset \cite{Zhao2018}, we use the metric named Average Precision based on Part ($AP^p$) which uses an average of part-level pixel IoU of different semantic part categories within a person instance to determine if one instance is a true positive and Percentage of Correctly Parsed Body Parts ($PCP$) which is the ratio between the correctly parsed categories and the total number of categories within a person. We use $AP^p$ with an IOU threshold of 0.5 ($AP^p_{50}$), the average of $AP^p$ with IOU thresholds ranging from 0.1 to 0.9 with a step size of 0.1 ($AP^p_{vol}$), and $PCP$ with an IOU threshold of 0.5 ($PCP_{50}$). All these metrics are proposed by \cite{Li2017} to evaluate the network performance on the MHP dataset.

As for the PASCAL-Person-Part dataset, we use Mean Average Precision ($AP^r$) first proposed for evaluating instance segmentation results by SDS \cite{Hariharan2014} and is used by nearly all methods to compare the performance of instance-level multi-human parsing result on the PASCAL-Person-Part dataset \cite{Chen2014}. We use $AP^r$ separately with IOU thresholds of 0.5, 0.6, 0.7, and $AP^r_{vol}$ as an average of $AP^r$ with IOU thresholds ranging from 0.1 to 0.9 with a step size of 0.1.

\subsection{Experimental results}
All our experiments are implemented on the MHP v2.0 dataset \cite{Zhao2018} with ResNet50 \cite{He2016} as the backbone and trained for 12 epochs.

\subsubsection{Unified prototype generation structure}
We conduct experiments on two settings of using 1) a unified collection of prototypes for humans and parts, as shown in Figure \ref{figure4}, 2) two groups of prototypes. As for the second set, we use the same structure as the first set but duplicate it twice to deal with humans and parts separately. We show the result in Table \ref{tab:table2}.

From Table \ref{tab:table2}, we can see that adopting a unified structure brings a better result. Besides, the second setting costs more memory since it doubles the feature pyramid fusion process.

\begin{table}[h!]
	\caption{Unified vs. Separate prototype generation structures. }
	\begin{center}
		\begin{tabular}{c c c c}
			\hline
			Prototypes & $AP^p_{50}$ & $AP^p_{vol}$ & $PCP_{50}$ \\\hline
			1) Unified & \textbf{41.8} & \textbf{46.2} & \textbf{41.9}  \\
			2) Separate & 41.5 & \textbf{46.2} & 41.8 \\  \hline
		\end{tabular}
		\label{tab:table2}
		
	\end{center}
	
\end{table}

\subsubsection{Not sharing weights between humans and parts}
We also conduct a series of experiments on whether to share weights across different levels between humans and parts on mask coefficient and category branches.

We show the results in Table \ref{tab:table3}. We can see that not sharing weights between them obtains the best results. For the category branch, we consider that this is because the classes predicted between humans and parts are different so that the network needs to learn diverse features. As for the mask coefficient branch, the smallest unit predicted in prototypes is in part level, and sometimes humans are fused by a combination of parts, thus requiring different features.
\begin{table}[h!]
	\caption{Share weights or not between humans and parts. Yes or no represents whether adding deformable convolutions or not. }
	\begin{center}
		\begin{tabular}{c c c c c}
			\hline
			Mask coefficients & Category & $AP^p_{50}$ & $AP^p_{vol}$ & $PCP_{50}$ \\\hline
			no & no & \textbf{41.8} & \textbf{46.2} & \textbf{41.9} \\ 
			 yes&no& 41.3 & 46.1 & 41.6  \\ 
			 yes&yes & 41.1 & 45.9 & 41.5  \\ \hline
		\end{tabular}
		\label{tab:table3}	
	\end{center}
	
\end{table}

\subsubsection{Prototype generation structure}
We compare two choices of mask prototype generation structure, 1) feature pyramid fusion proposed in \cite{Kirillov2019} with Coordconv \cite{RosanneLiuJoelLehmanPieroMolino2018} as shown in Figure 4, 2) the structure similar to the one used in YOLACT \cite{Bolya2019}. As for the second choice, we use $4$$\times$$convs$ ($3$$\times$$3$) for feature extraction and one ($1$$\times$$1$) for prediction after the finest level of FPN with the final resolution 1/4 of the original image.

Our network utilizes the unique attribute of FPN \cite{Lin2017a}, which concentrates on small objects on lower levels but big ones on higher levels. The features obtained by choice two are not enough to generate prototypes with a large range of scales. On the contrary, choice one obtains detailed information and semantic information simultaneously. As shown in Table \ref{tab:table4}, using the finest level alone degrades the performance by a large margin.

\begin{table}[h!]
	\caption{Prototype generation structure. }
	\begin{center}
		\begin{tabular}{c c c c}
			\hline
			Generation choice & $AP^p_{50}$ & $AP^p_{vol}$ & $PCP_{50}$ \\\hline
			1) choice one & \textbf{41.8} & \textbf{46.2} & \textbf{41.9} \\ 
			2) choice two  & 32.1 & 41.4 & 33.7  \\ \hline
		\end{tabular}
		\label{tab:table4}	
	\end{center}
\end{table}

\subsubsection{Other experiments}
We find that adding deformable convolutions \cite{Zhu2019} in our network can obtain considerable improvement. We add deformable convolutions \cite{Zhu2019} in the backbone and the prototype generation module. Besides, increasing iterations is a common method to improve the performance. We also try training our network for 36 epochs. We show the results in Table \ref{tab:table5}.
	
\begin{table}[h!]
	\begin{center}
		\caption{Other experiments on MHP v2.0. DCN means deformable convolutions. Yes or no represents whether adding deformable convolutions or not.}
		\begin{tabular}{c  c  c c c c}
			\hline
			Baseline & DCN & Epochs & $AP^p_{50}$ & $AP^p_{vol}$ & $PCP_{50}$ \\\hline
			\multirow{3}* {ResNet50}  & no & 12 & 41.8 & 46.2 & 41.9 \\ 
			& yes & 12& 46.0 &47.7 & 45.2\\
			&yes &36 & \textbf{51.1} & \textbf{49.5} & \textbf{49.9} \\ \hline
		\end{tabular}
		\label{tab:table5}	
	\end{center}
\end{table}

\subsection{Comparisons with the state-of-the-art methods}

We compare NTHP with state-of-the-art methods on the MHP v2.0 \cite{Zhao2018} and PASCAL-Person-Part \cite{Chen2014} datasets.

\begin{figure}[htbp]
	\centering
	\subfigure
	{
		\begin{minipage}[b]{0.16071\linewidth}
			\centering
			\includegraphics[height=1.5cm,width=2.25cm]{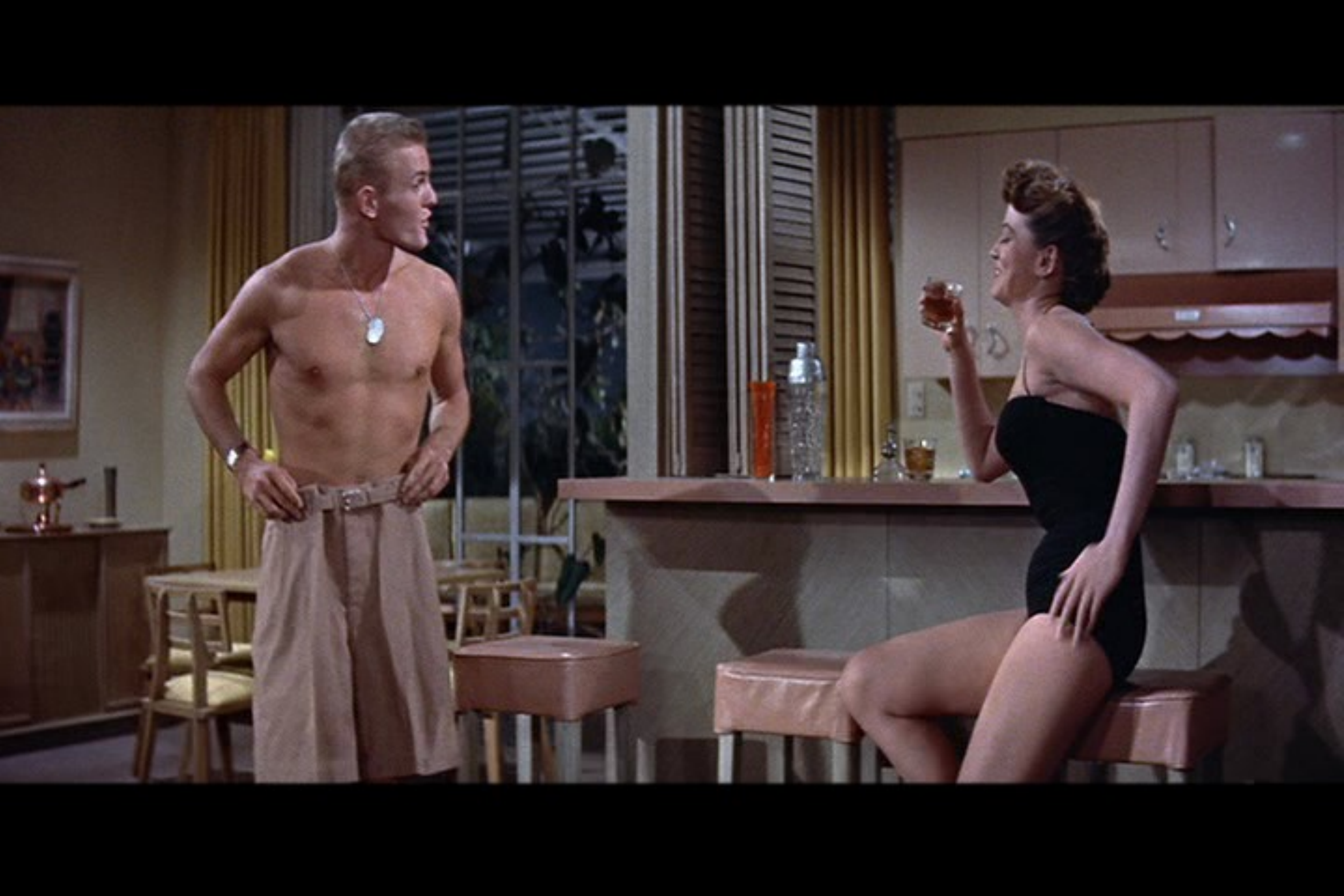} \\
			\includegraphics[height=1.5cm,width=2.25cm]{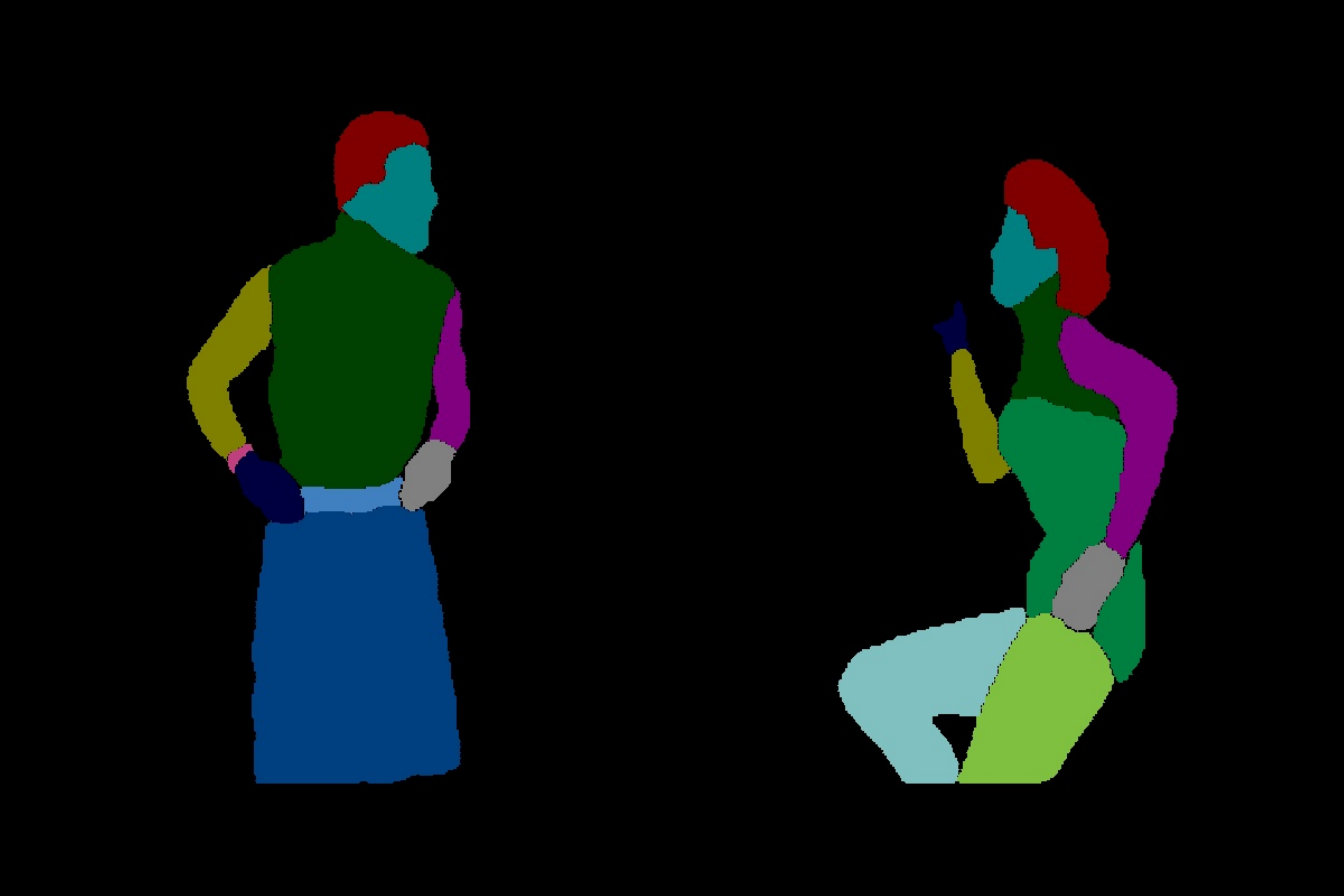} \\
			\includegraphics[height=1.5cm,width=2.25cm]{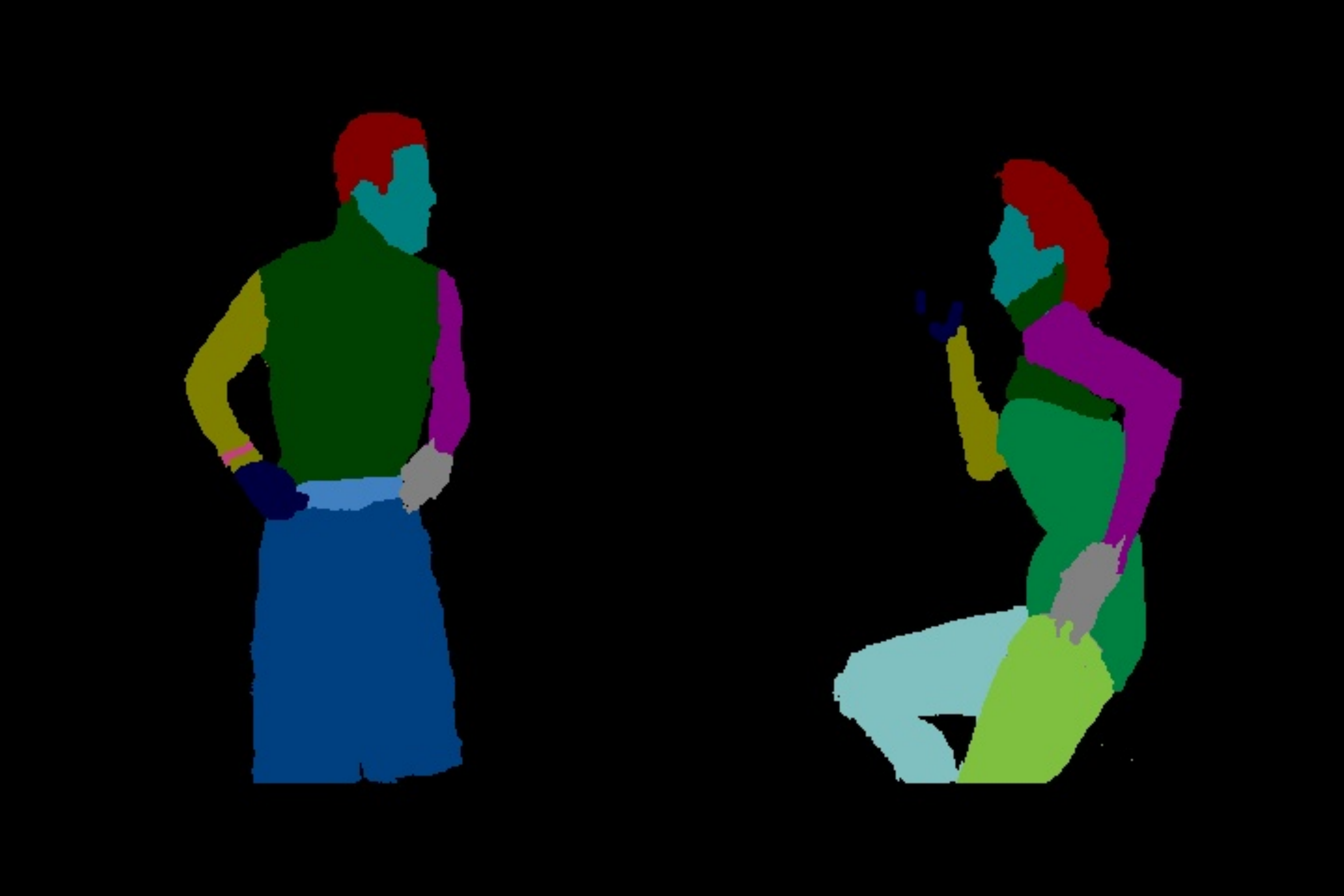} \\
			\includegraphics[height=1.5cm,width=2.25cm]{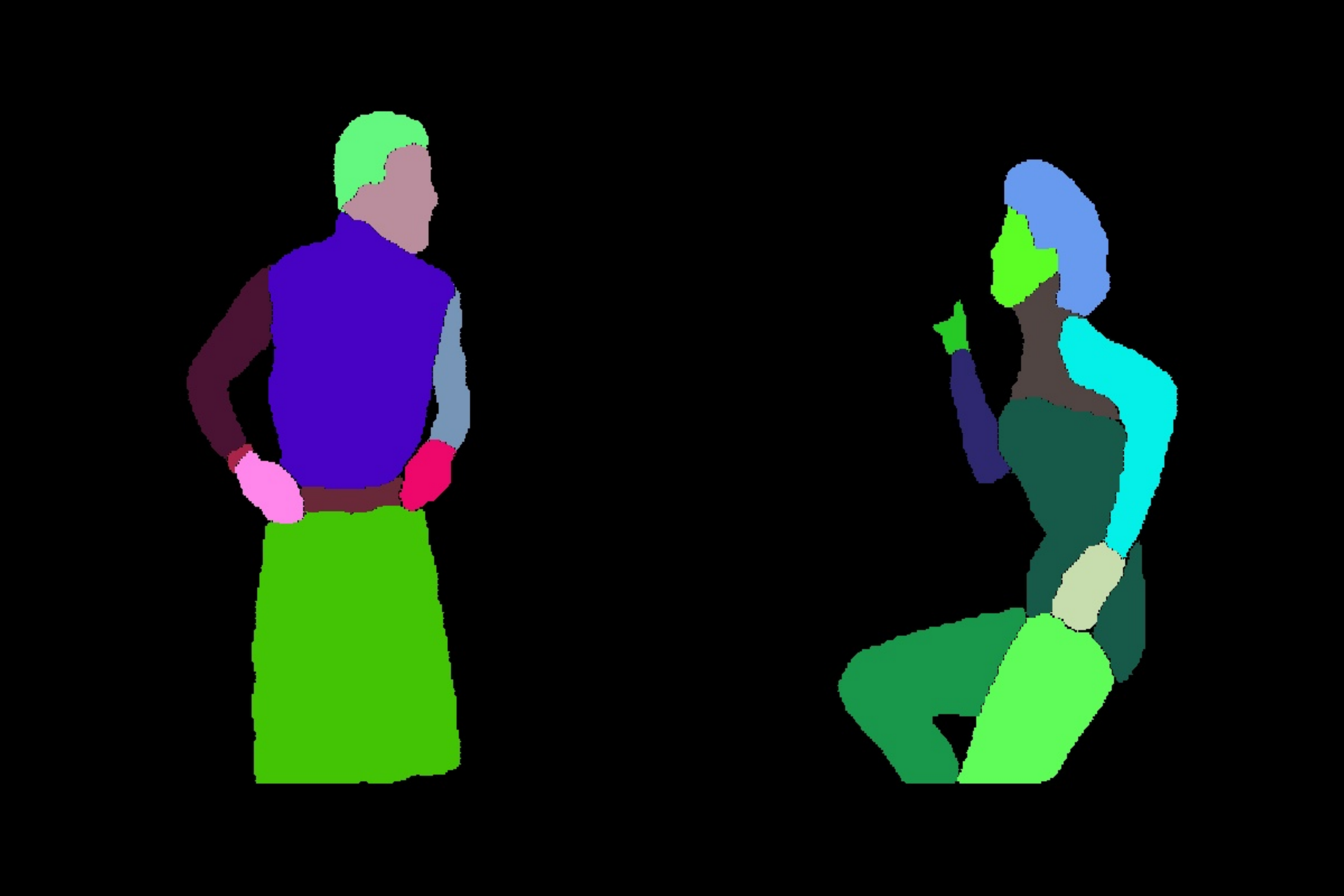} \\
			\includegraphics[height=1.5cm,width=2.25cm]{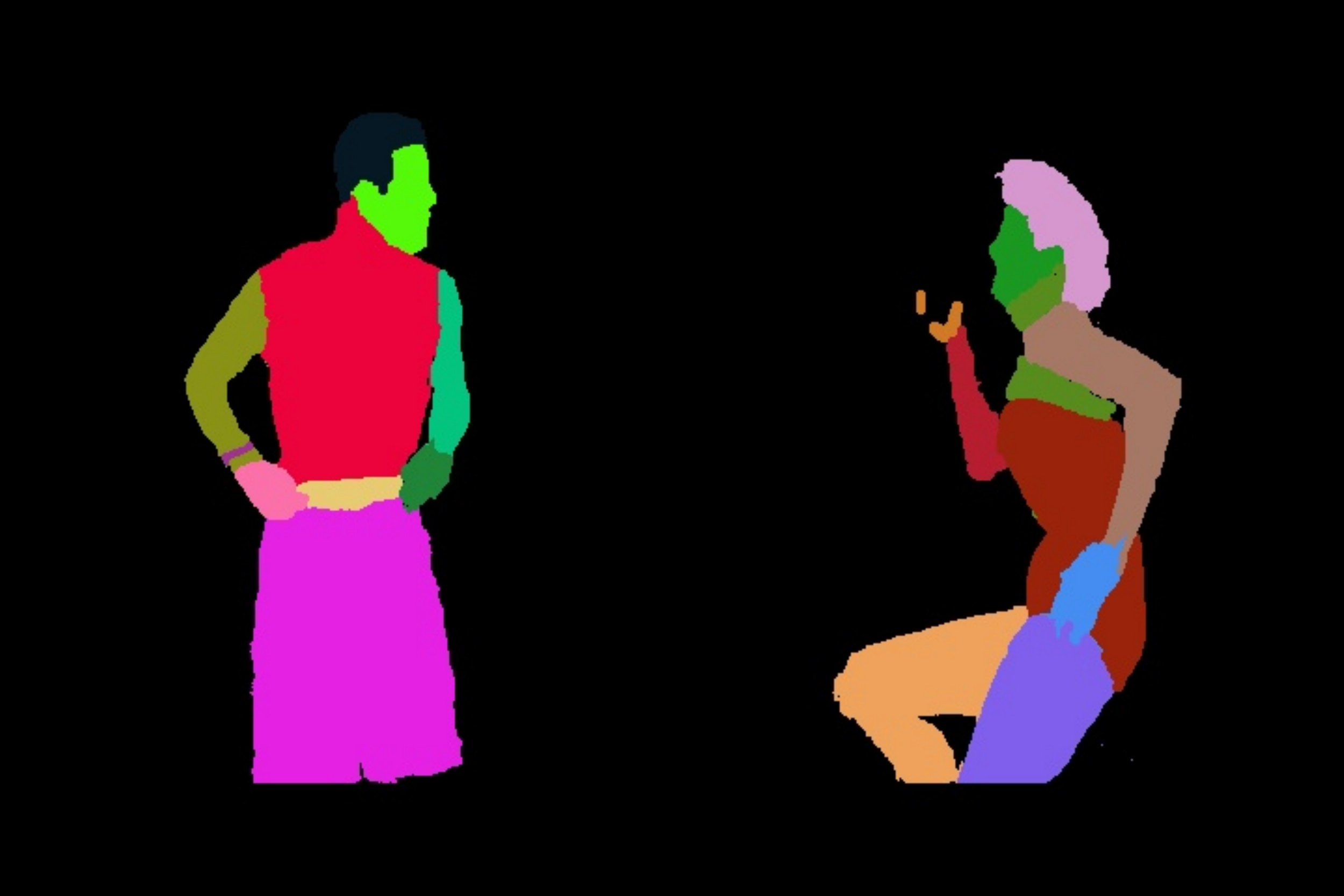} \\
			\includegraphics[height=1.5cm,width=2.25cm]{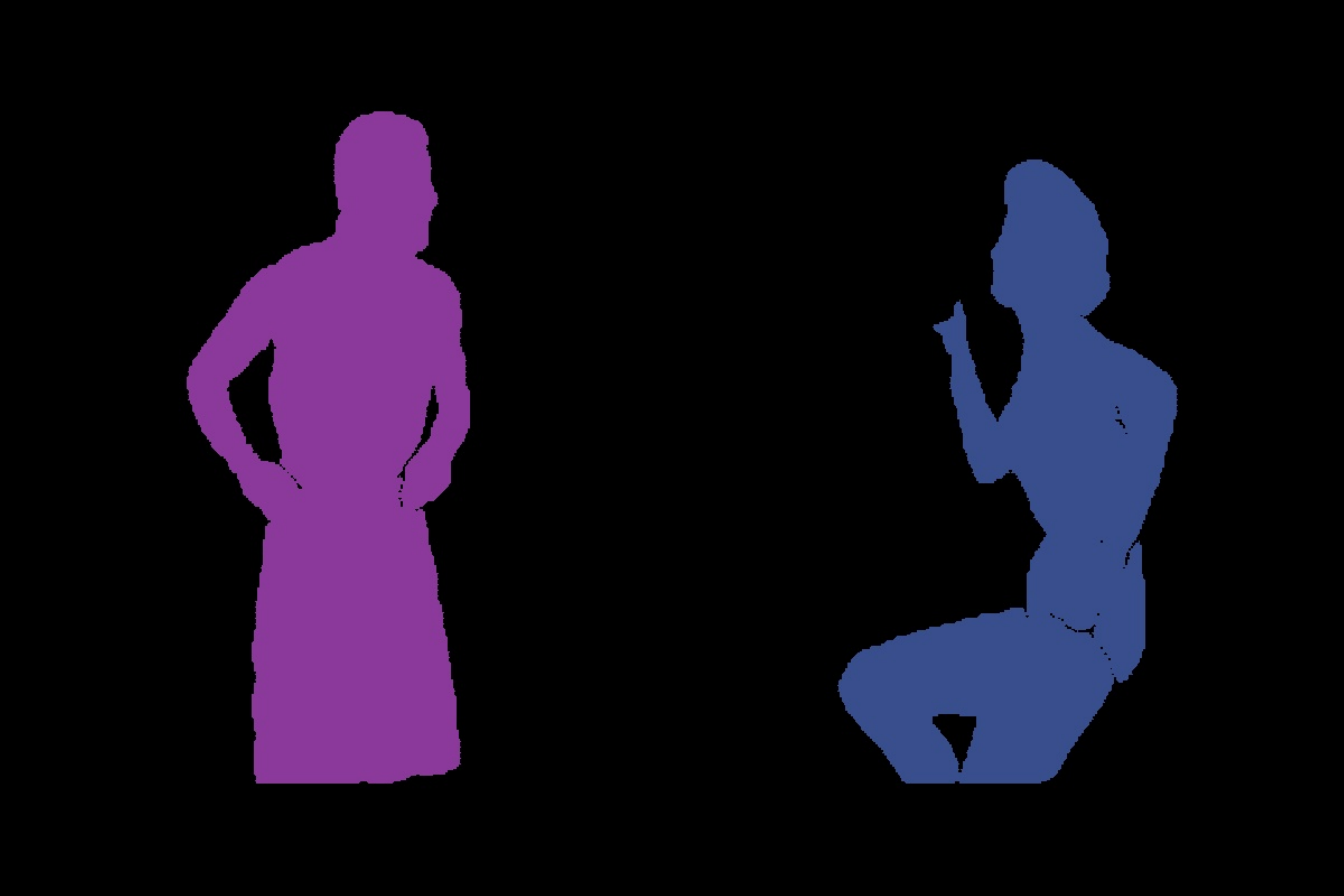} \\
			\includegraphics[height=1.5cm,width=2.25cm]{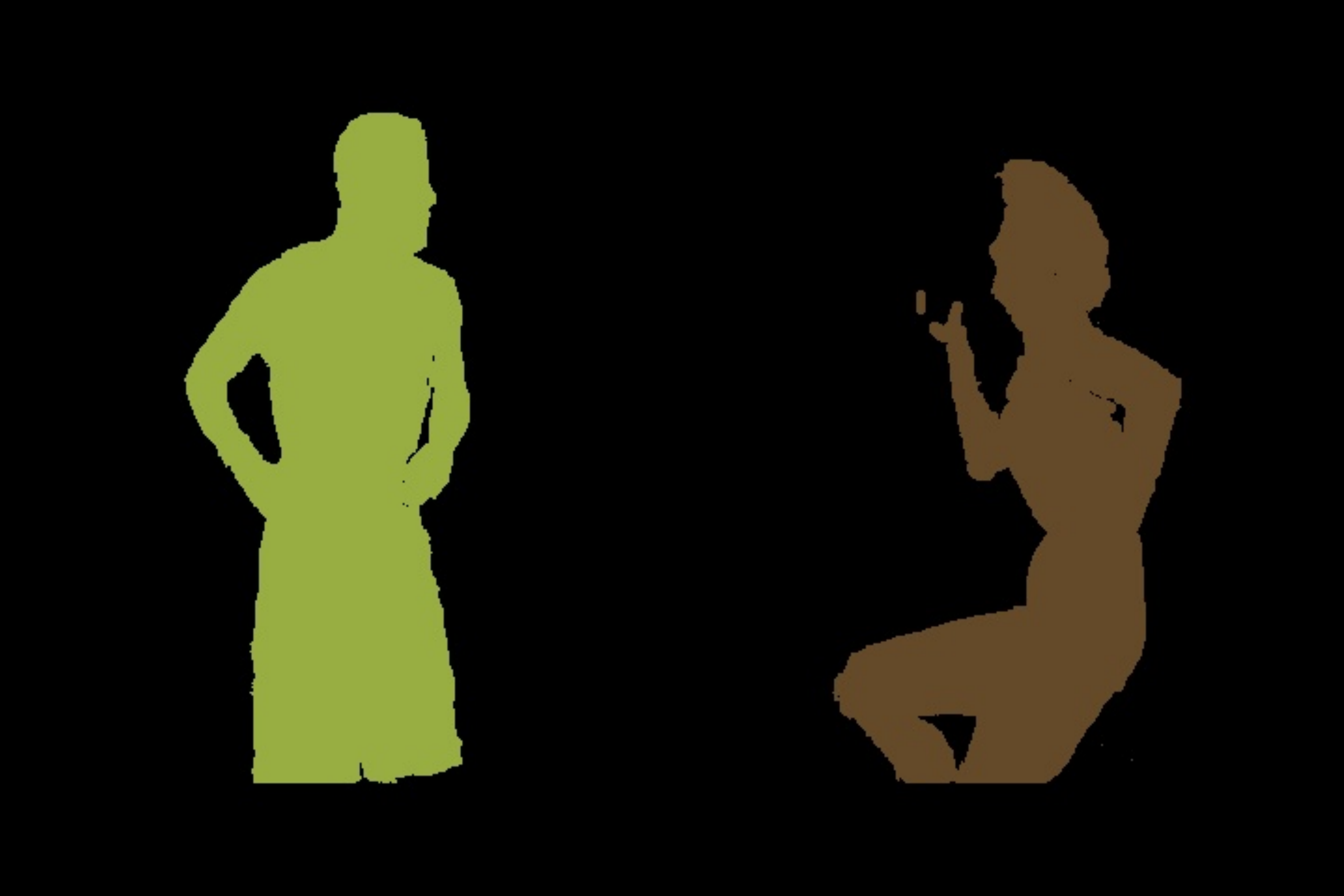} \\
		\end{minipage}
	}
	\subfigure
	{
	\begin{minipage}[b]{0.057\linewidth}
		\centering
		\includegraphics[height=1.5cm,width=1cm]{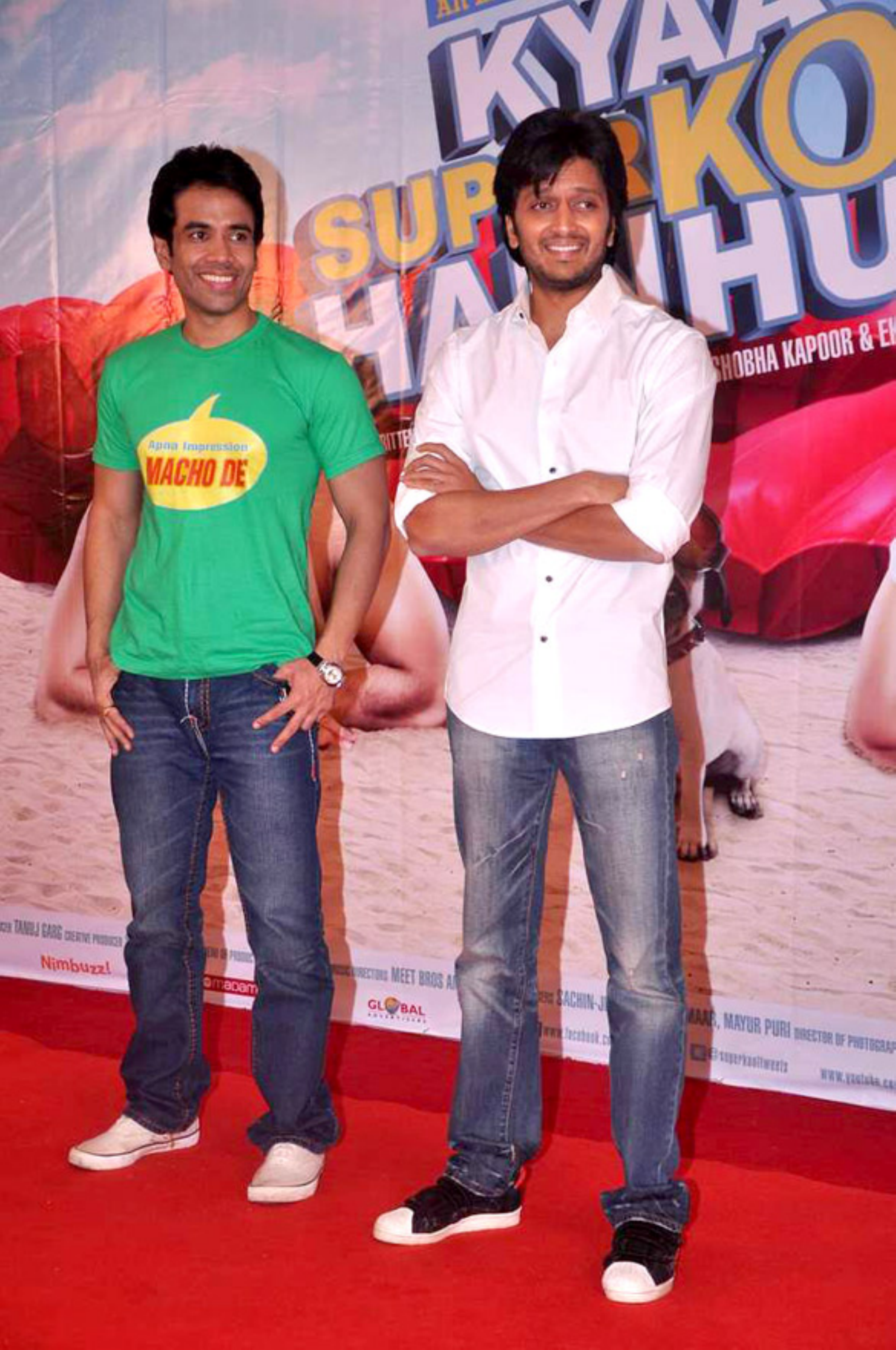} \\
		\includegraphics[height=1.5cm,width=1cm]{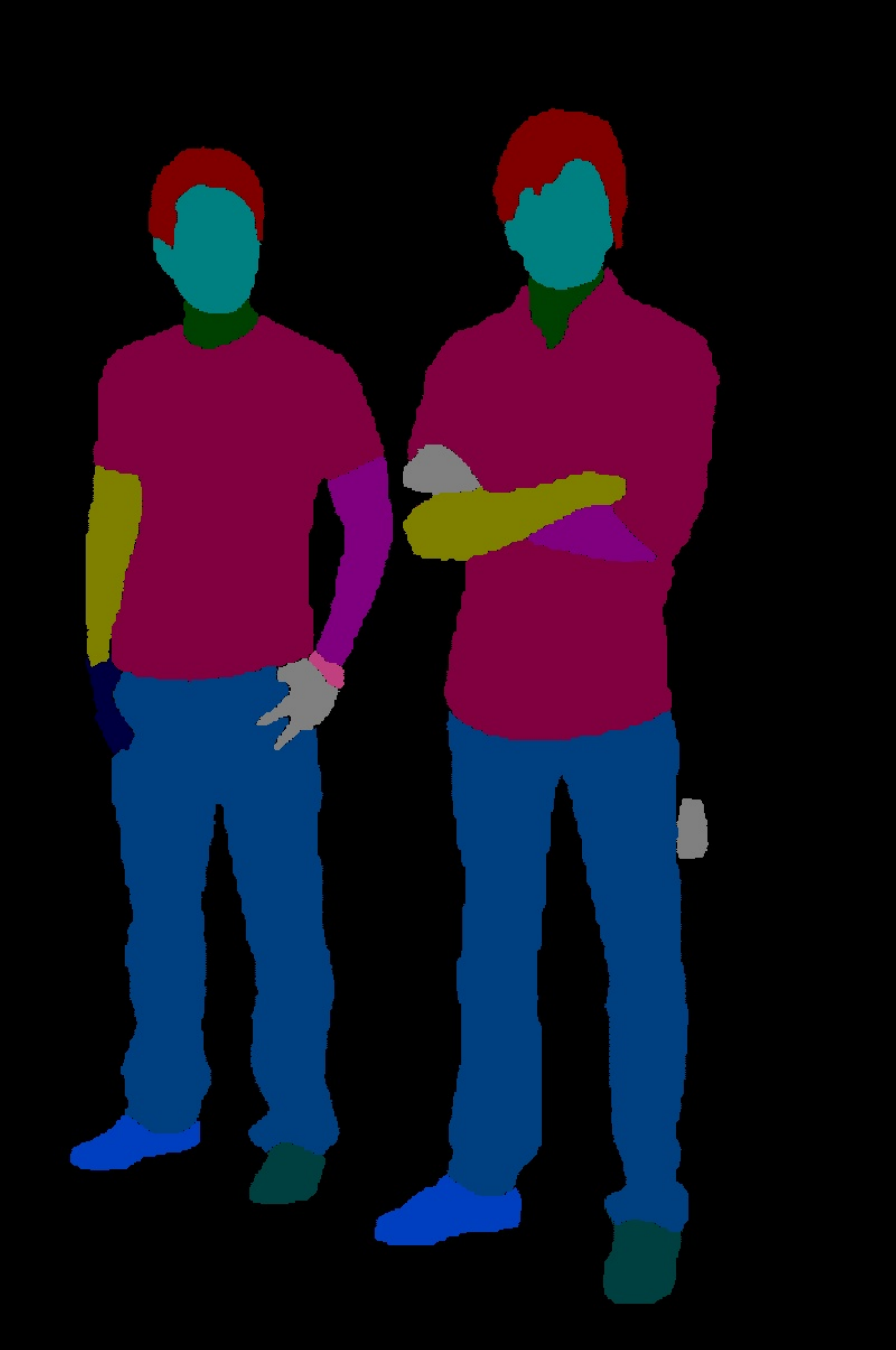} \\
		\includegraphics[height=1.5cm,width=1cm]{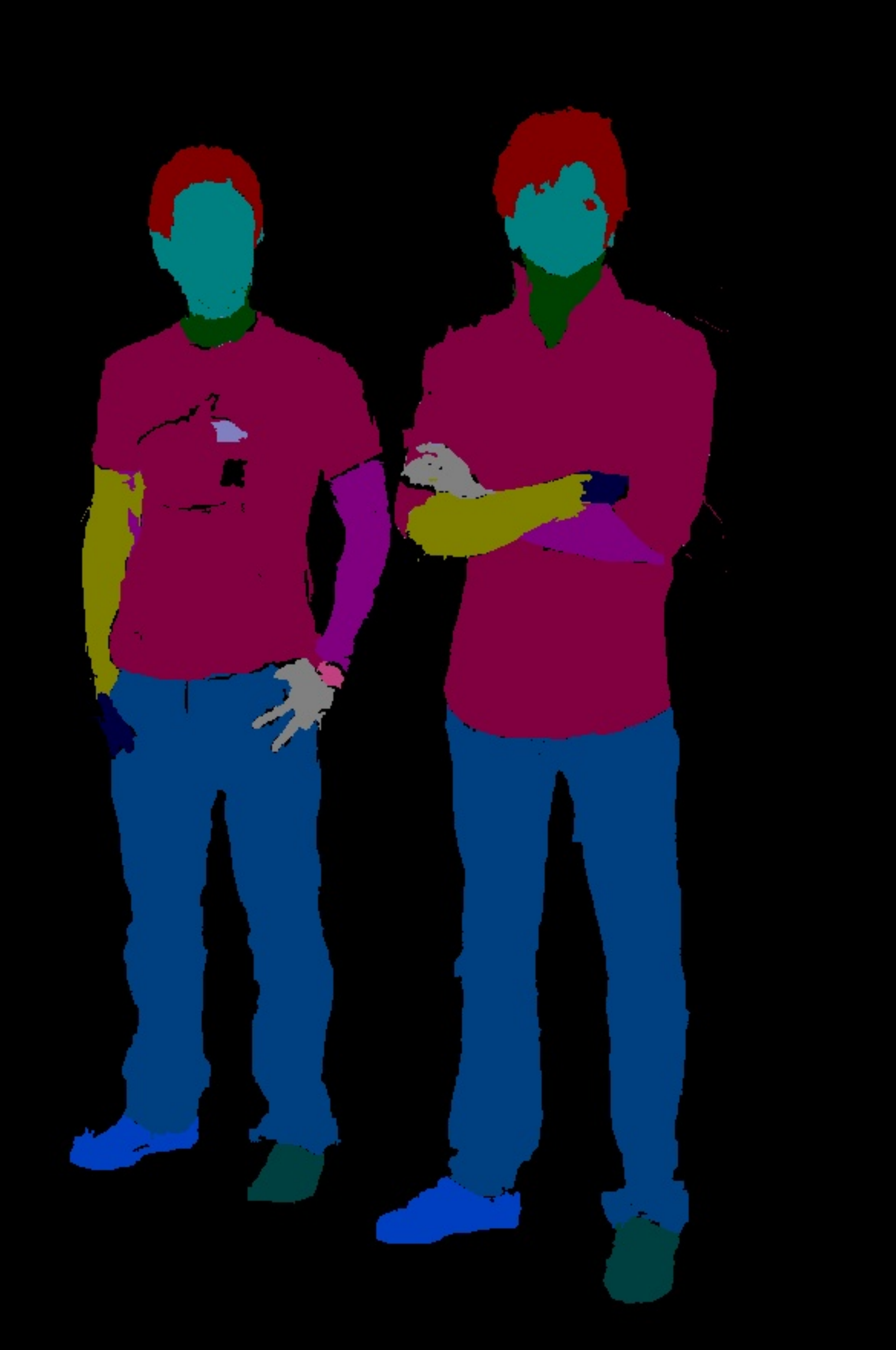} \\
		\includegraphics[height=1.5cm,width=1cm]{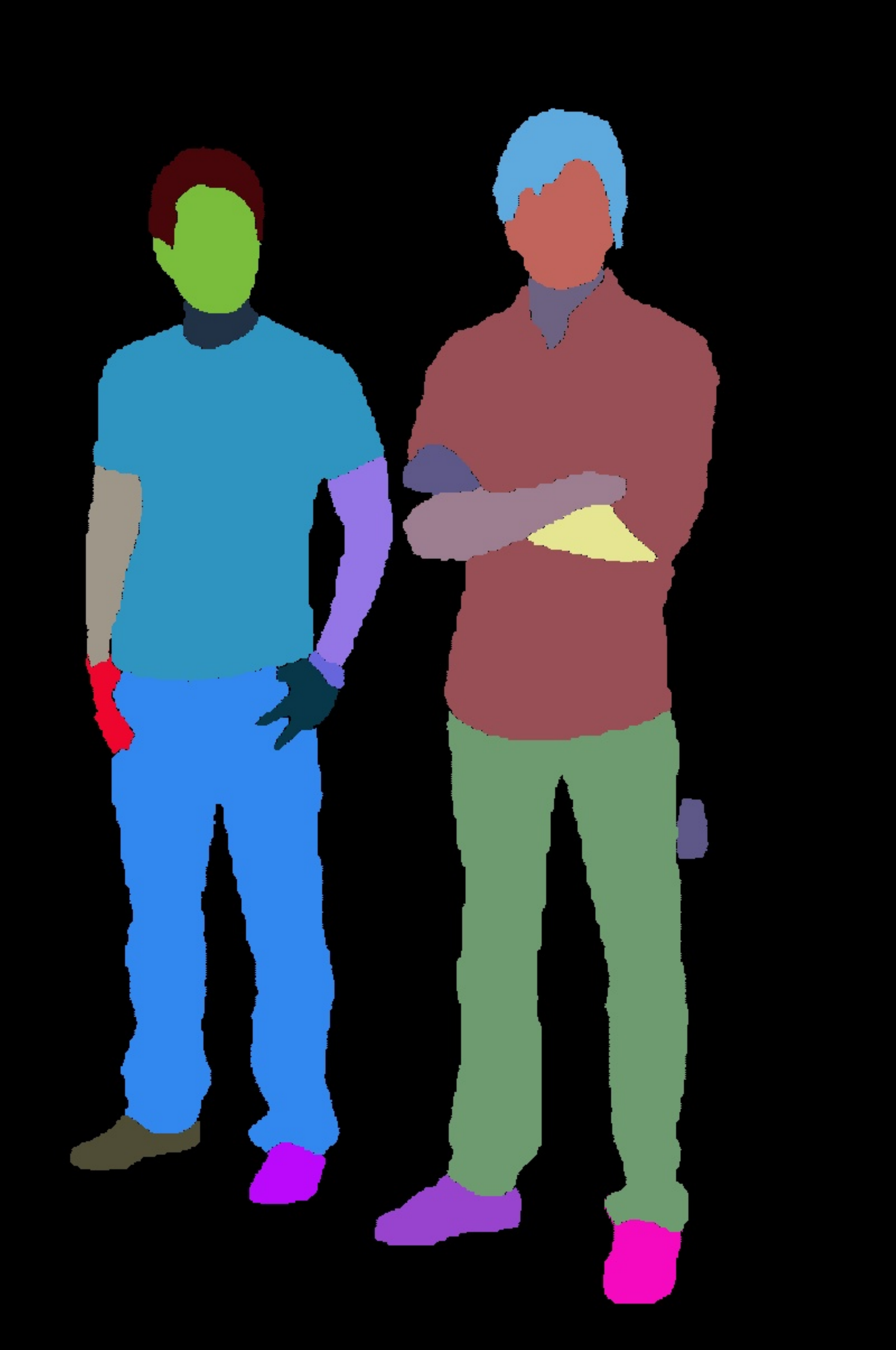} \\
		\includegraphics[height=1.5cm,width=1cm]{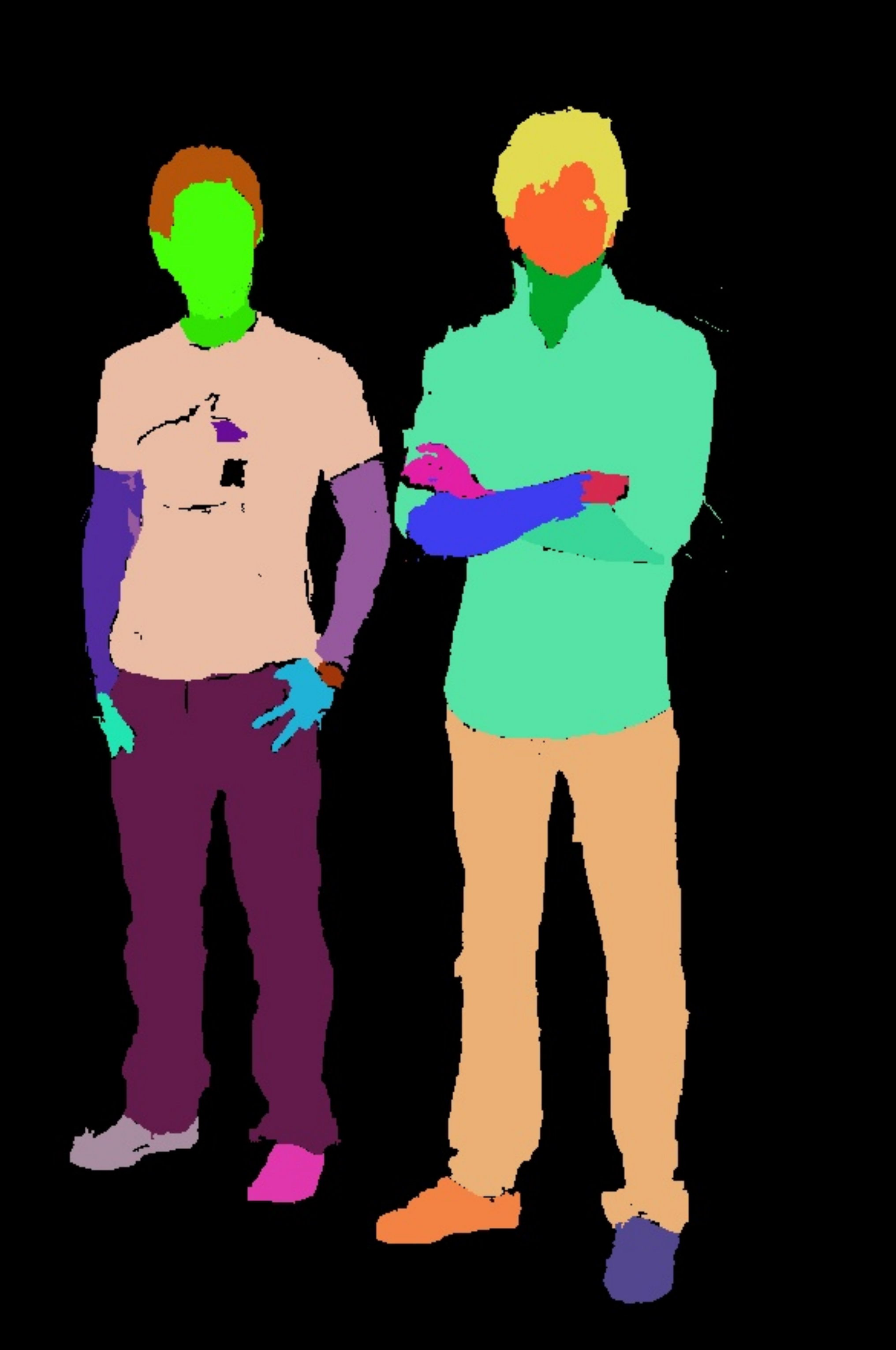} \\
		\includegraphics[height=1.5cm,width=1cm]{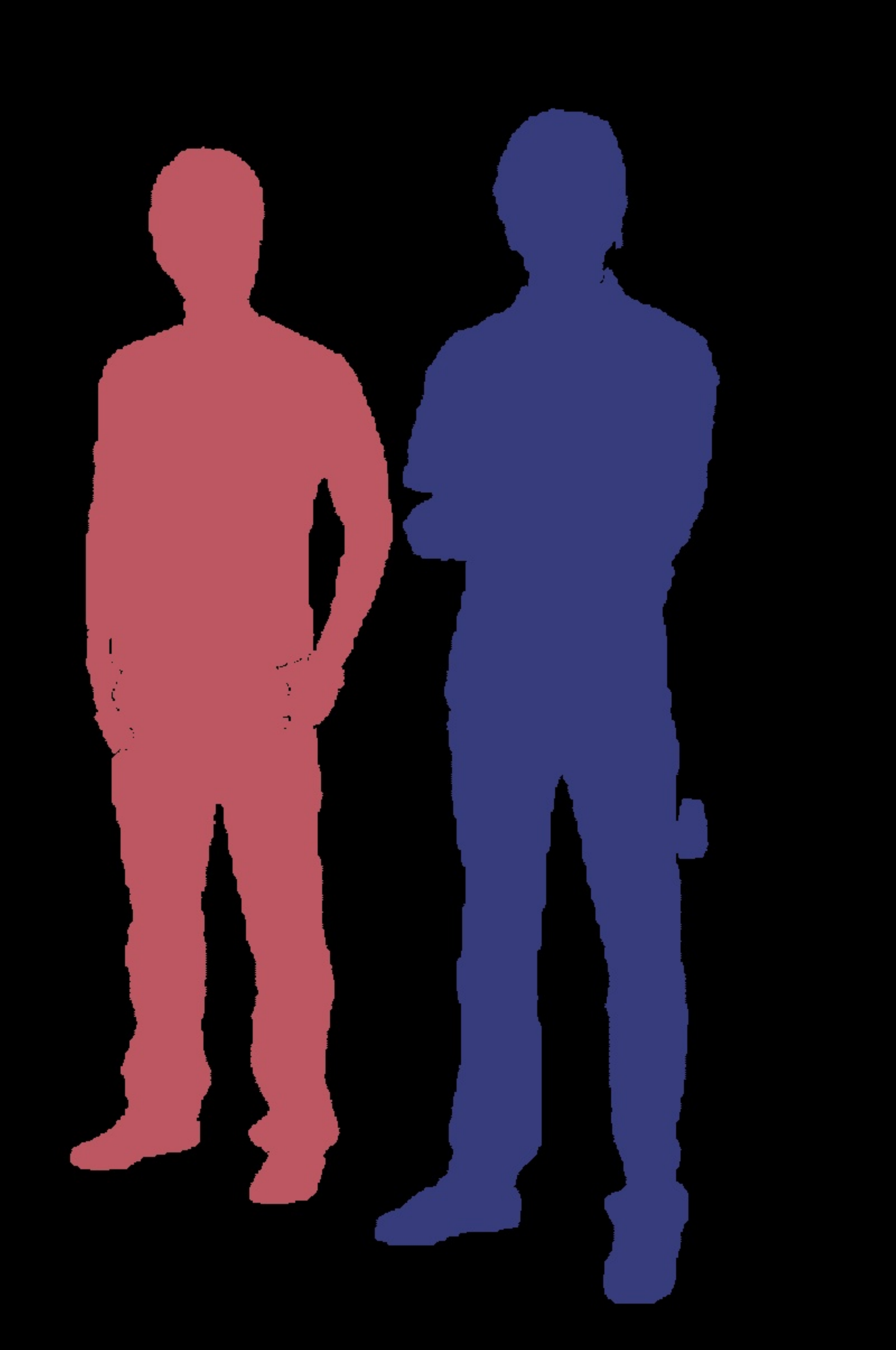} \\
		\includegraphics[height=1.5cm,width=1cm]{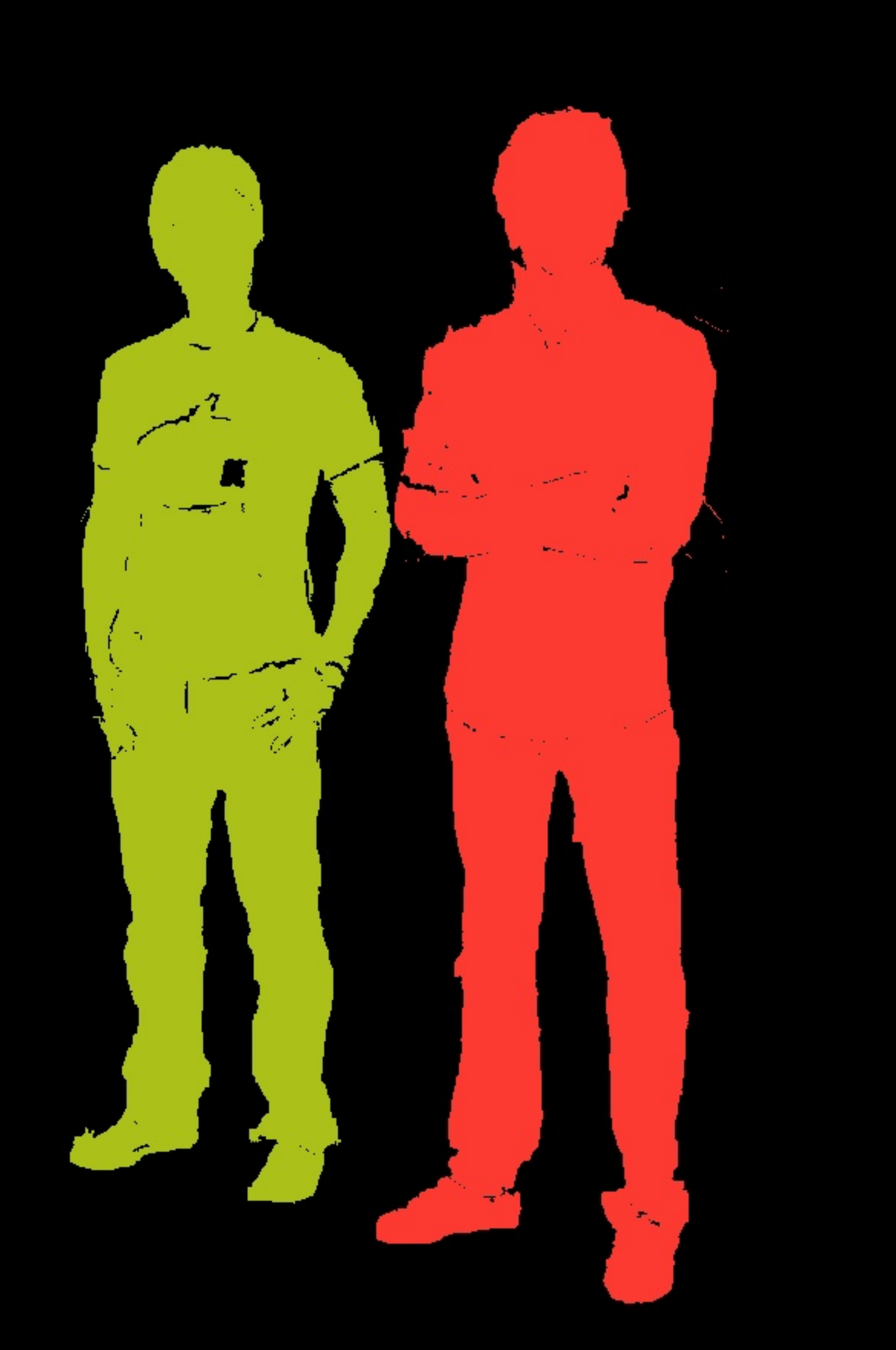} \\
	\end{minipage}
	}
	\subfigure
	{
		\begin{minipage}[b]{0.16071\linewidth}
			\centering
			\includegraphics[height=1.5cm,width=2.25cm]{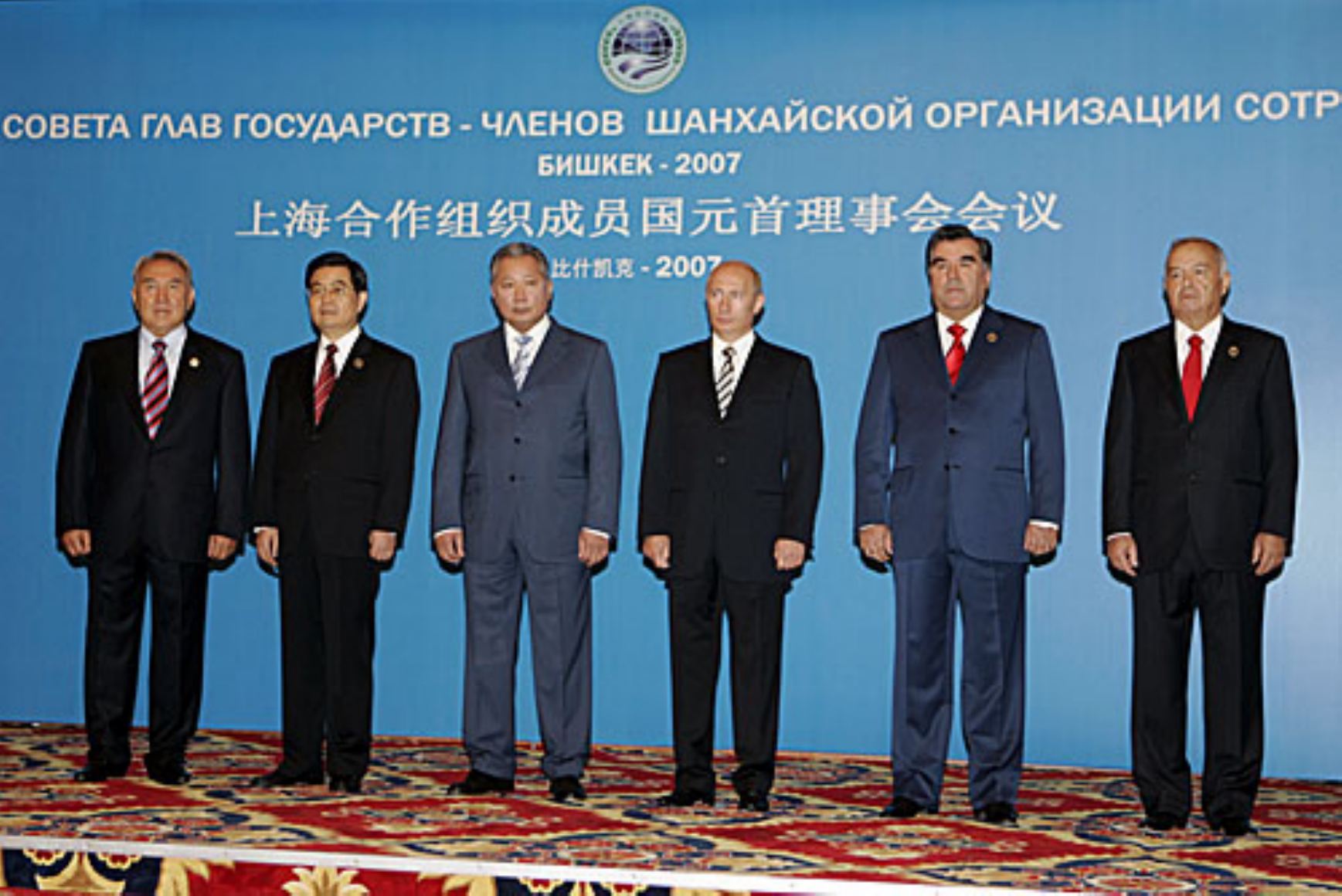} \\
			\includegraphics[height=1.5cm,width=2.25cm]{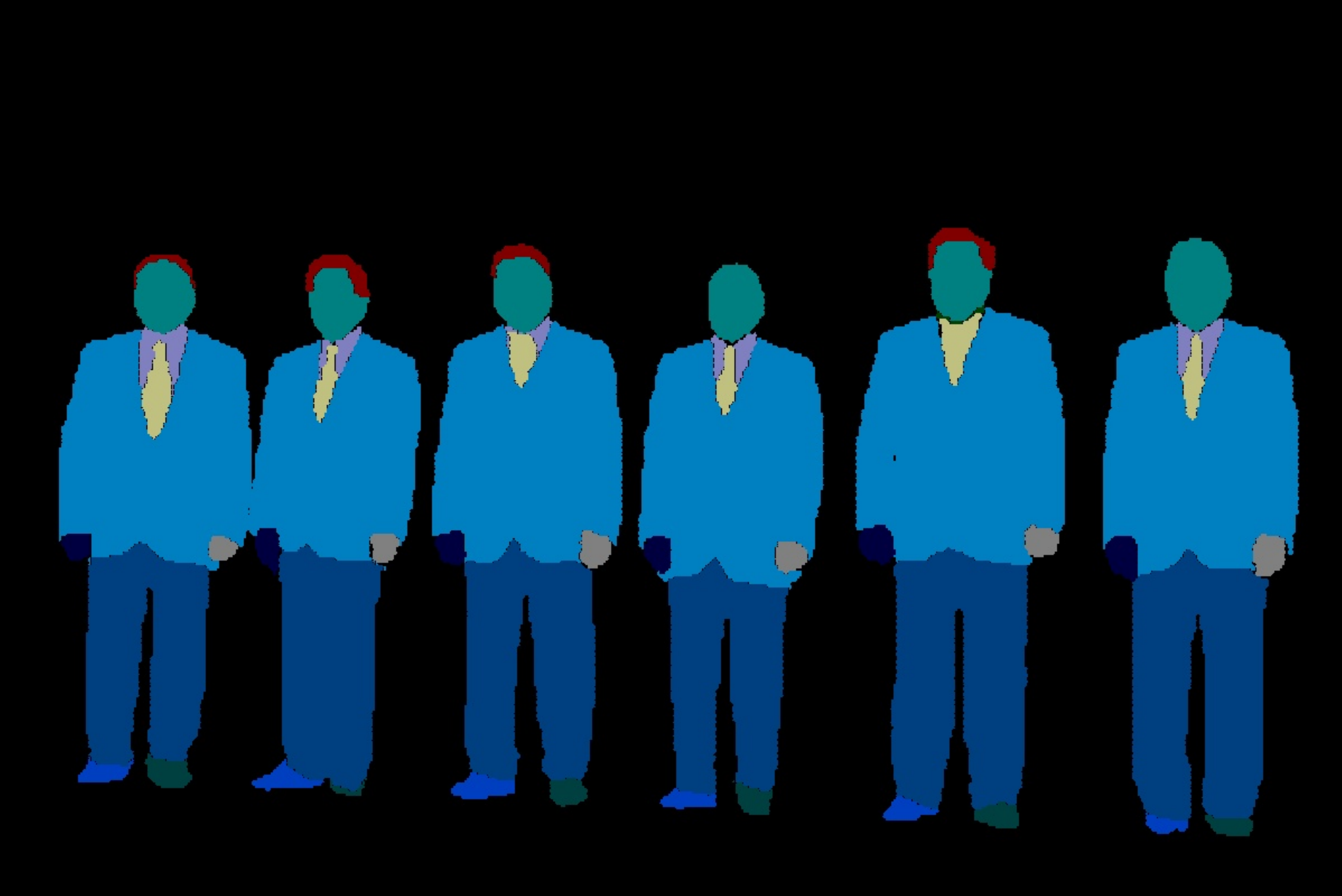} \\
			\includegraphics[height=1.5cm,width=2.25cm]{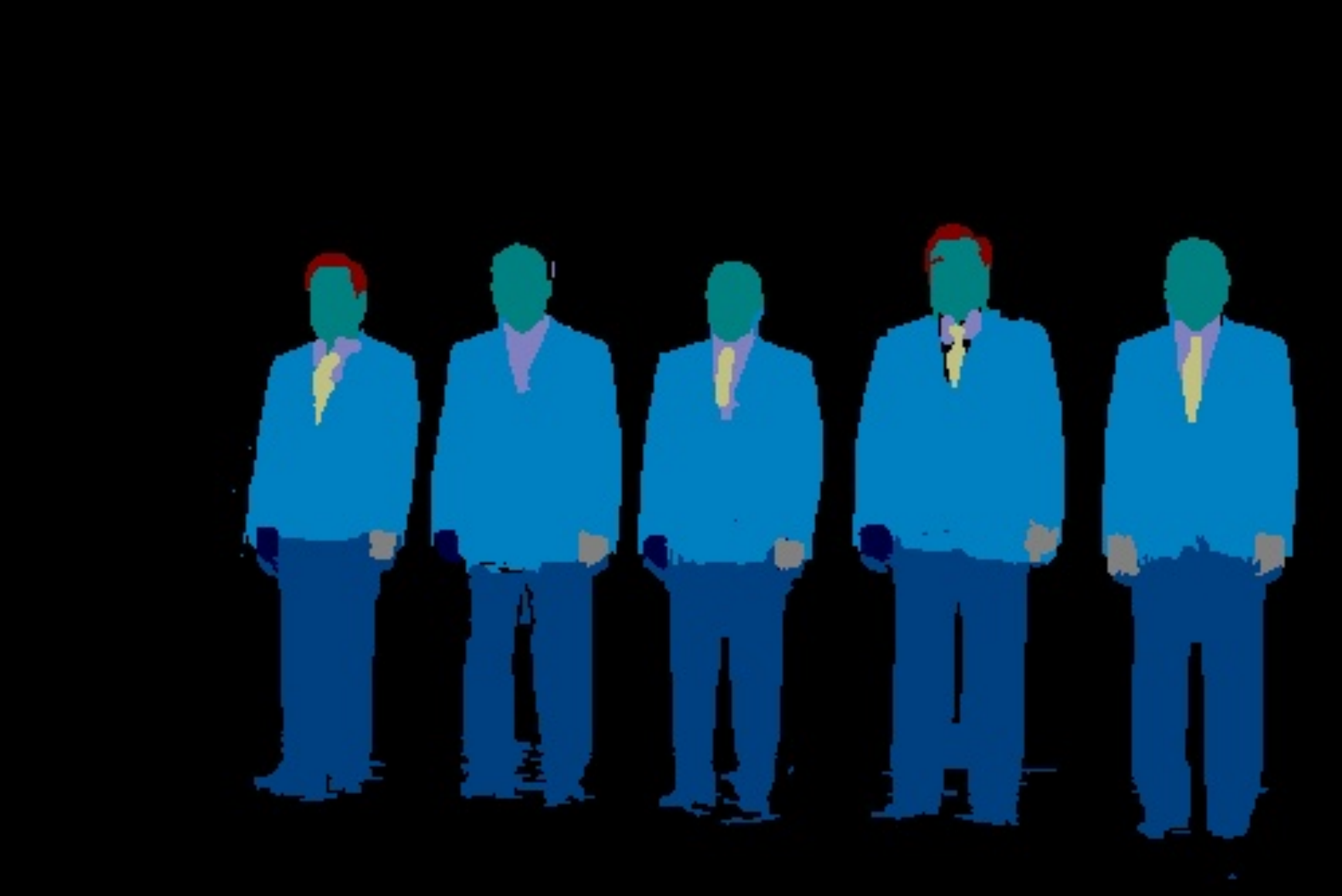} \\
			\includegraphics[height=1.5cm,width=2.25cm]{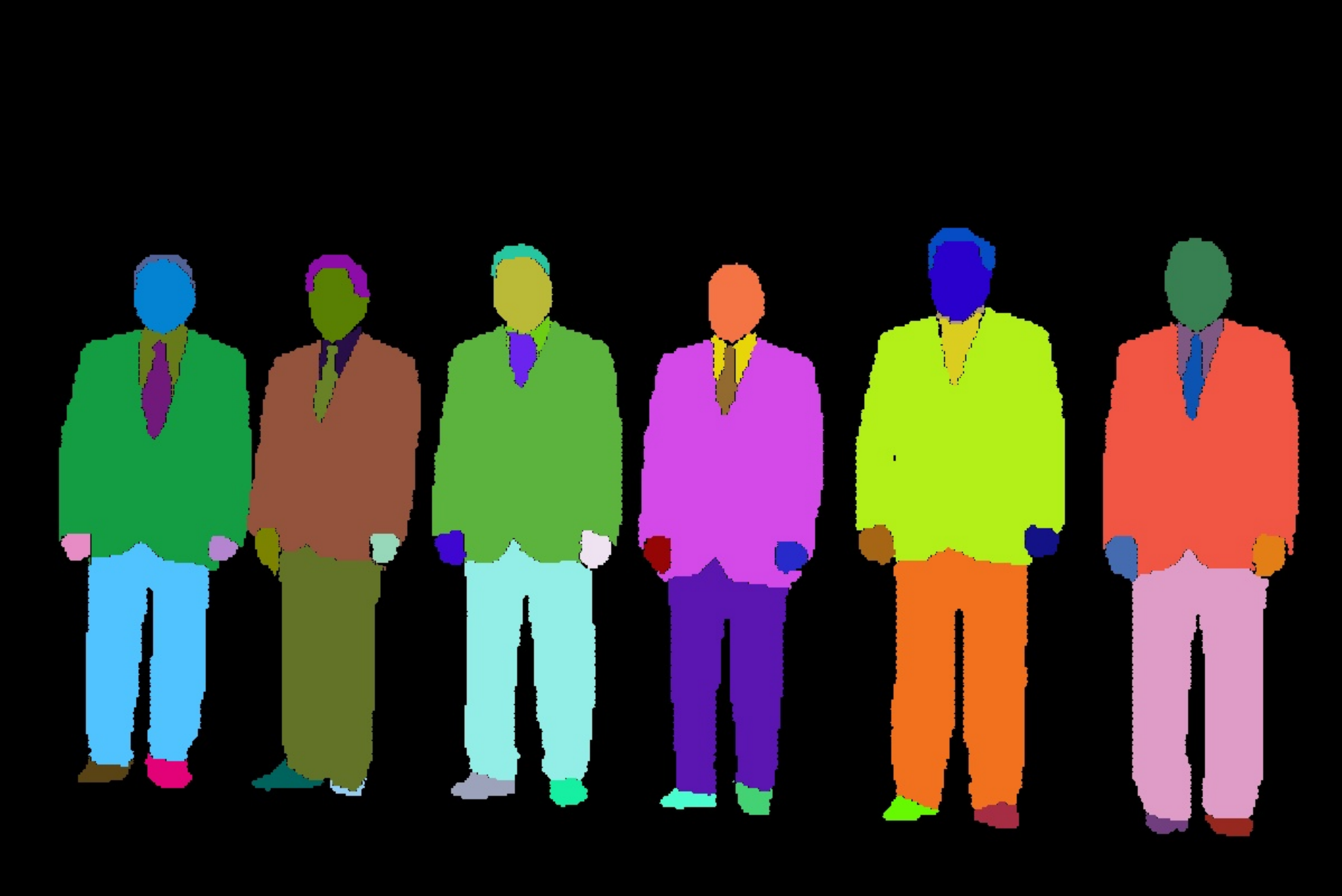} \\
			\includegraphics[height=1.5cm,width=2.25cm]{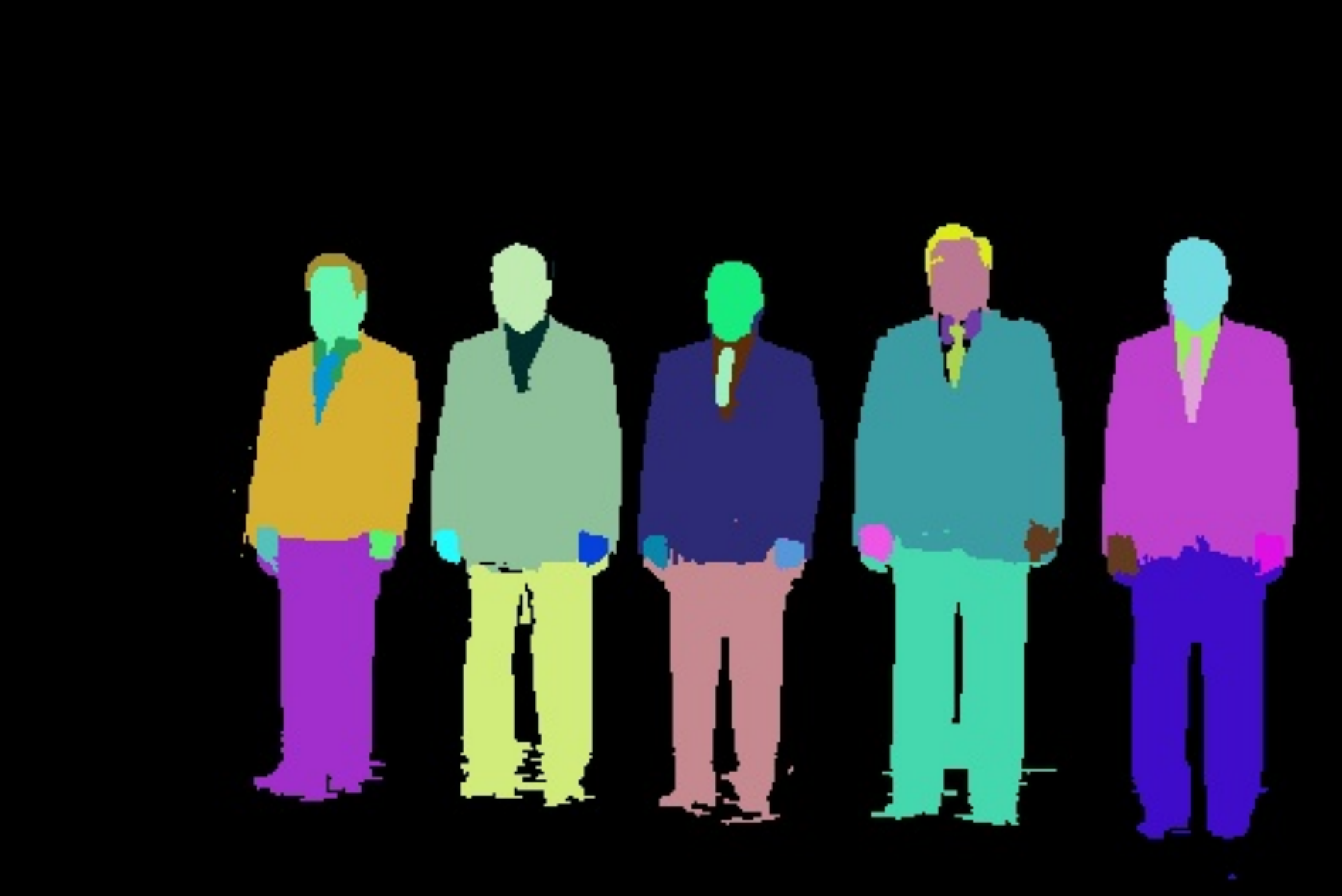} \\
			\includegraphics[height=1.5cm,width=2.25cm]{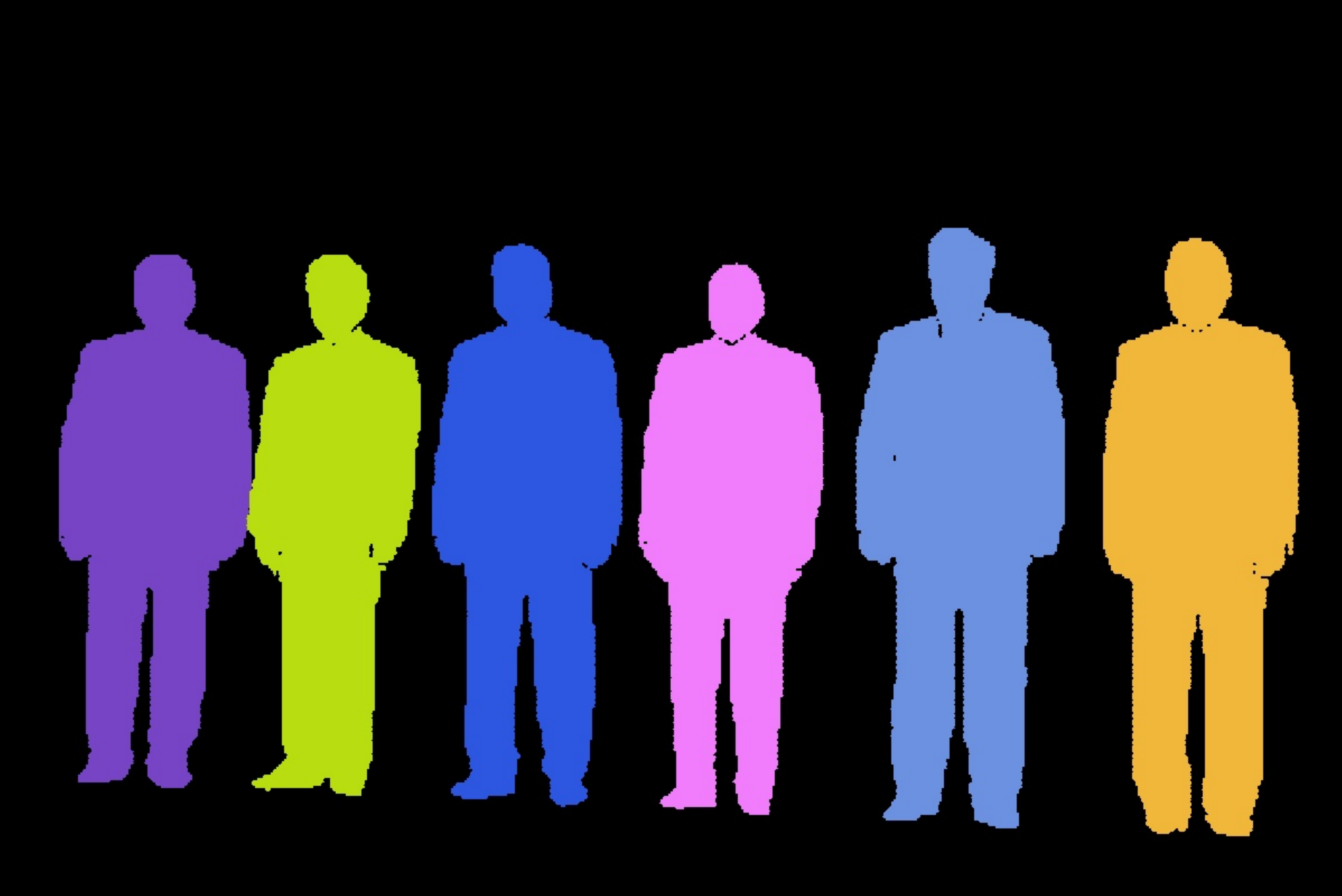} \\
			\includegraphics[height=1.5cm,width=2.25cm]{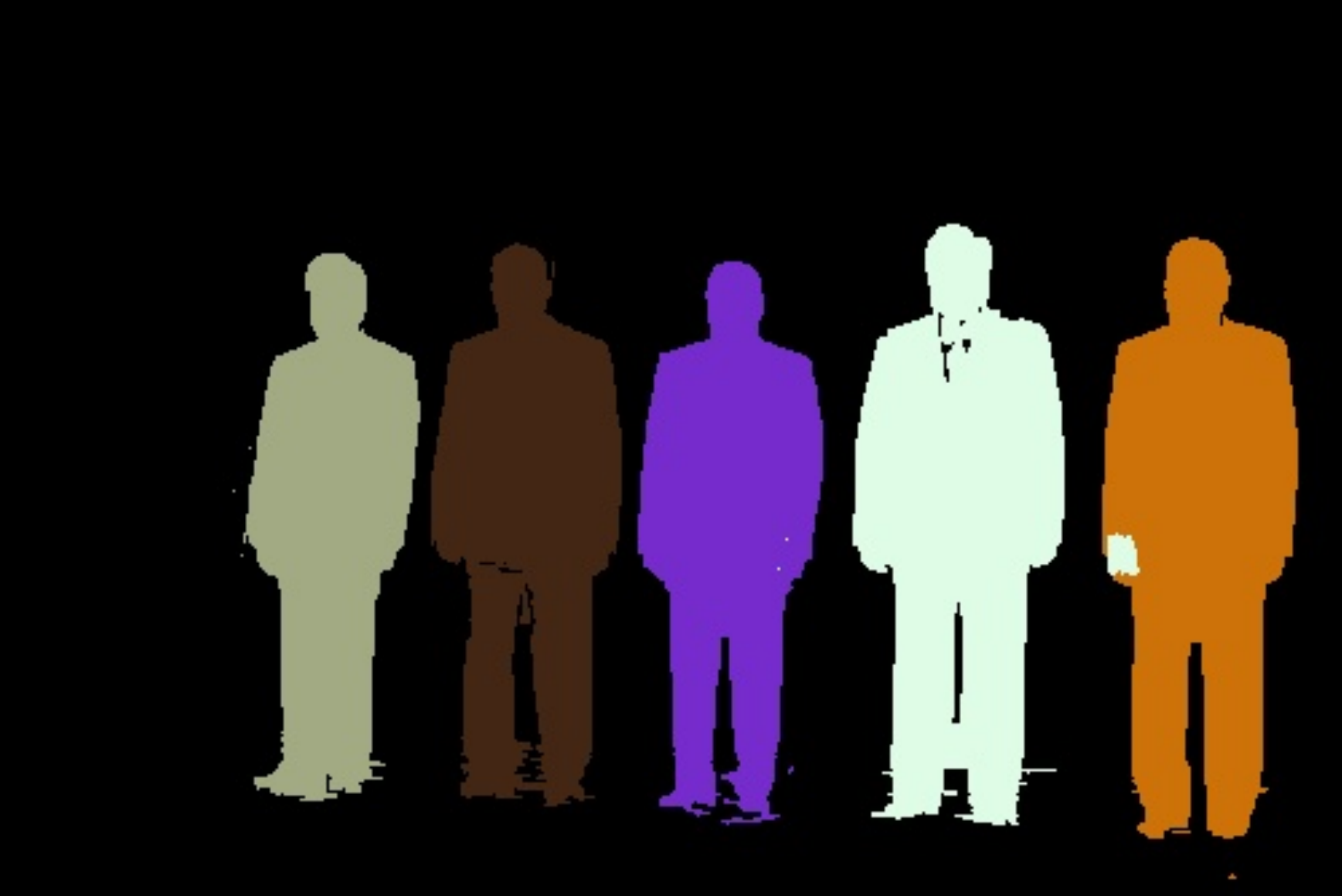} \\
		\end{minipage}
	}
	\subfigure
	{
		\begin{minipage}[b]{0.16071\linewidth}
			\centering
			\includegraphics[height=1.5cm,width=2.25cm]{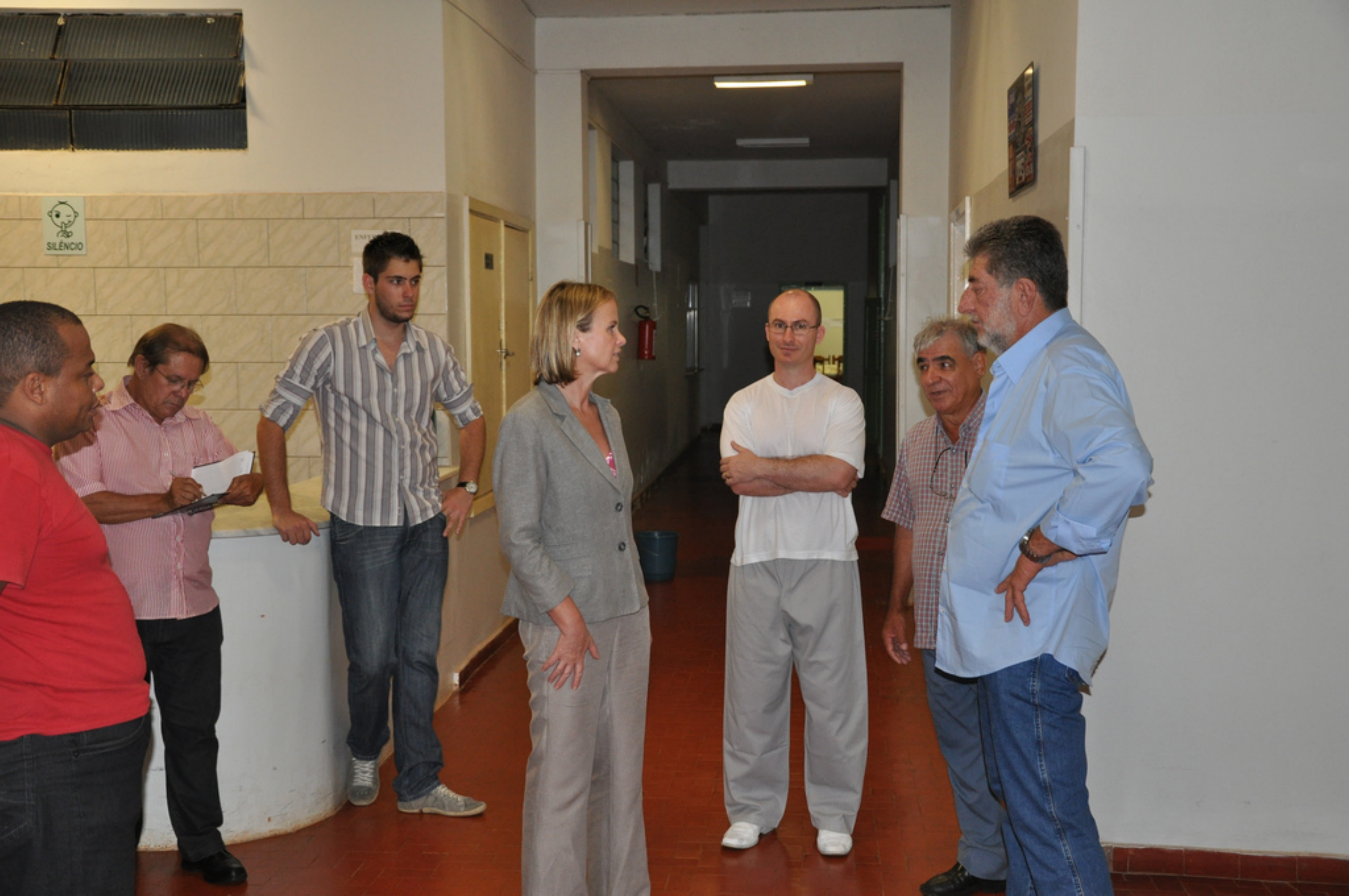} \\
			\includegraphics[height=1.5cm,width=2.25cm]{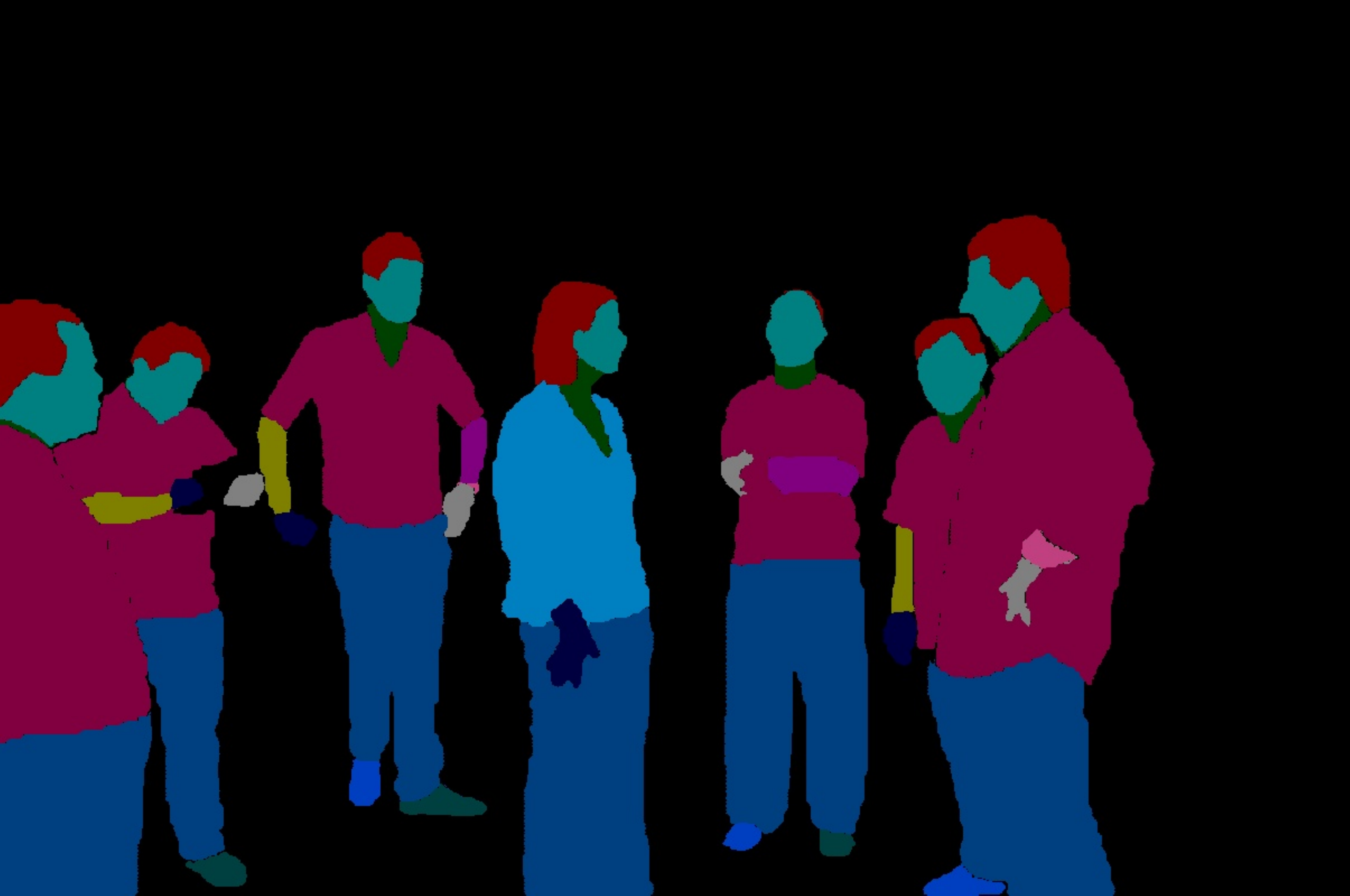} \\
			\includegraphics[height=1.5cm,width=2.25cm]{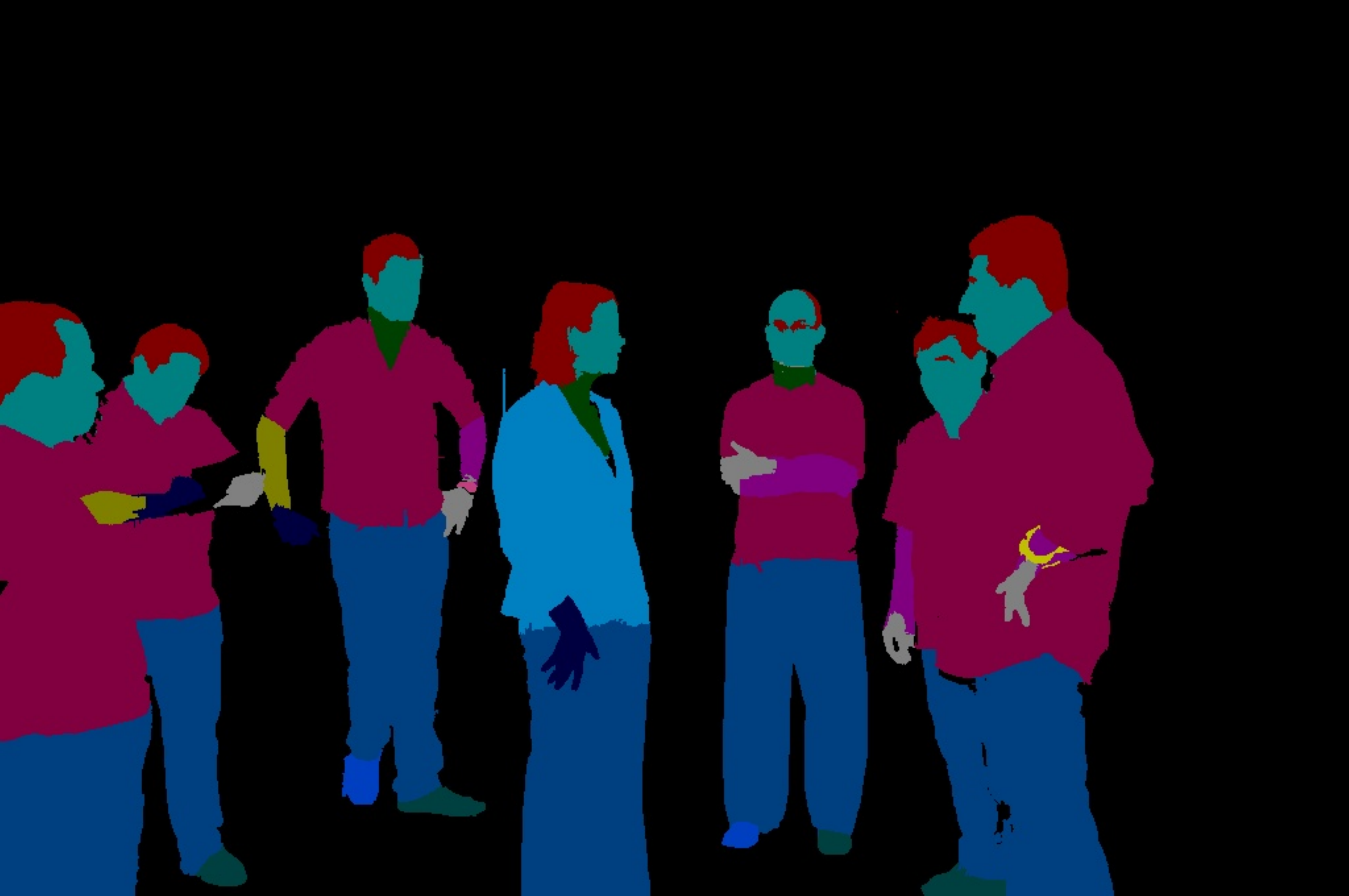} \\
			\includegraphics[height=1.5cm,width=2.25cm]{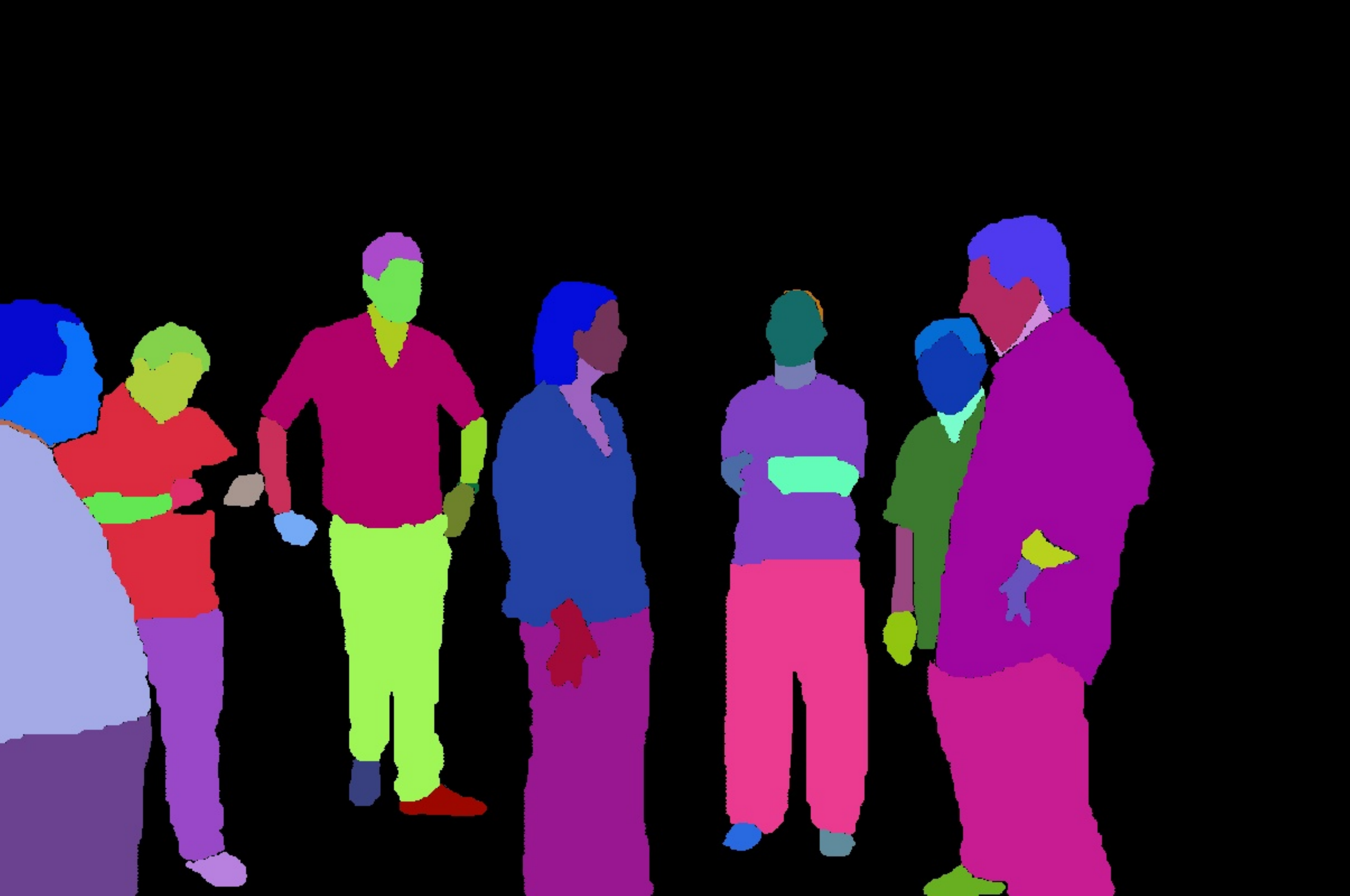} \\
			\includegraphics[height=1.5cm,width=2.25cm]{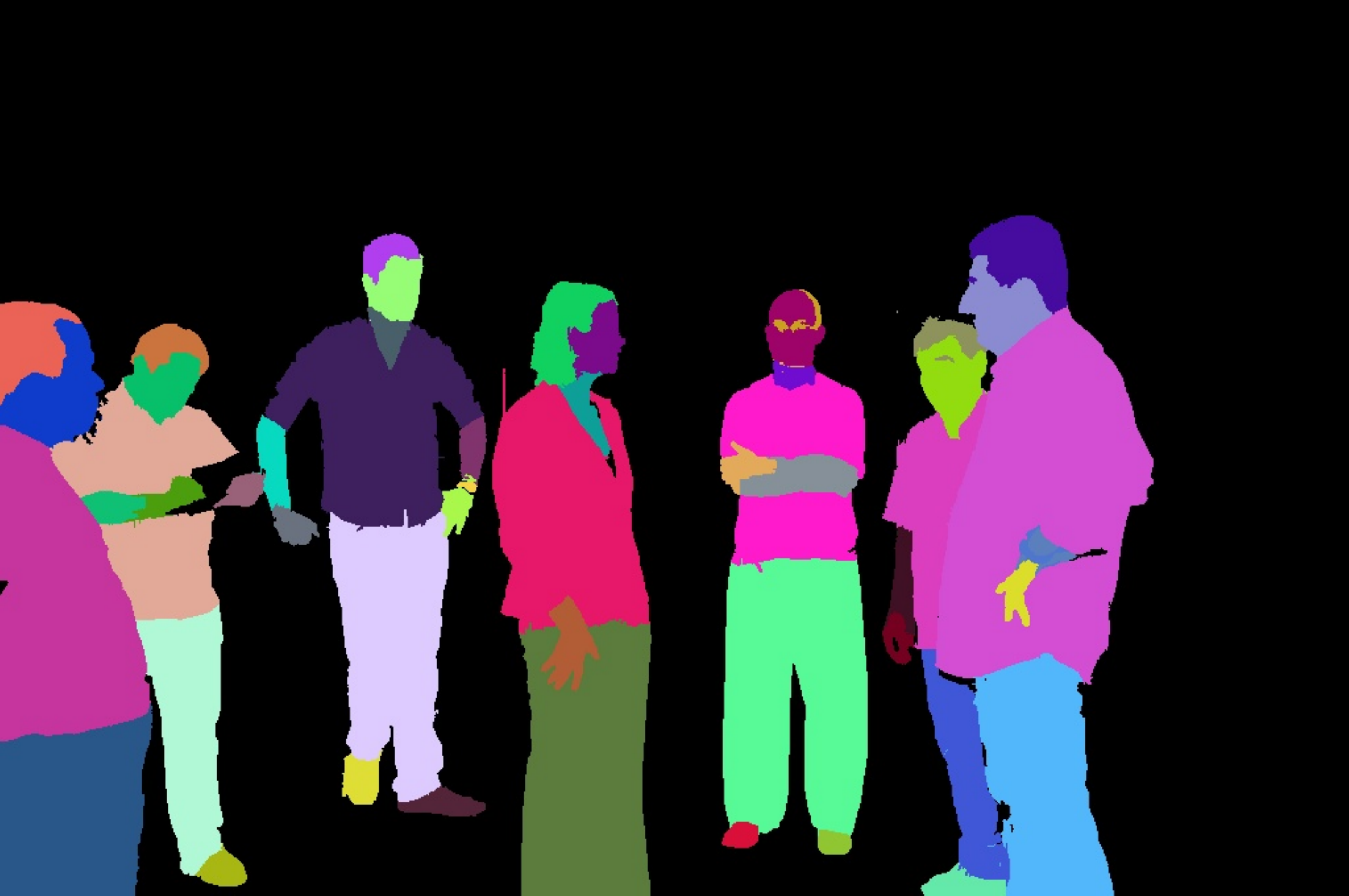} \\
			\includegraphics[height=1.5cm,width=2.25cm]{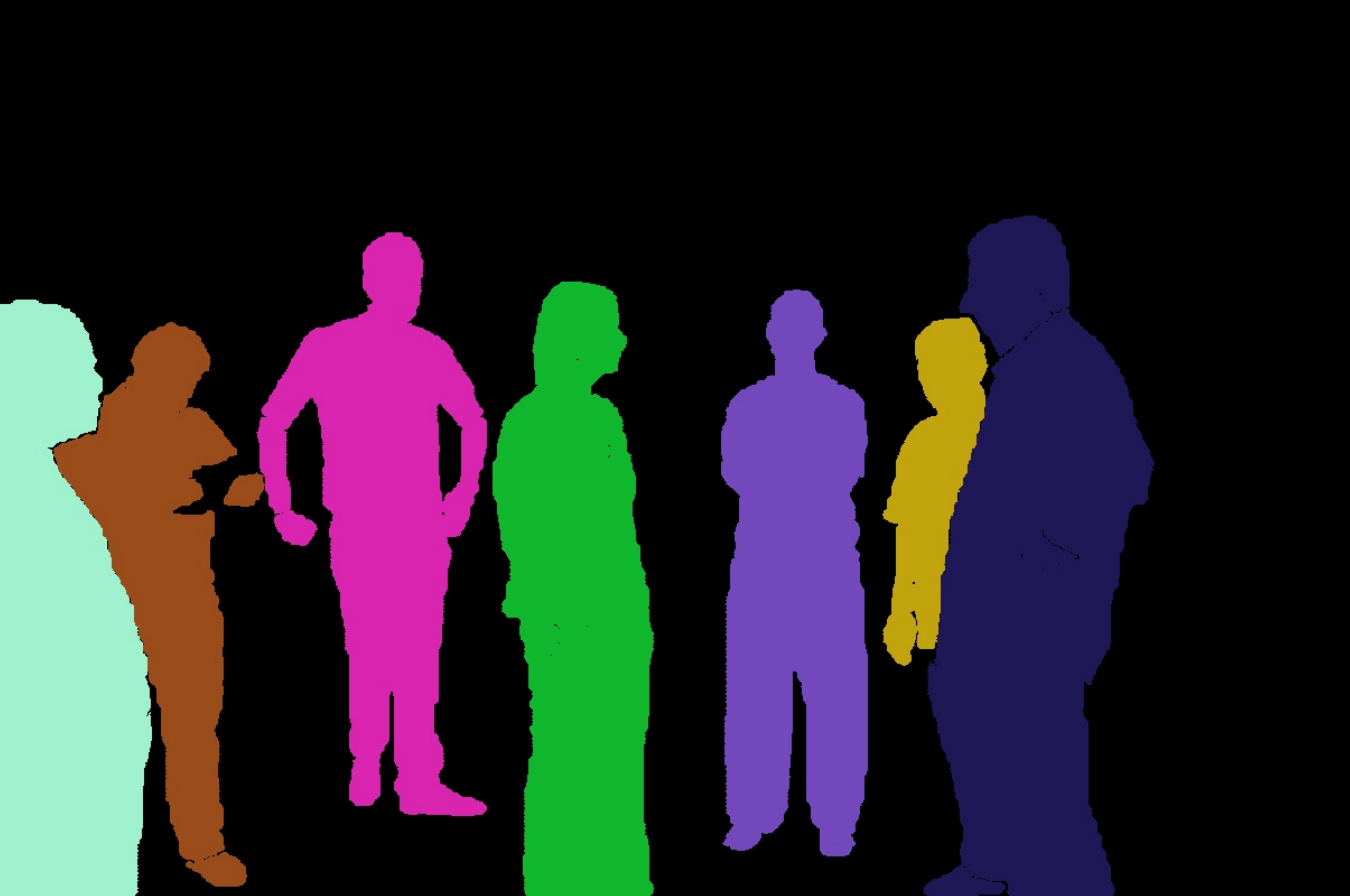} \\
			\includegraphics[height=1.5cm,width=2.25cm]{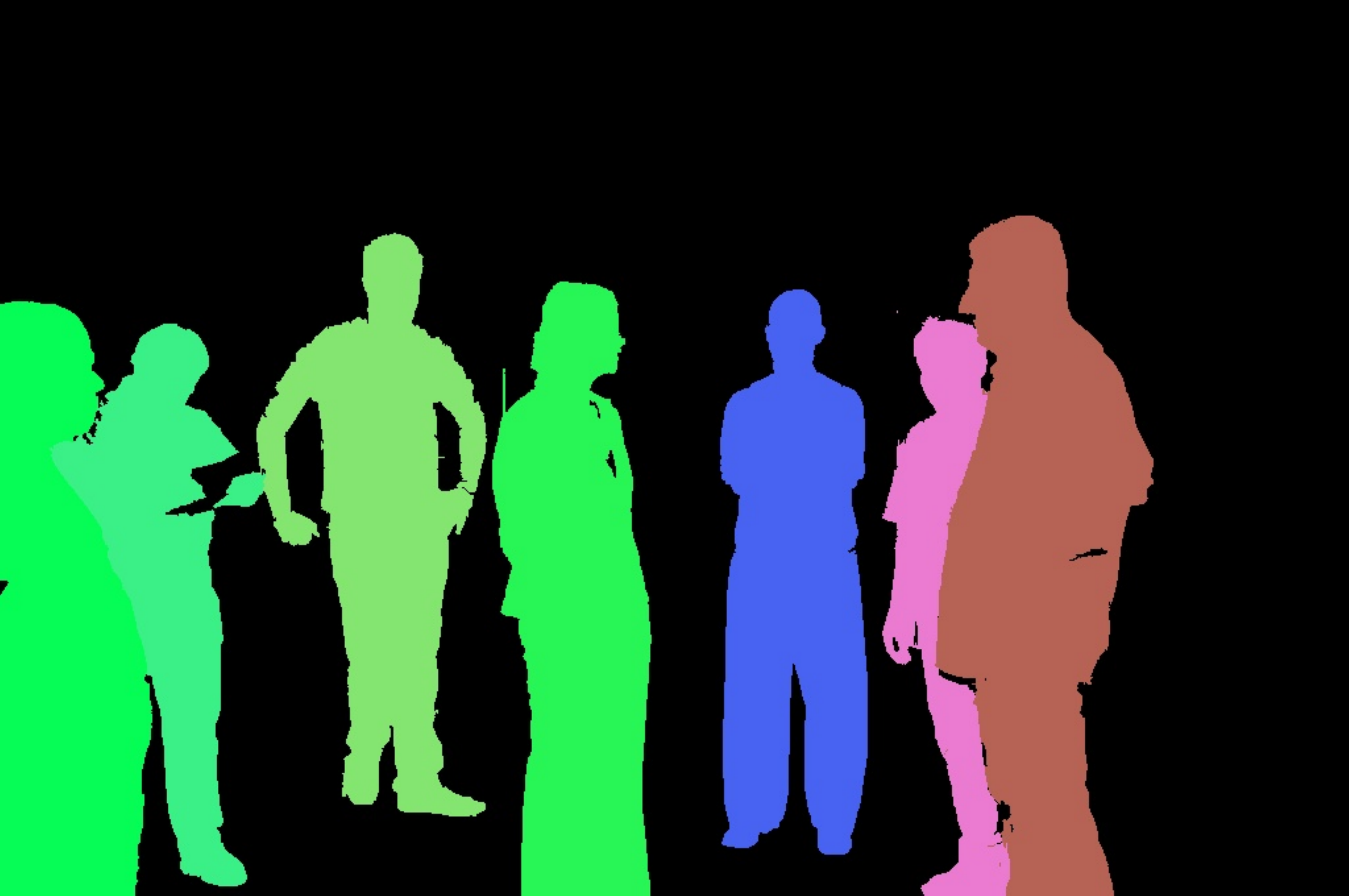} \\
		\end{minipage}
	}
	\subfigure
	{
		\begin{minipage}[b]{0.16071\linewidth}
			\centering
			\includegraphics[height=1.5cm,width=2.25cm]{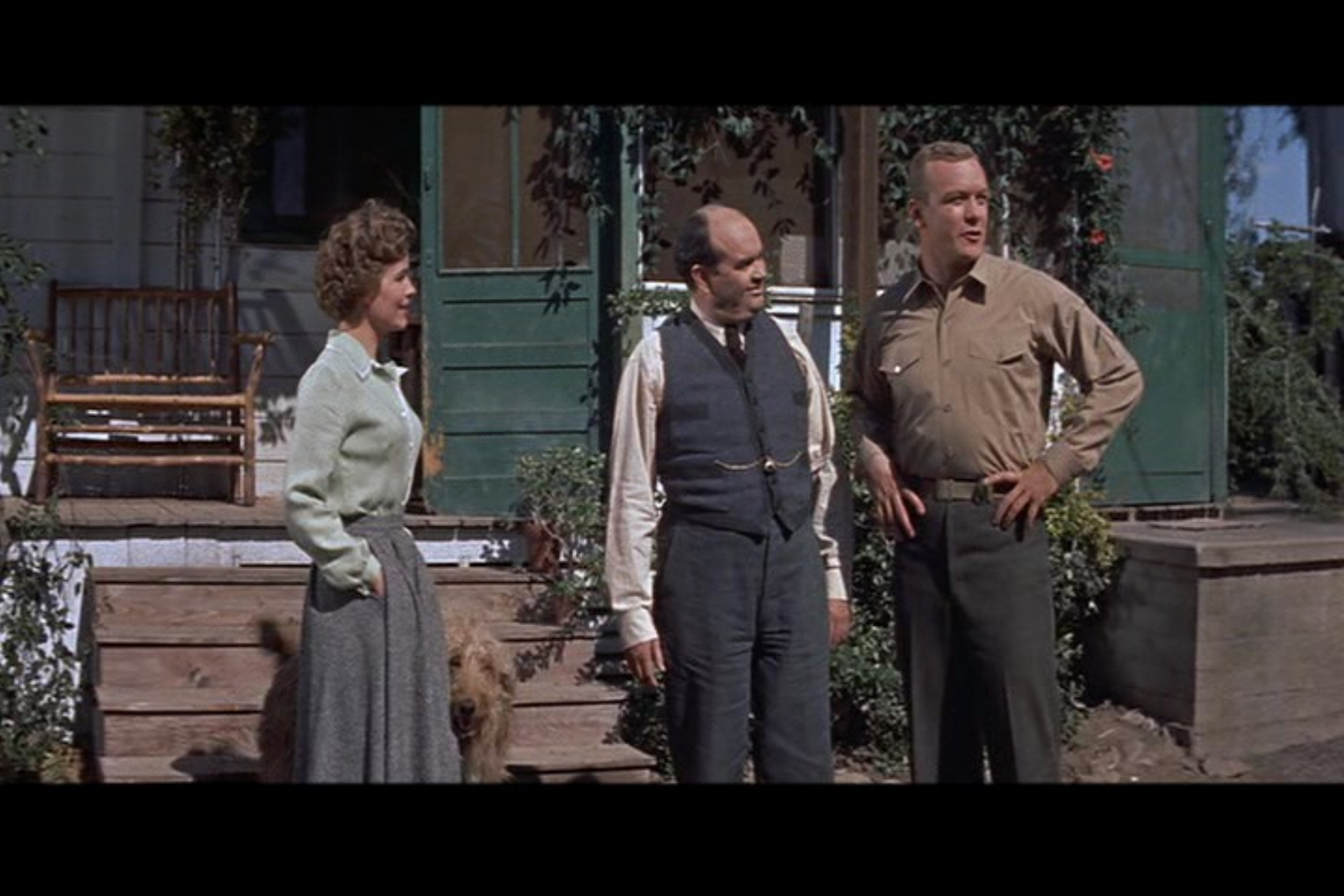} \\
			\includegraphics[height=1.5cm,width=2.25cm]{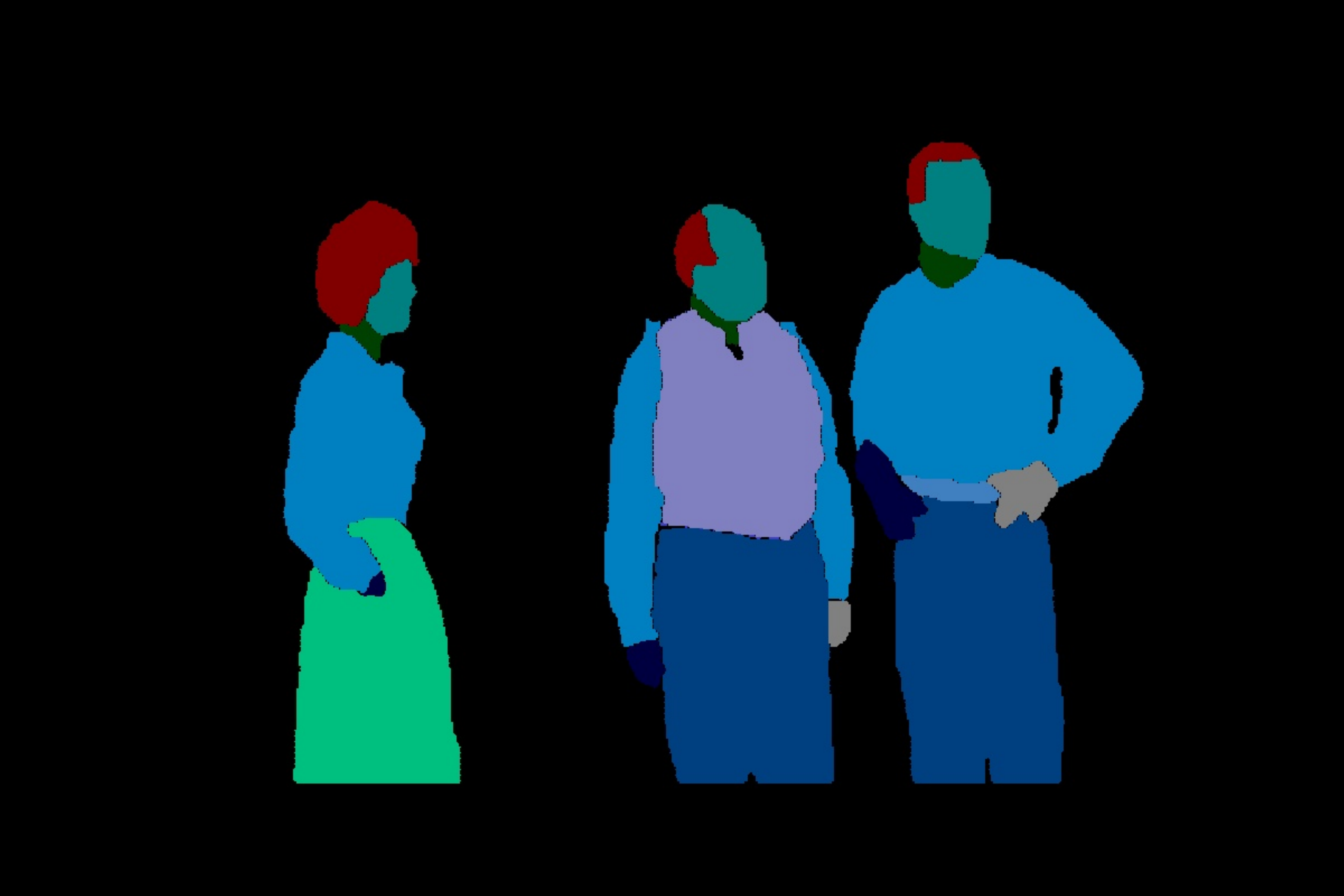} \\
			\includegraphics[height=1.5cm,width=2.25cm]{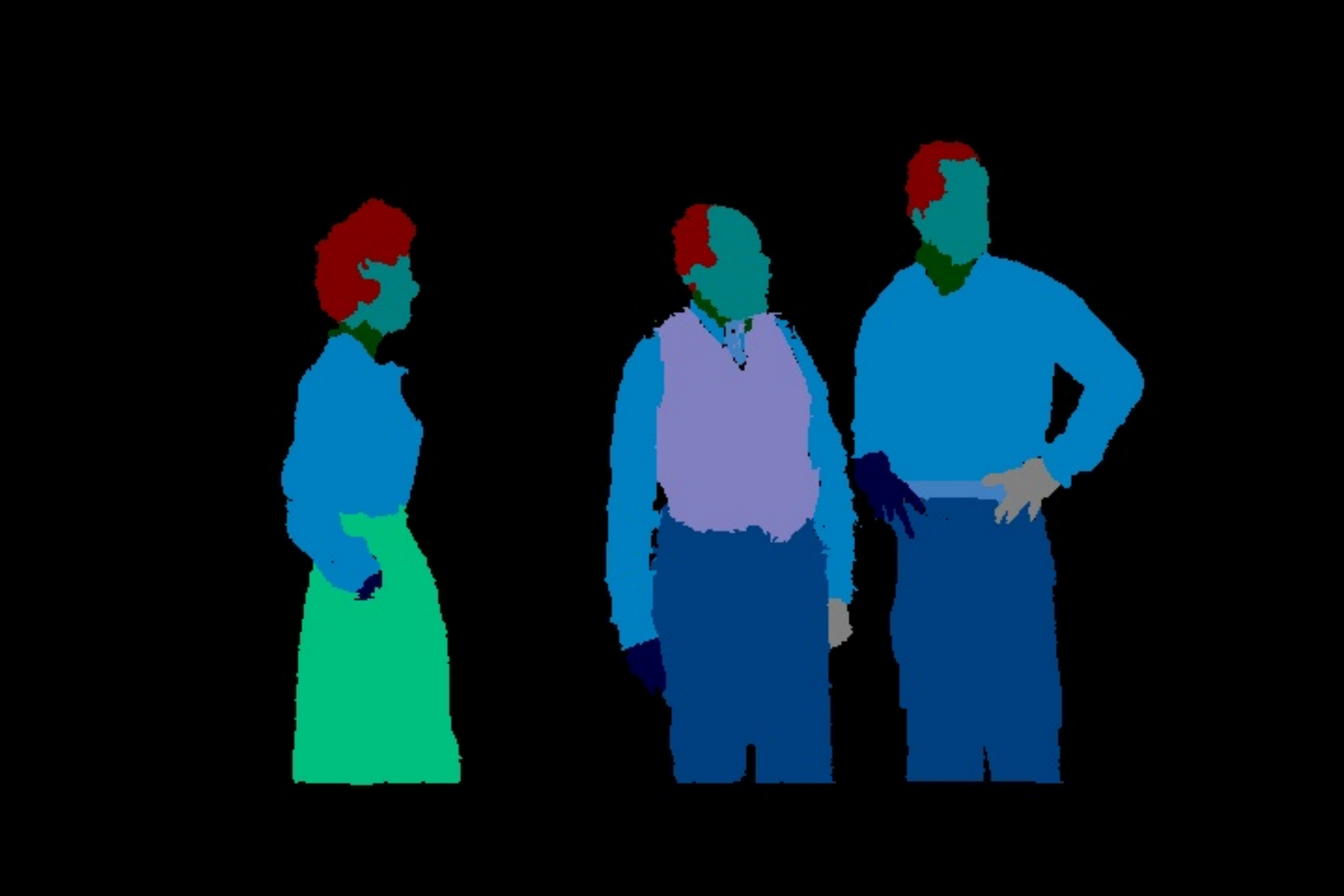} \\
			\includegraphics[height=1.5cm,width=2.25cm]{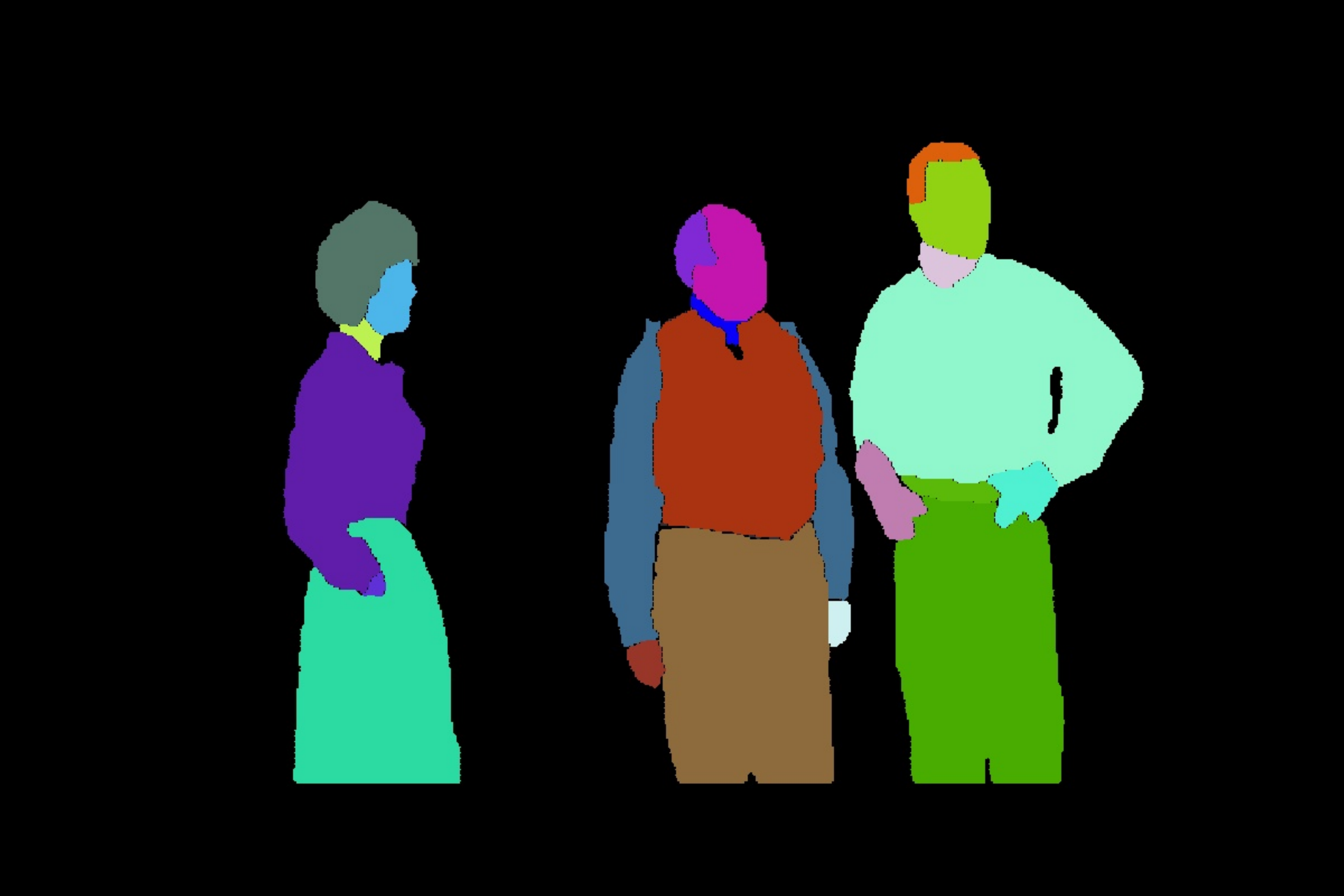} \\
			\includegraphics[height=1.5cm,width=2.25cm]{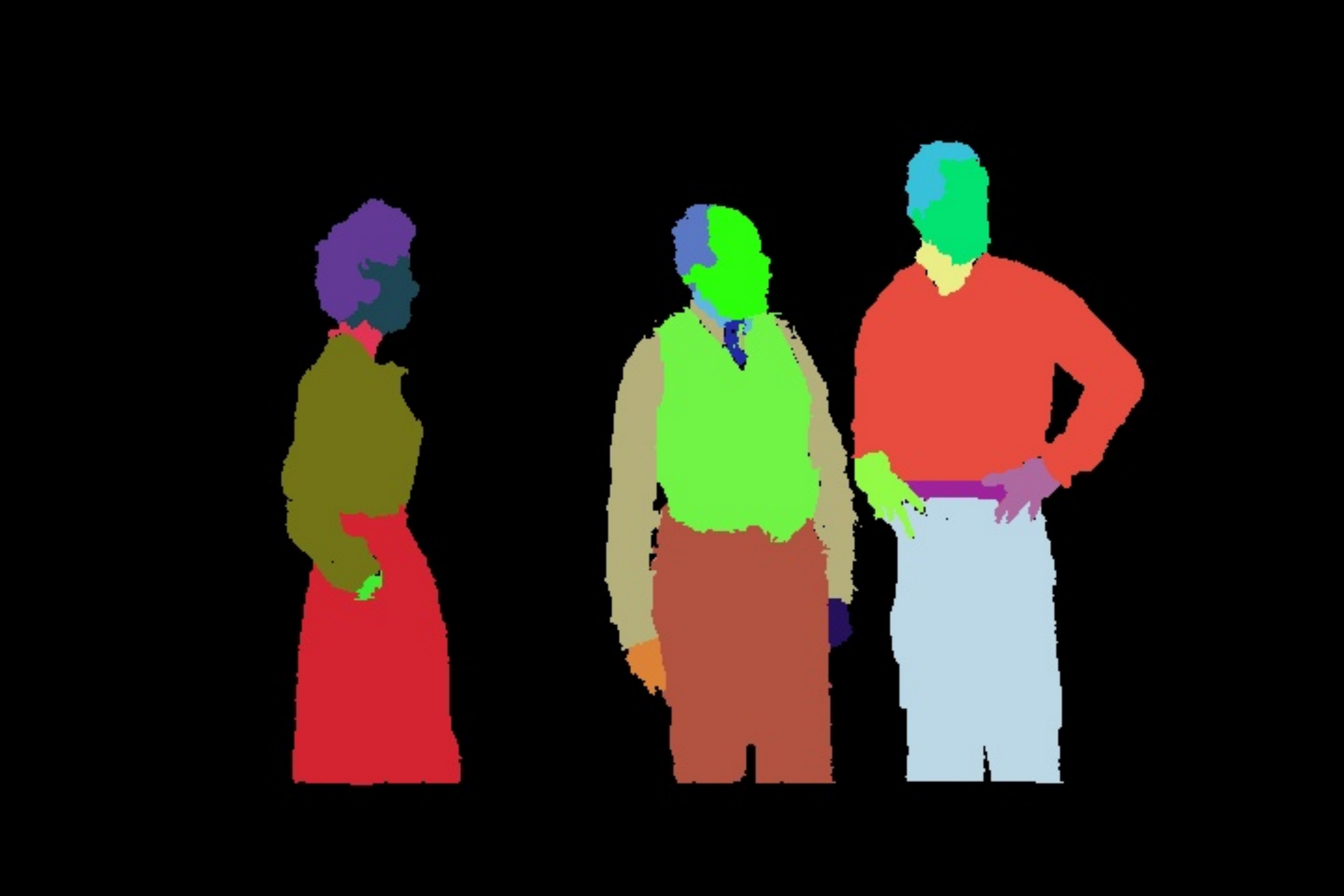} \\
			\includegraphics[height=1.5cm,width=2.25cm]{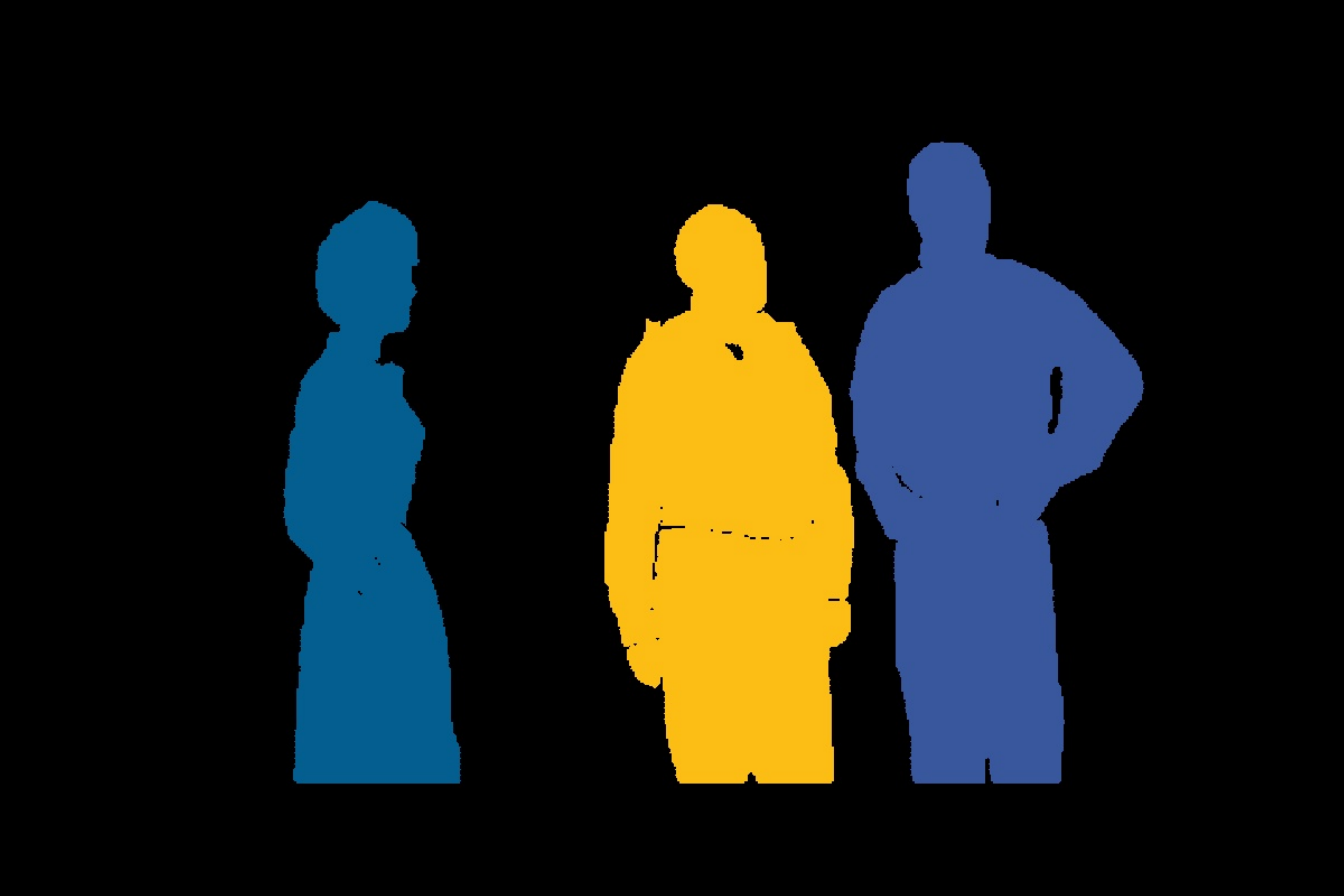} \\
			\includegraphics[height=1.5cm,width=2.25cm]{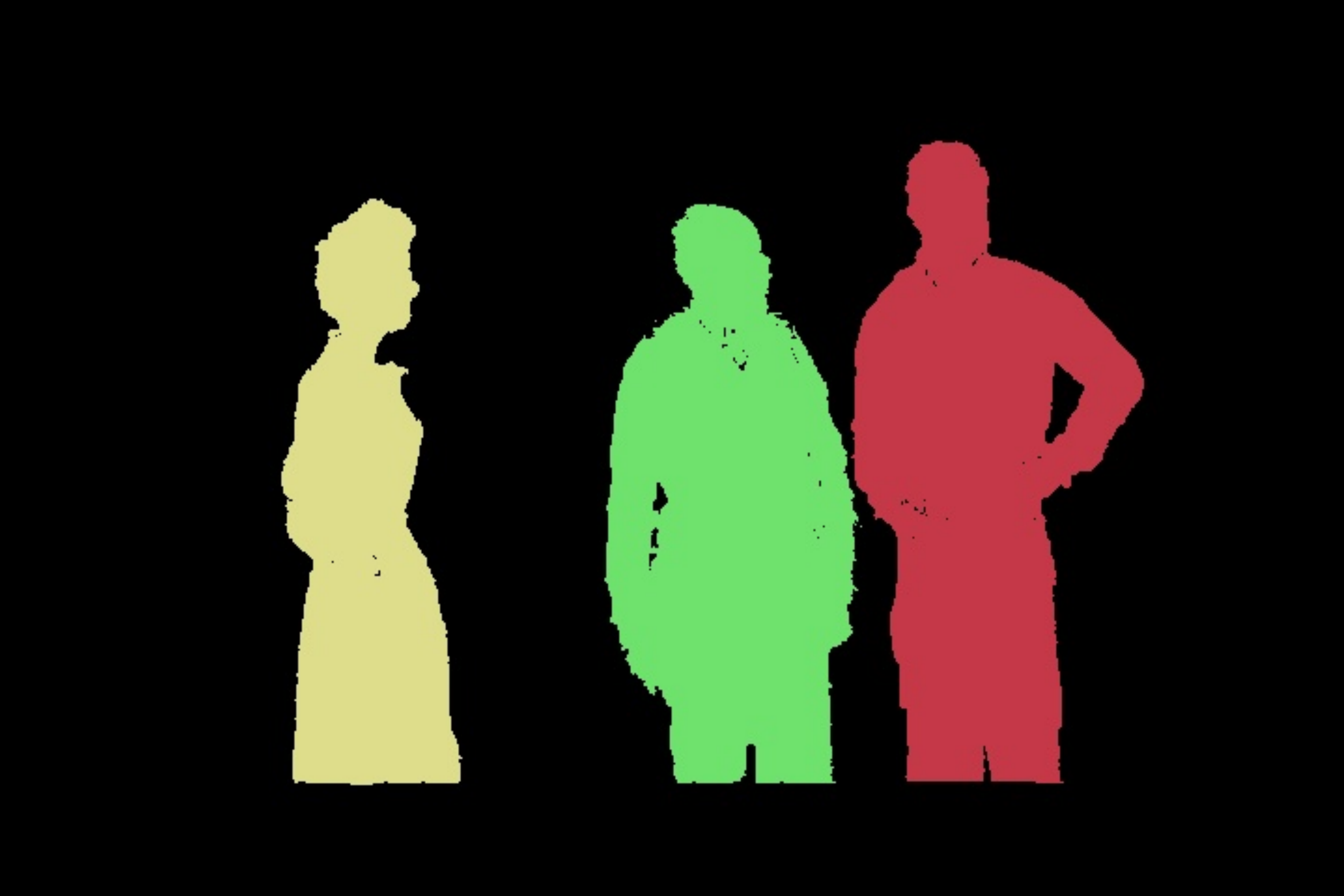} \\
		\end{minipage}
	}
	\subfigure
	{
	\begin{minipage}[b]{0.057\linewidth}
		\centering
		\includegraphics[height=1.5cm,width=1cm]{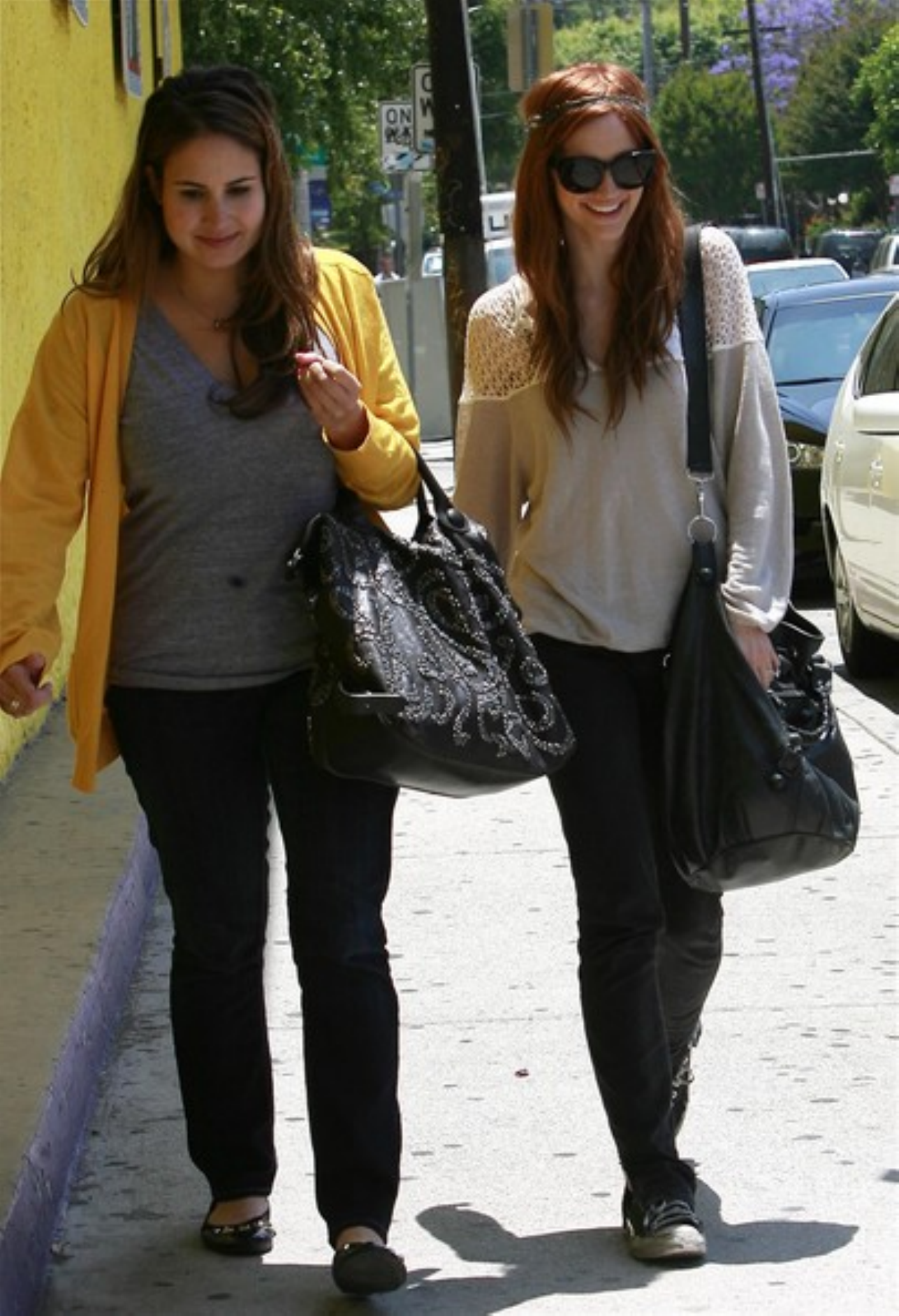} \\
		\includegraphics[height=1.5cm,width=1cm]{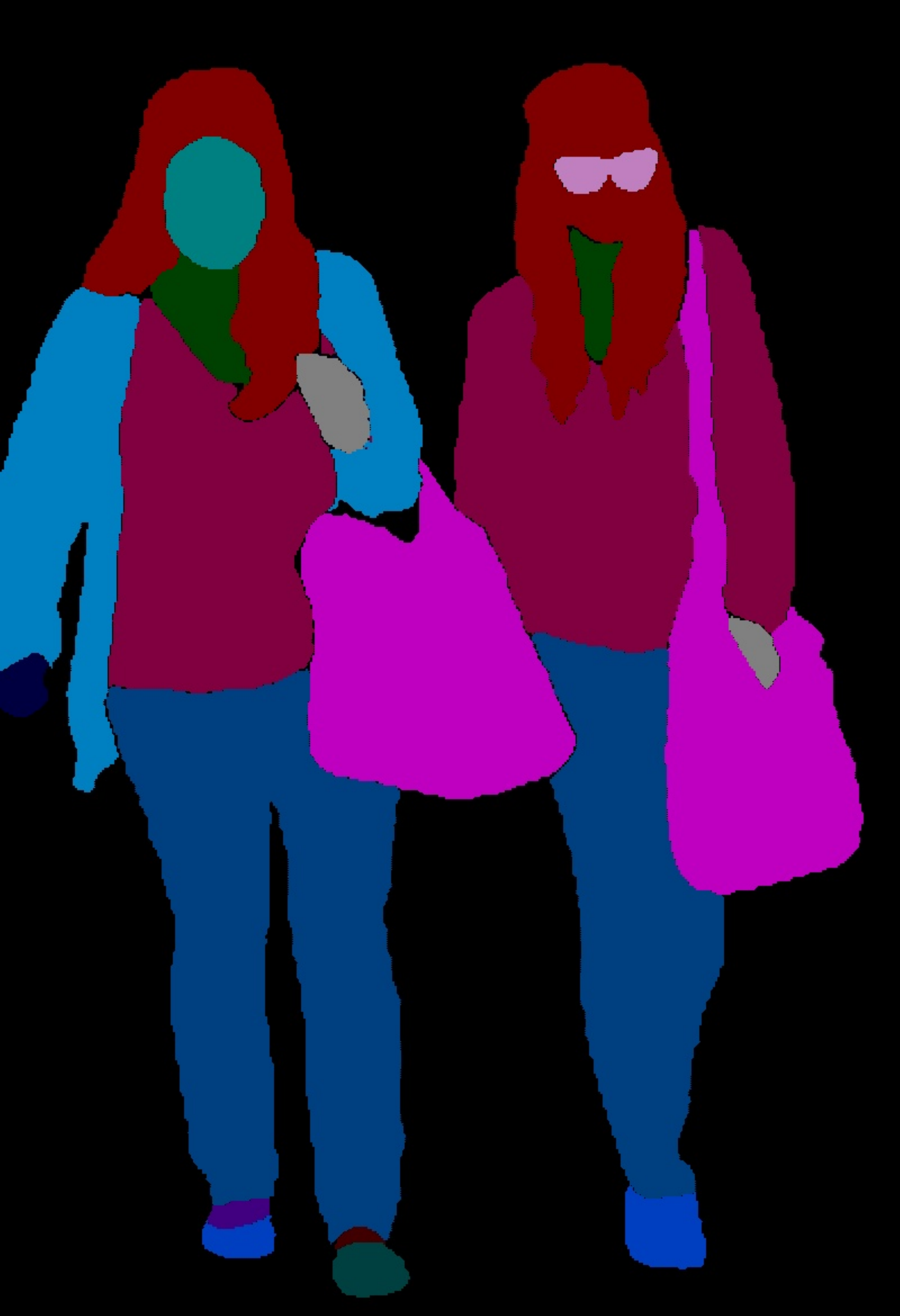} \\
		\includegraphics[height=1.5cm,width=1cm]{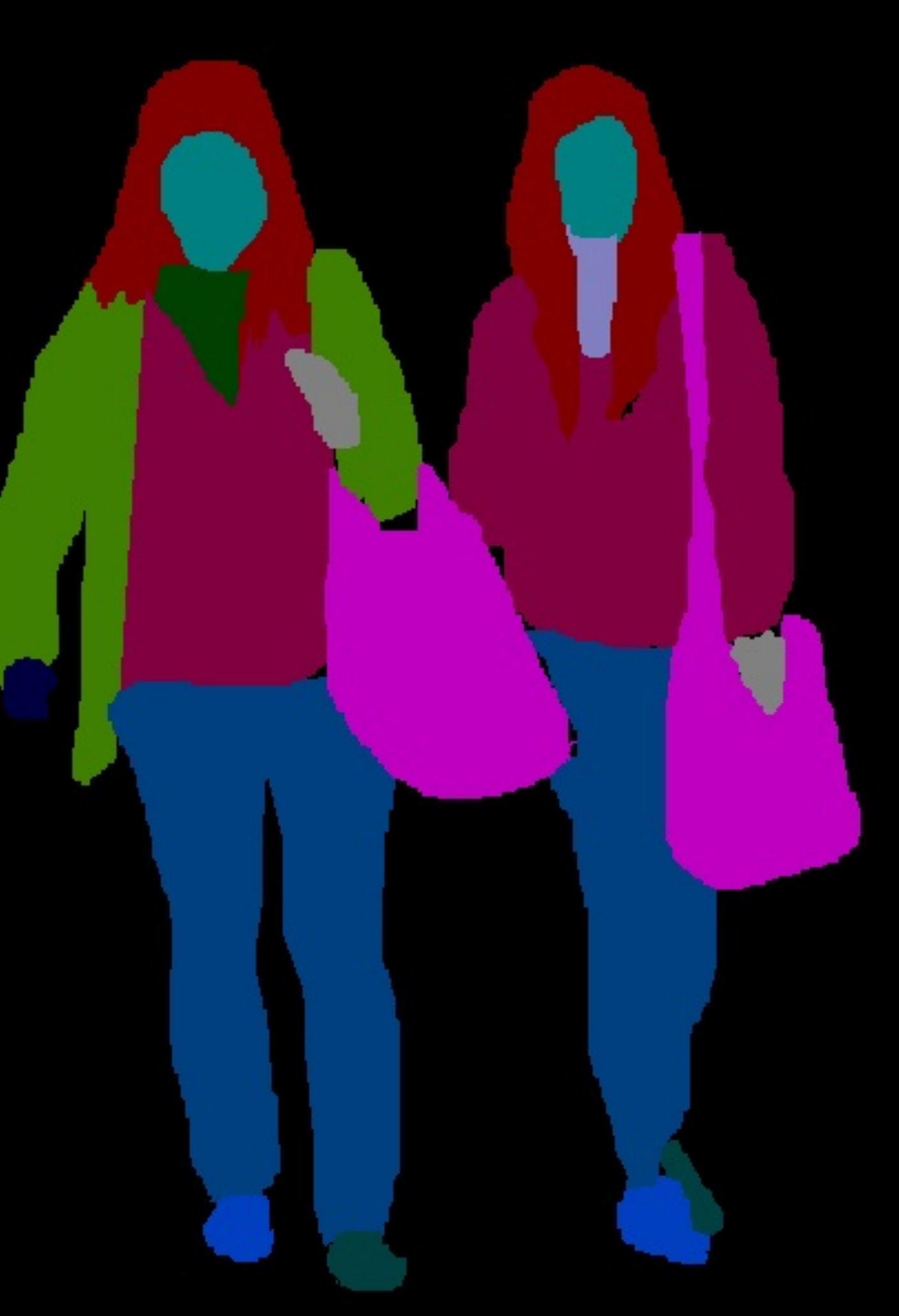} \\
		\includegraphics[height=1.5cm,width=1cm]{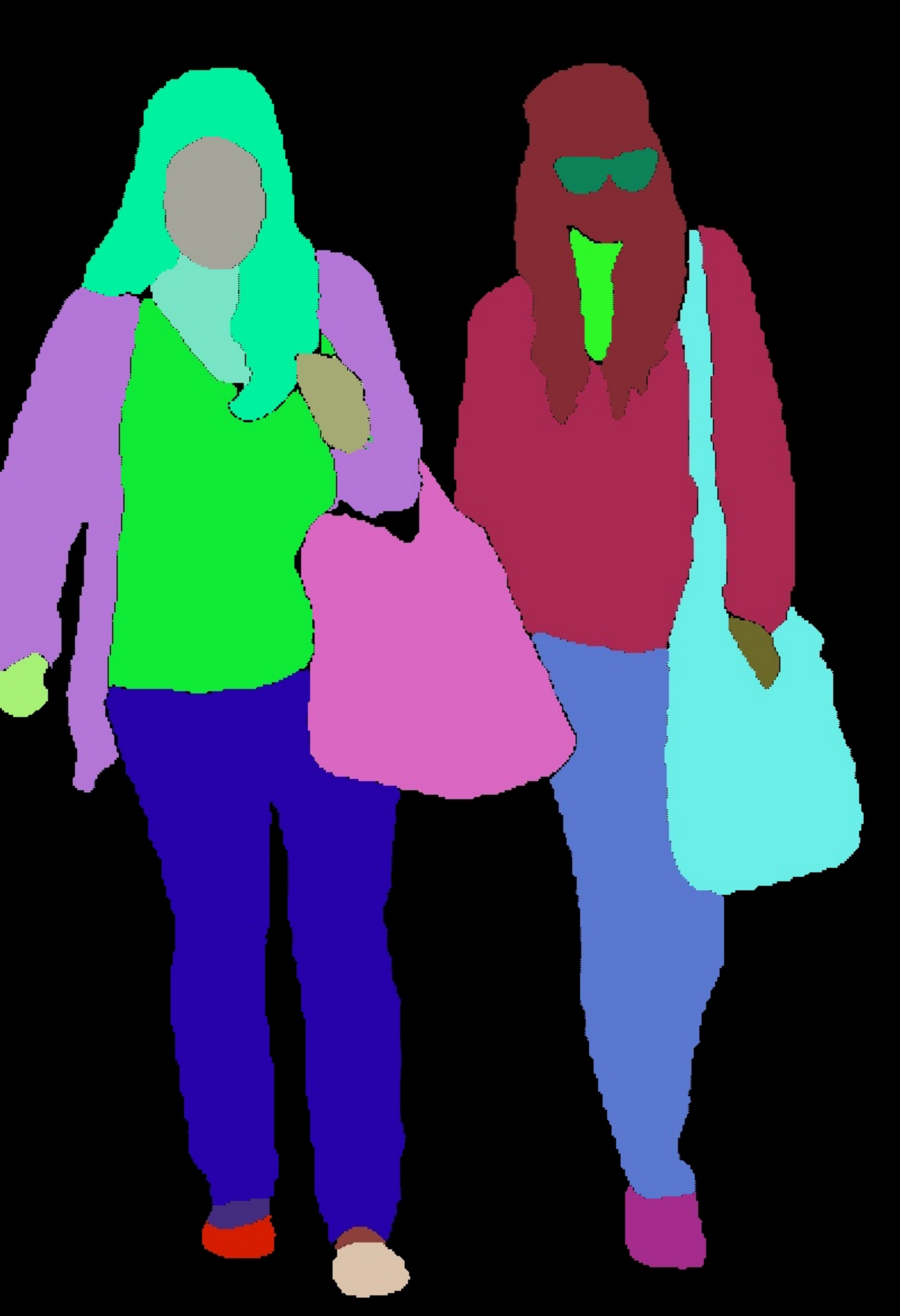} \\
		\includegraphics[height=1.5cm,width=1cm]{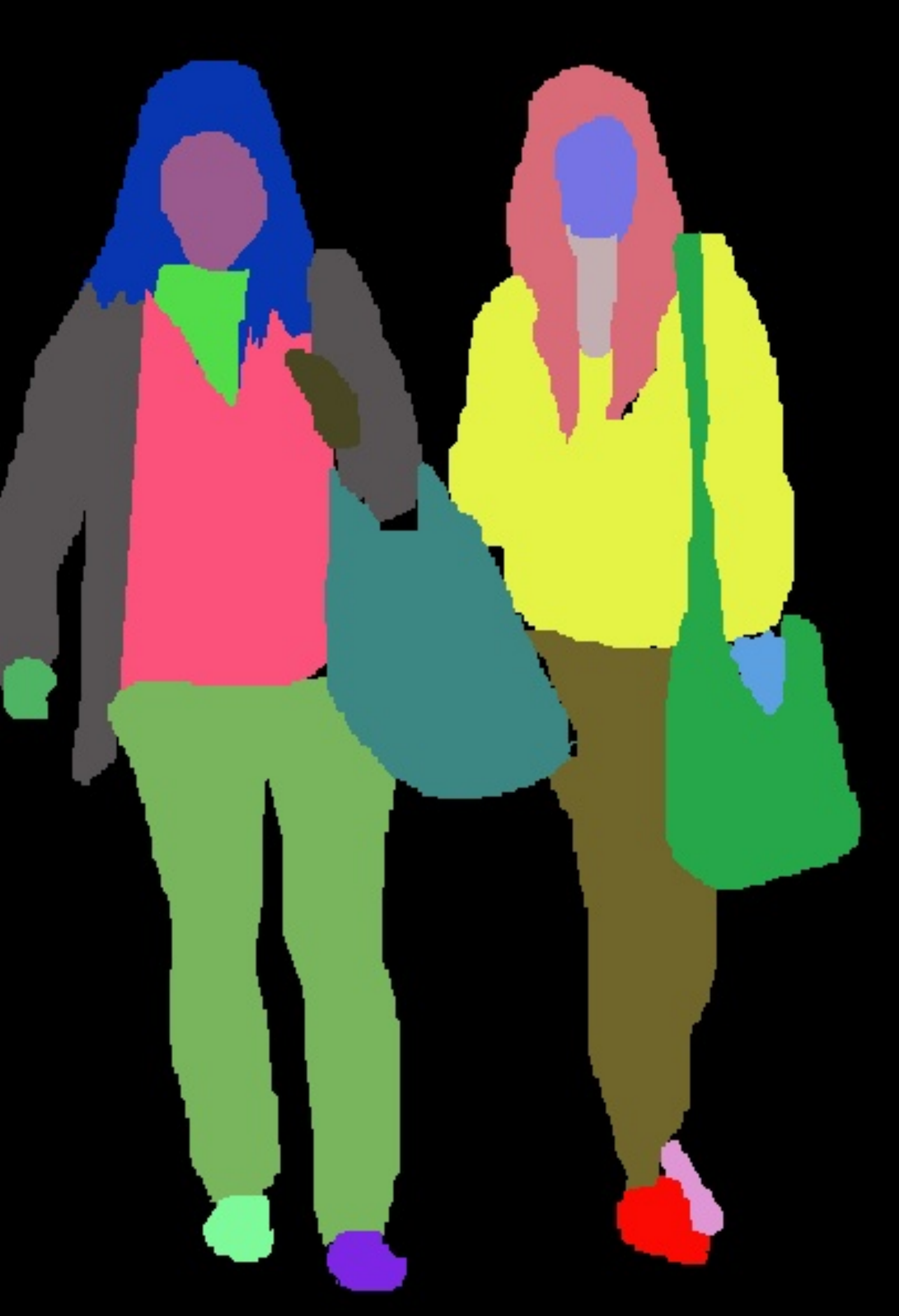} \\
		\includegraphics[height=1.5cm,width=1cm]{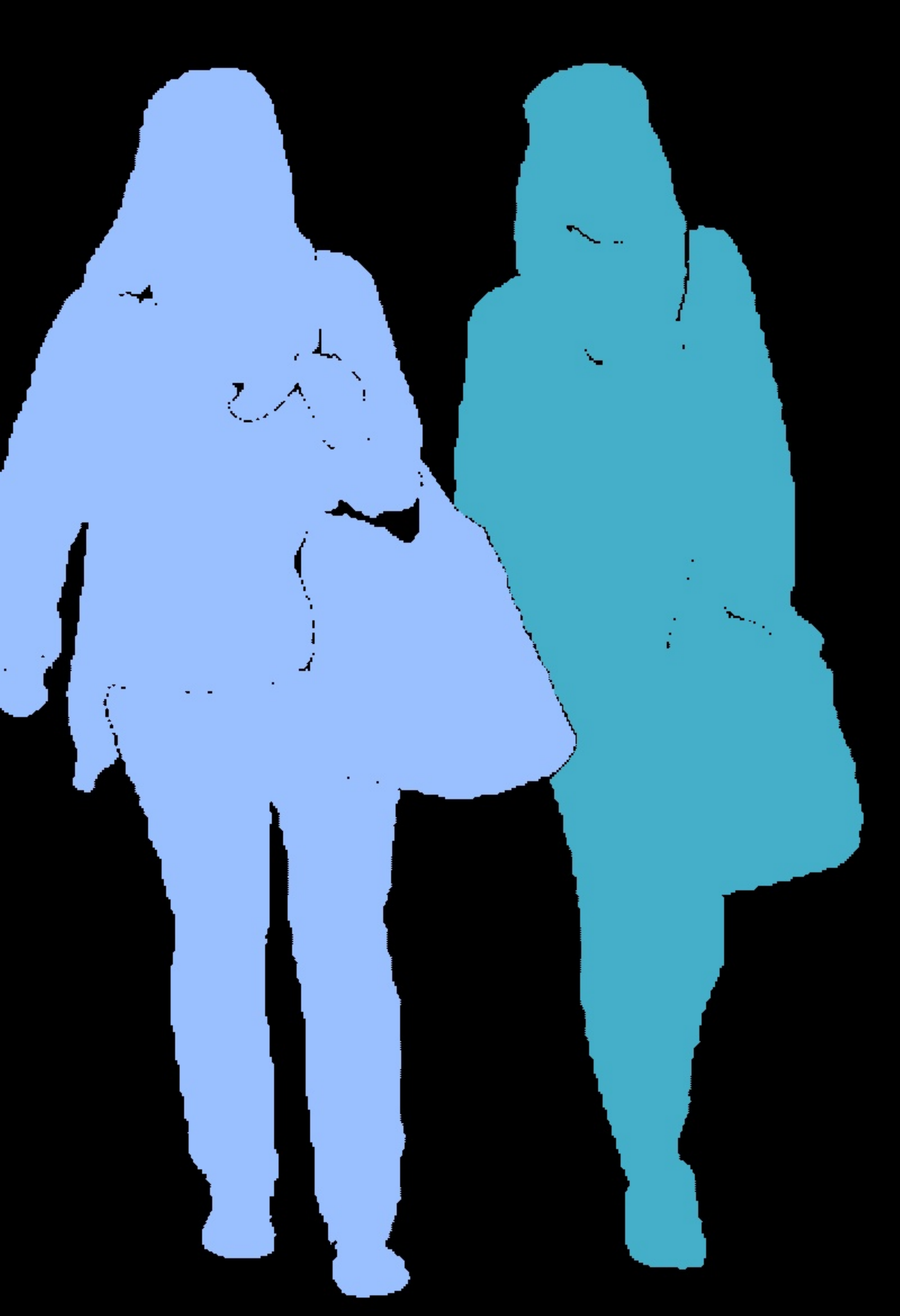} \\
		\includegraphics[height=1.5cm,width=1cm]{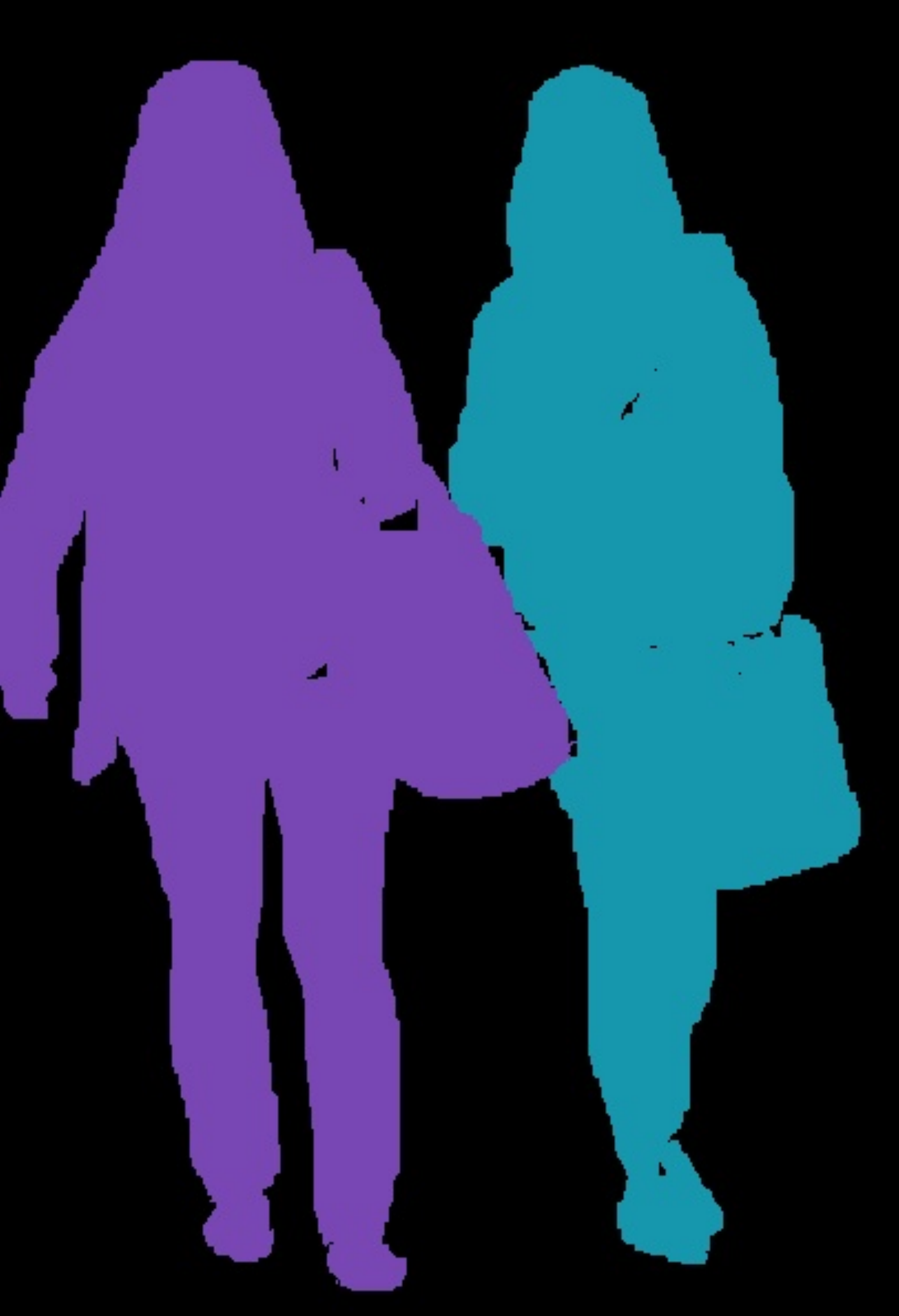} \\
	\end{minipage}
	}

	\caption{Visualization of NTHP on the MHP v2.0 dataset. Pictures on the first line are original images, pictures on the second, fourth, and sixth lines are predictions, and those on the third, fifth, and last lines are ground truths.}
	\label{figure6}
\end{figure}

We show the results on MHP v2.0 in Table \ref{tab:table6}. For a fair comparison with the best previous method RP R-CNN \cite{Yang2020a}, we use ResNet50 \cite{He2016} as the backbone. From Table 6, we can see that we outperform all state-of-the-art methods with  46.0 $AP^p_{50}$, 47.7 $AP^p_{vol}$, 45.2 $PCP_{50}$ by training for only 12 epochs. Our best results are obtained by training for 36 epochs with 51.1 $AP^p_{50}$, 49.5 $AP^p_{vol}$, 49.9 $PCP_{50}$, with a margin of 5.8 points $AP^p_{50}$, 2.7 points $AP^p_{vol}$, and 6.1 points $AP^p_{50}$ over the best previous entry. Note that we do not use flipping operation in training or any test-time augmentation on the MHP v2.0 dataset. We visualize good or bad results in Figure \ref{figure6}. We contrast the part-level and human-level instance segmentation results. And to demonstrate our classification performance, we use the same color to represent instances with the same category in the second and third lines.

\begin{table}[htbp]
	\caption{Multi-human parsing results on the MHP v2.0 val set. $\rm ^\ast$denotes longer learning schedule. }
	\begin{center}
		\begin{tabular}{c  c  c c c}
			\hline
			Mehods              & Epochs  &$AP^p_{50}$ & $AP^p_{vol}$ & $PCP_{50}$ \\\hline
			MH-Parser \cite{Li2017}        &    -    &    17.9   &       36.0  & 26.9 \\
			Parsing R-CNN \cite{Yang2019a}      &     75  &       24.5   &     39.5      & 37.2 \\
			NAN \cite{Zhao2018}               &$\sim$80 &      25.1 &      41.7  & 32.2 \\
			M-CE2P \cite{Ruan2018}            &    150  &       34.5  &    42.7        &43.8  \\
			SemaTree \cite{Ji2019}       &     200 &       34.4    &   42.5         & 43.5 \\
			RP R-CNN \cite{Yang2020a}        &   150   &  45.3    &   46.8          & 43.8 \\
			NTHP (ours)           &     12  &    46.0       &     47.7        & 45.2 \\
			$\rm NTHP^\ast$ (ours)            &   36    &   \textbf{51.1}  &  \textbf{49.5}& \textbf{49.9} \\
			\hline
		\end{tabular}
		\label{tab:table6}
	\end{center}
\end{table}

\begin{table}[htbp]
	\caption{Multi-human parsing results on the PASCAL-Person-Part test dataset. $\rm ^\#$denotes using pretraining on other datasets. $\rm ^\dagger$denotes test-time augmentation. }
	\begin{center}
		\begin{tabular}{c c c r r r}
			\hline
			\multirow{2}*{Methods} & \multirow{2}*{Epochs} & \multirow{2}*{$AP^r_{vol}$} & \multicolumn{3}{c} {IoU threshold}\\\cline{4-6}
			&           &      &  0.5& 0.6& 0.7  \\\hline
			$\rm MNC^\#$\cite{Dai2016}     &   $\sim$117        &    36.7     &       38.8  & 28.1 & 19.3 \\
			$\rm Holistic^\#$\cite{Li2017a}    &$\sim$100 &     38.4   &      40.6  & 30.4& 19.1\\
			$\rm PGN^{\#\dagger}$\cite{Gong2018}     &    $\sim$80            &       39.2   &    39.6  &29.9 &20.0 \\
			$\rm Unified^{\#\dagger}$\cite{Qin2020}      &     $\sim$600            &      43.1   &   48.1    &38.3 &25.7\\
			NTHP (ours)            &     54              &   43.9     &     49.1  & 40.0 &28.1\\
			$\rm NTHP^\#$(ours)            &  54                 &\textbf{47.1}&  \textbf{53.9}& \textbf{44.7}& \textbf{31.9} \\\hline
		\end{tabular}
		\label{tab:table7}
	\end{center}
\end{table}

\begin{figure}[htbp]
	\centering
	\subfigure
	{
		\begin{minipage}[b]{0.14286\linewidth}
			\centering
			\includegraphics[height=1.5cm,width=2.0cm]{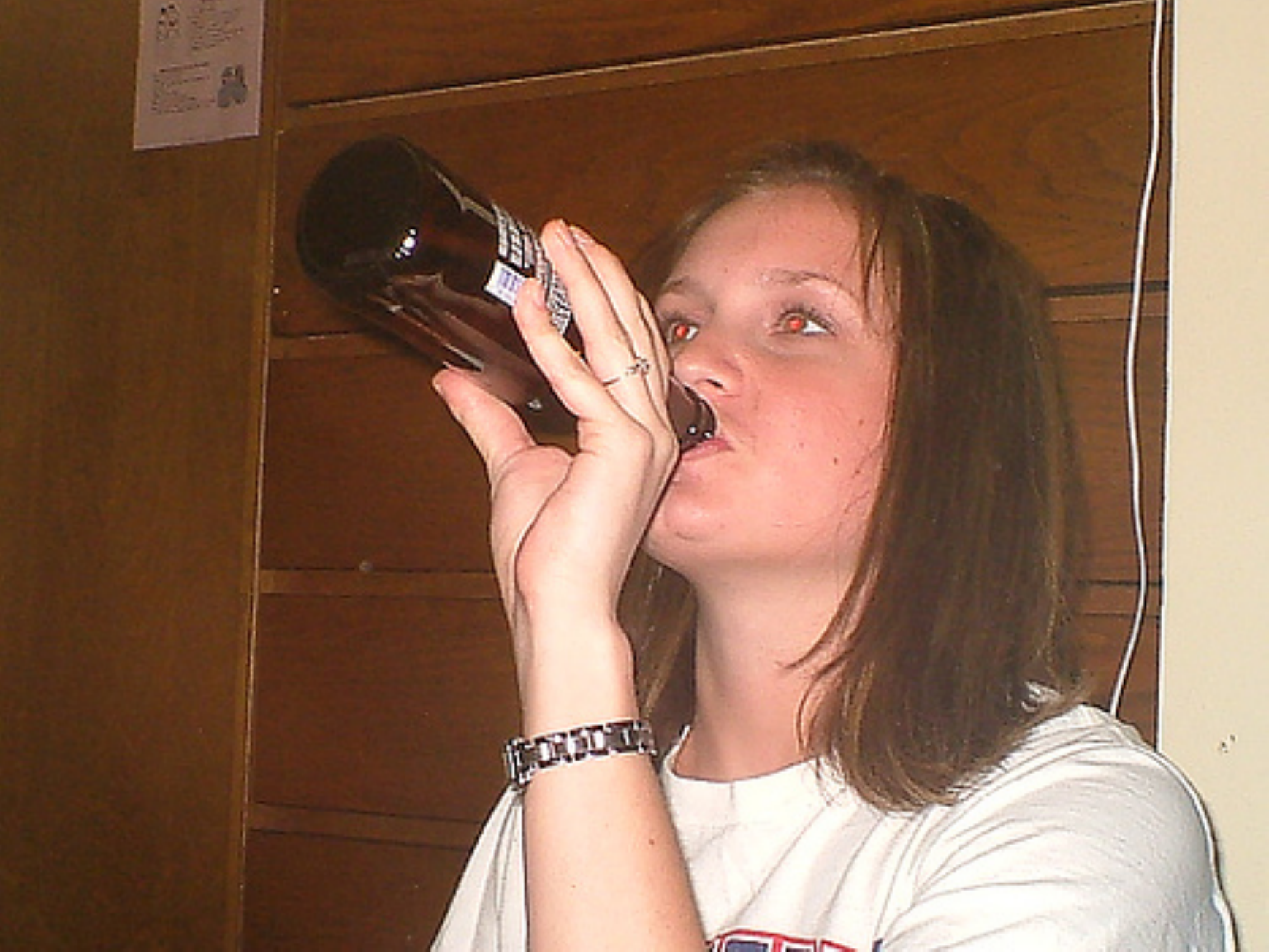} \\
			\includegraphics[height=1.5cm,width=2.0cm]{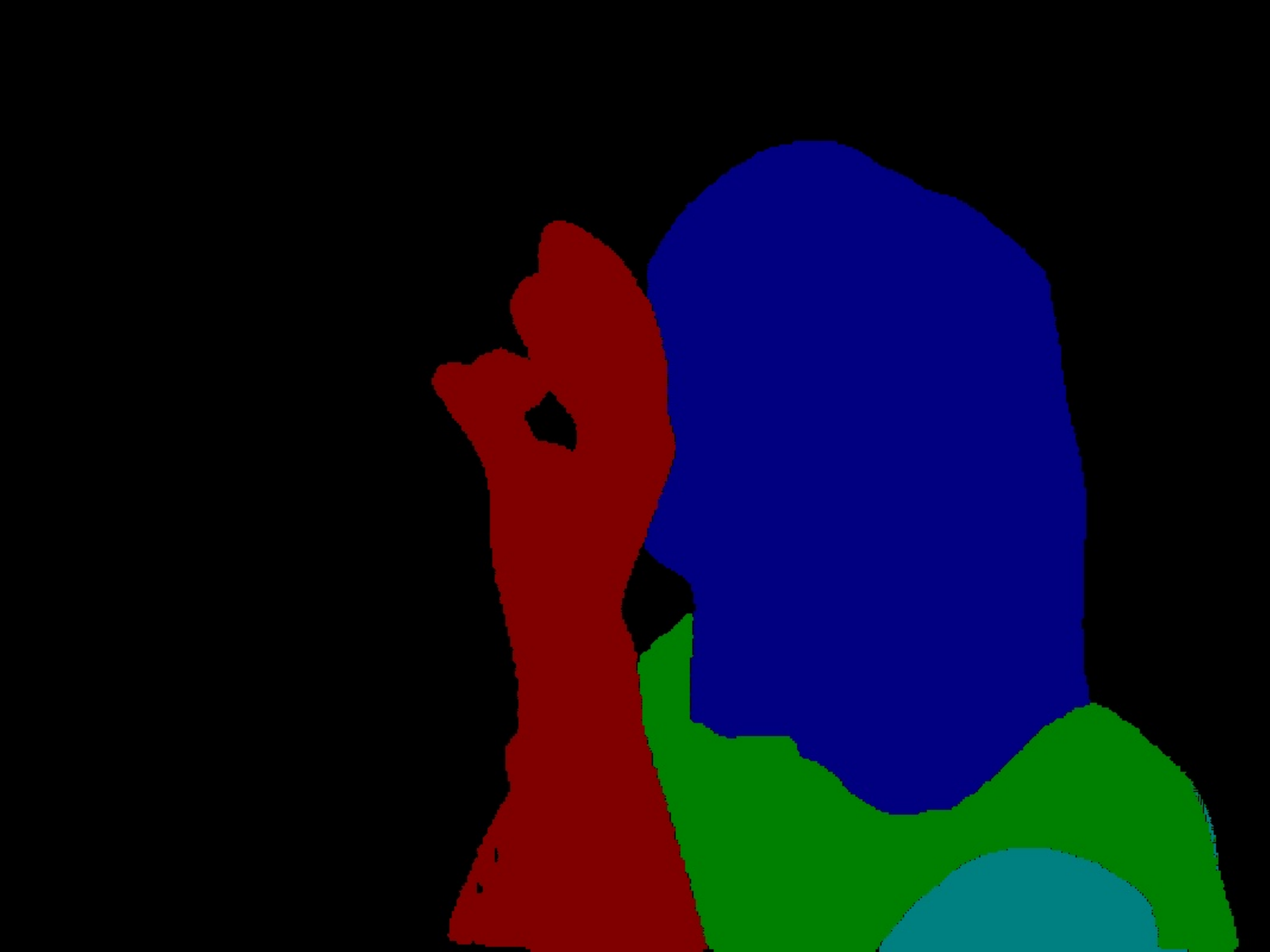} \\
			\includegraphics[height=1.5cm,width=2.0cm]{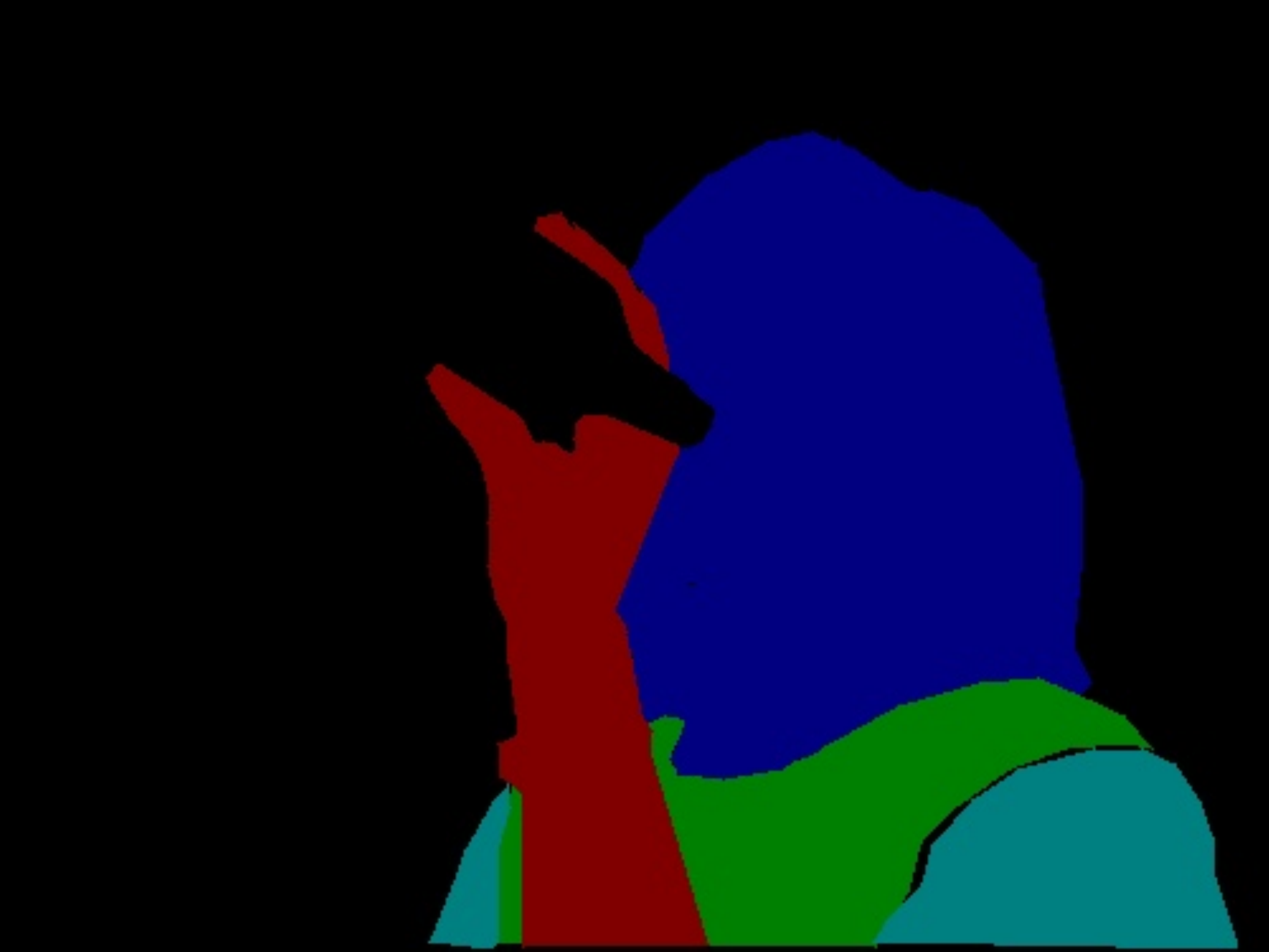} \\
			\includegraphics[height=1.5cm,width=2.0cm]{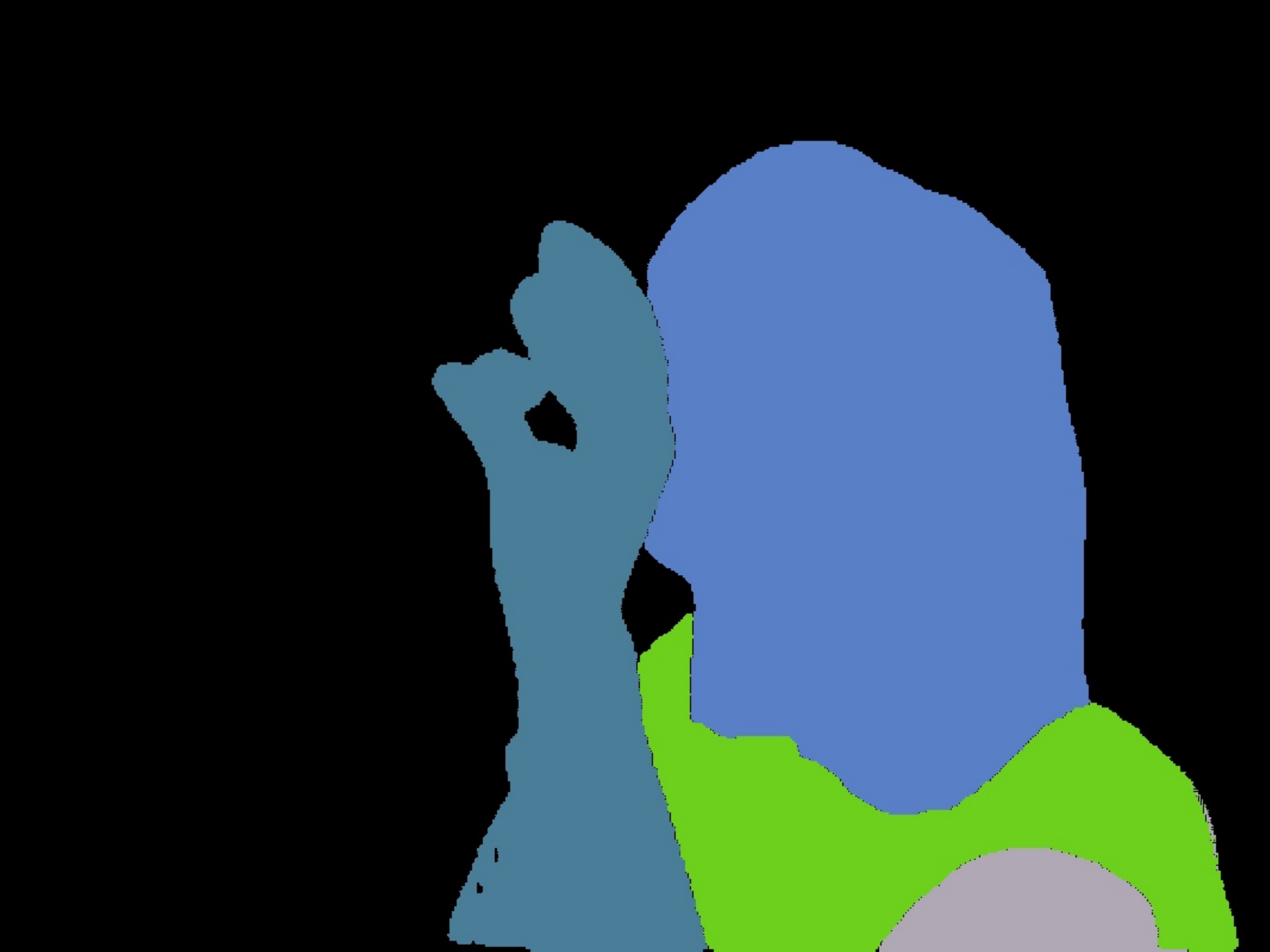} \\
			\includegraphics[height=1.5cm,width=2.0cm]{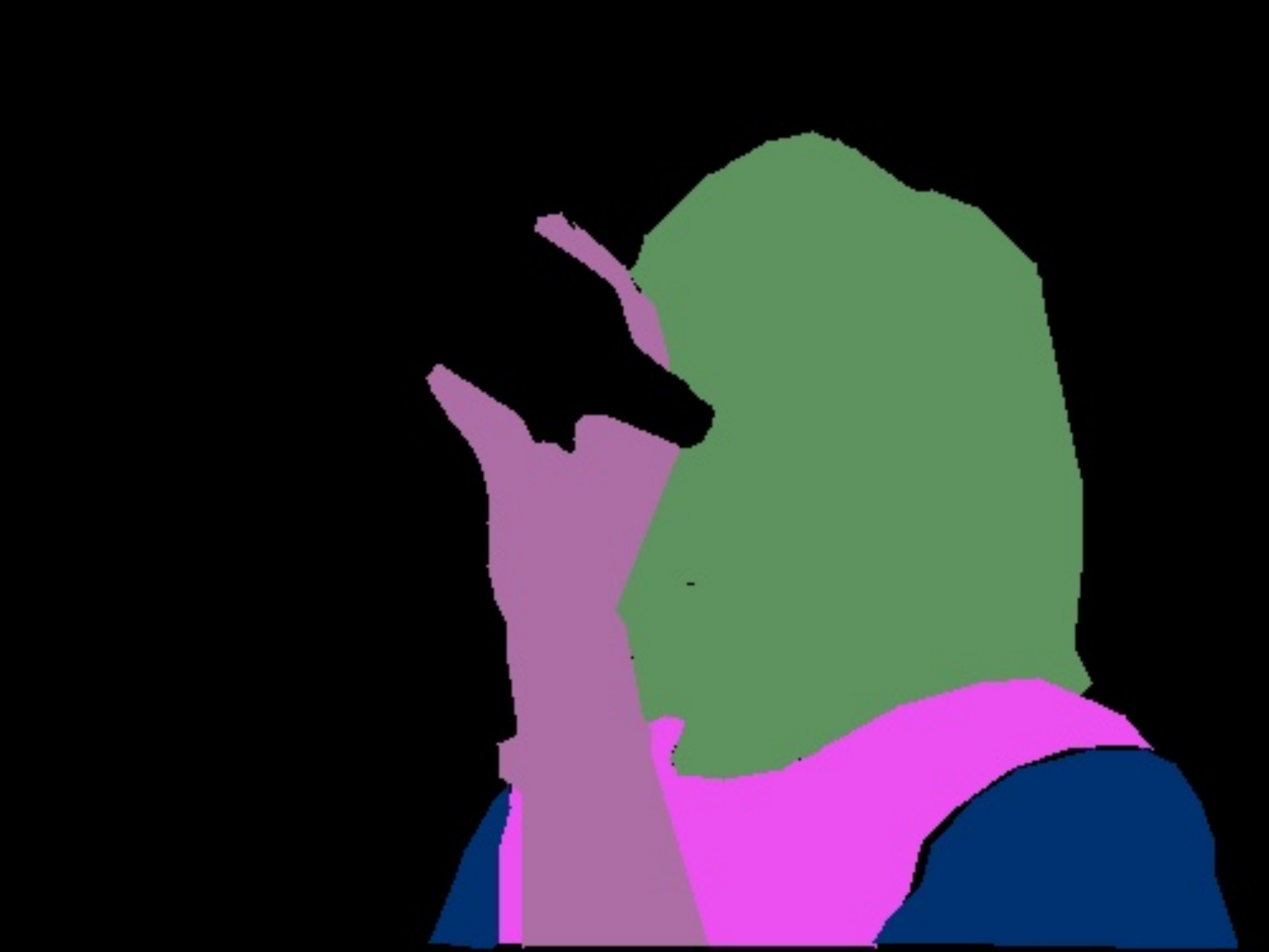} \\
			\includegraphics[height=1.5cm,width=2.0cm]{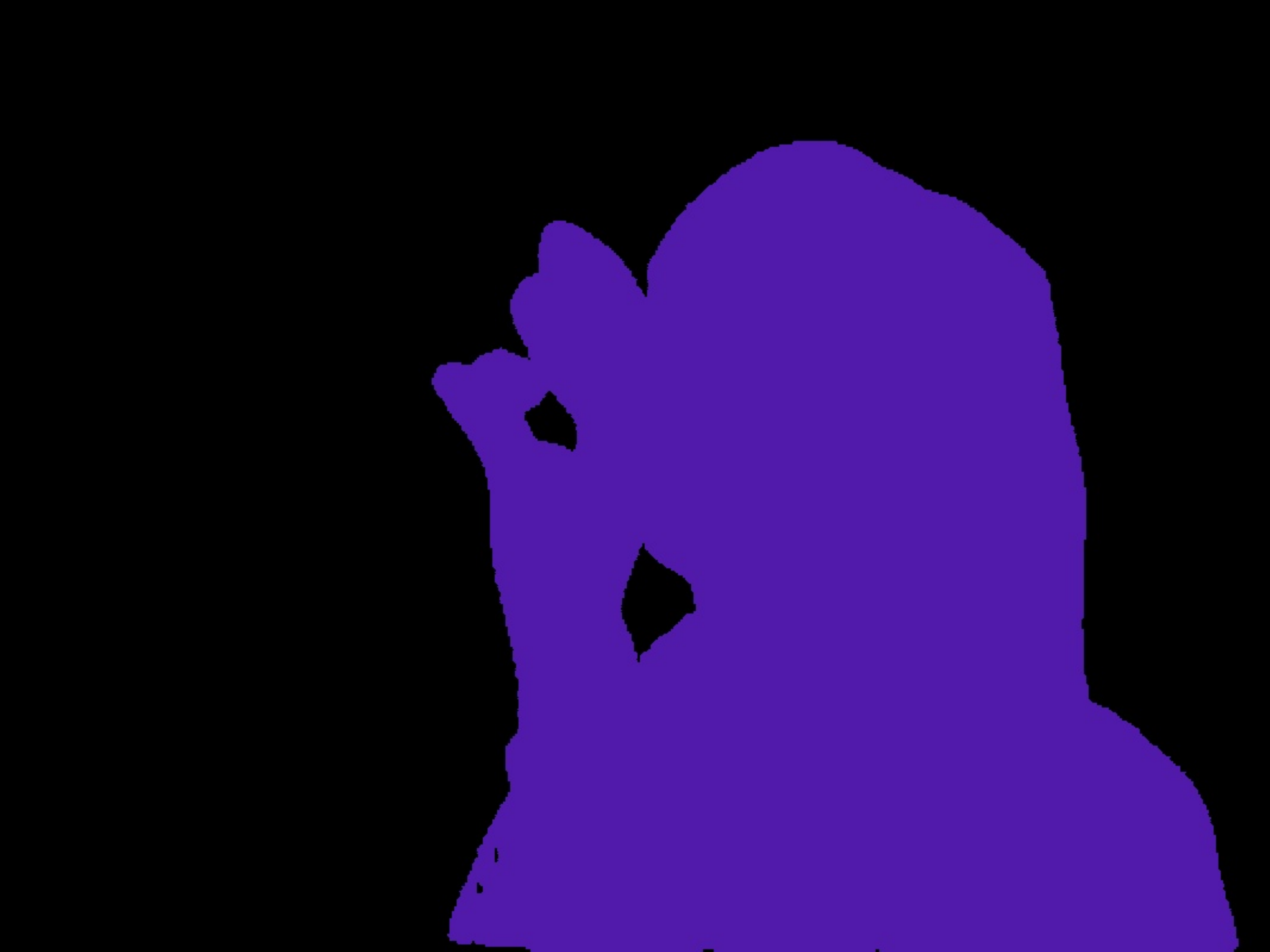} \\
			\includegraphics[height=1.5cm,width=2.0cm]{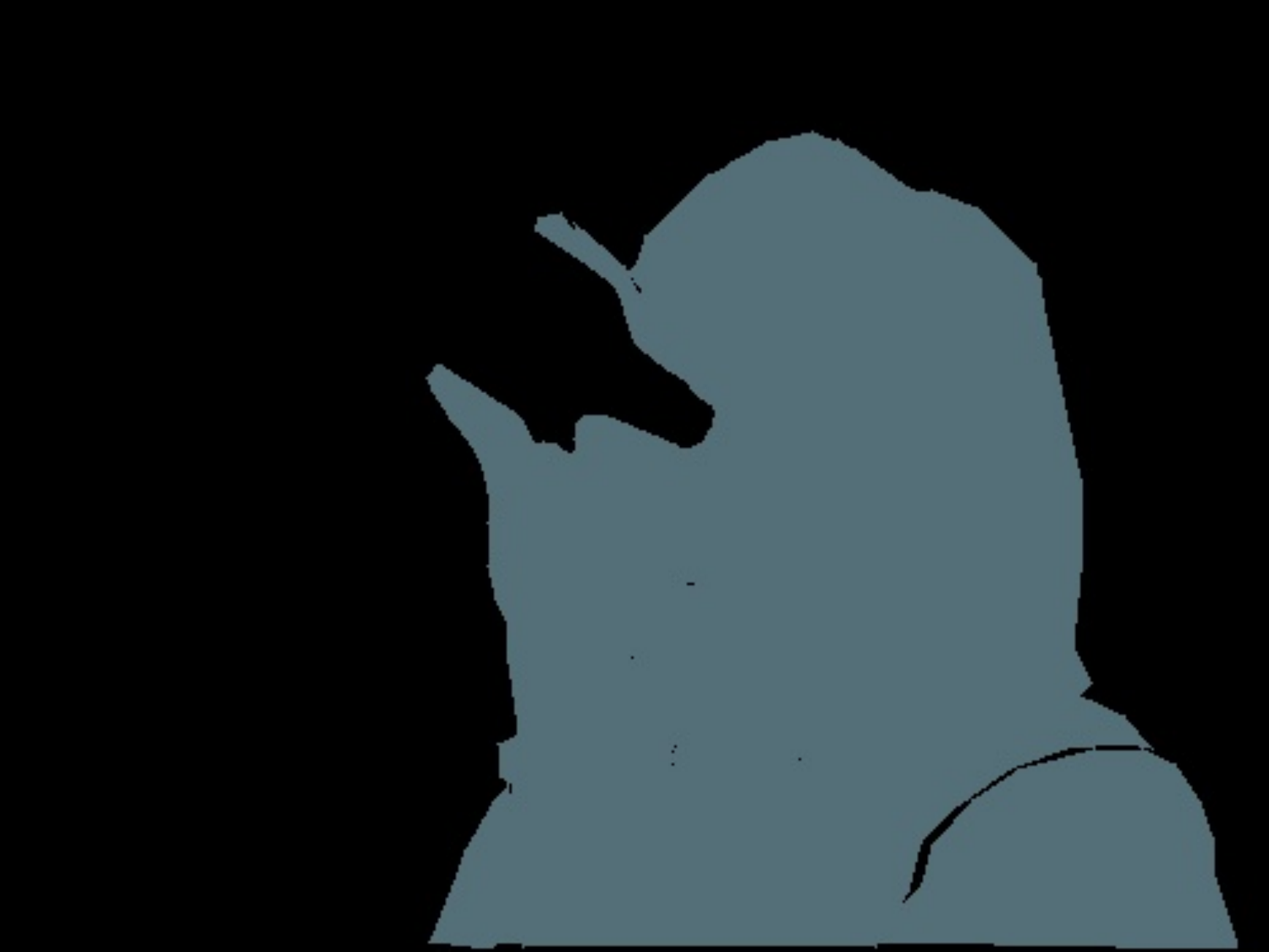} \\
		\end{minipage}
	}
	\subfigure
	{
		\begin{minipage}[b]{0.14286\linewidth}
			\centering
			\includegraphics[height=1.5cm,width=2.0cm]{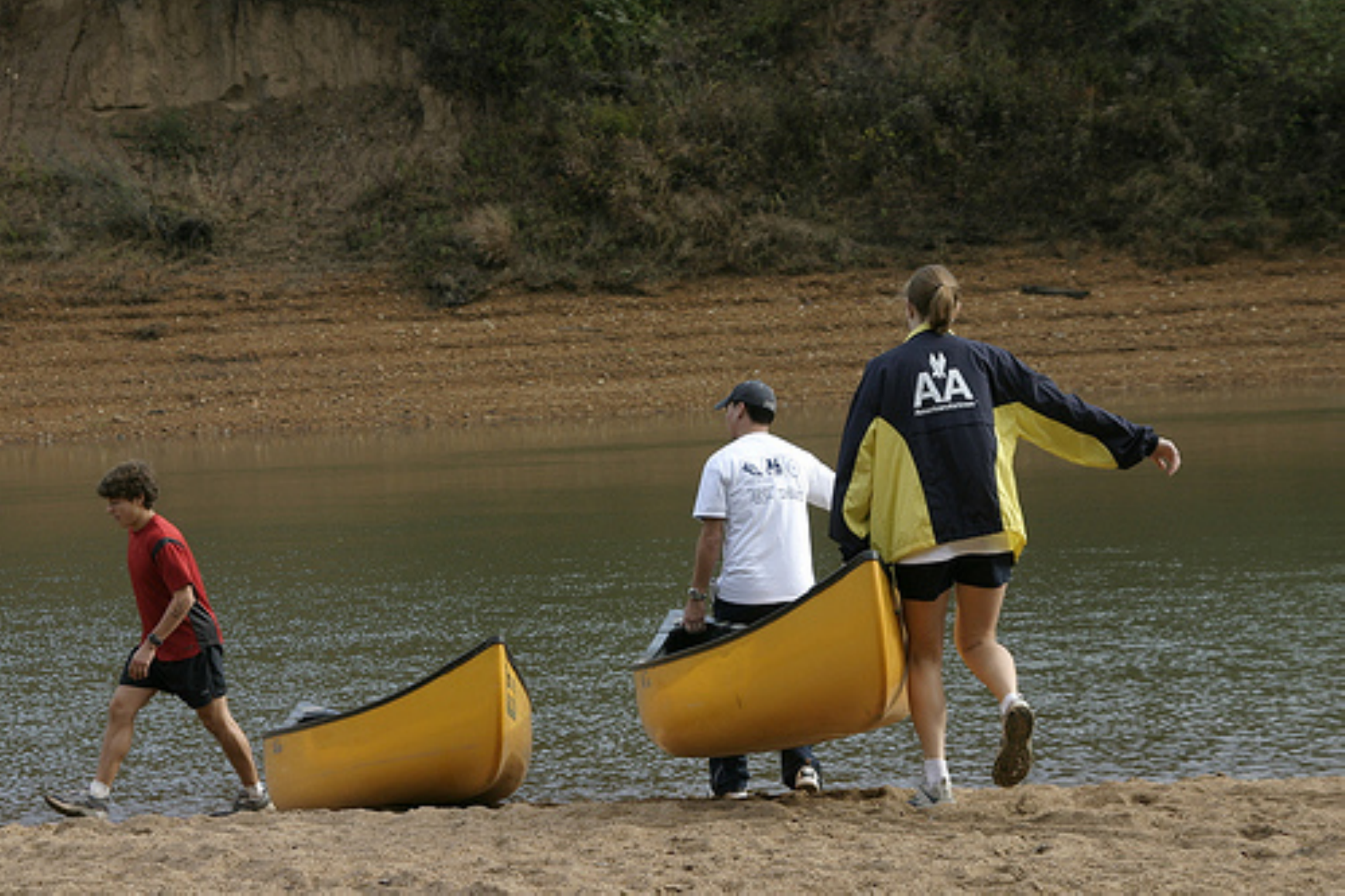} \\
			\includegraphics[height=1.5cm,width=2.0cm]{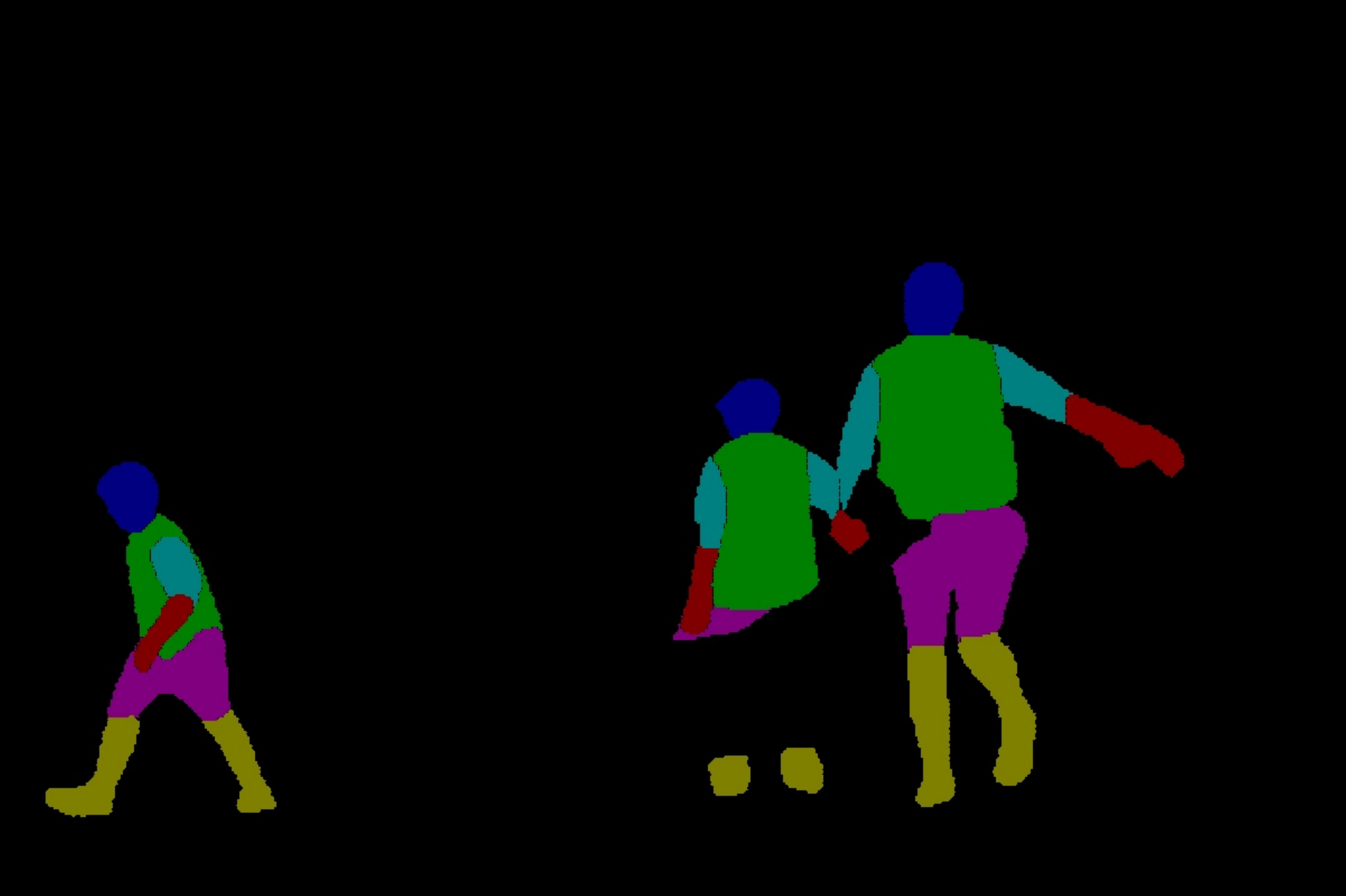} \\
			\includegraphics[height=1.5cm,width=2.0cm]{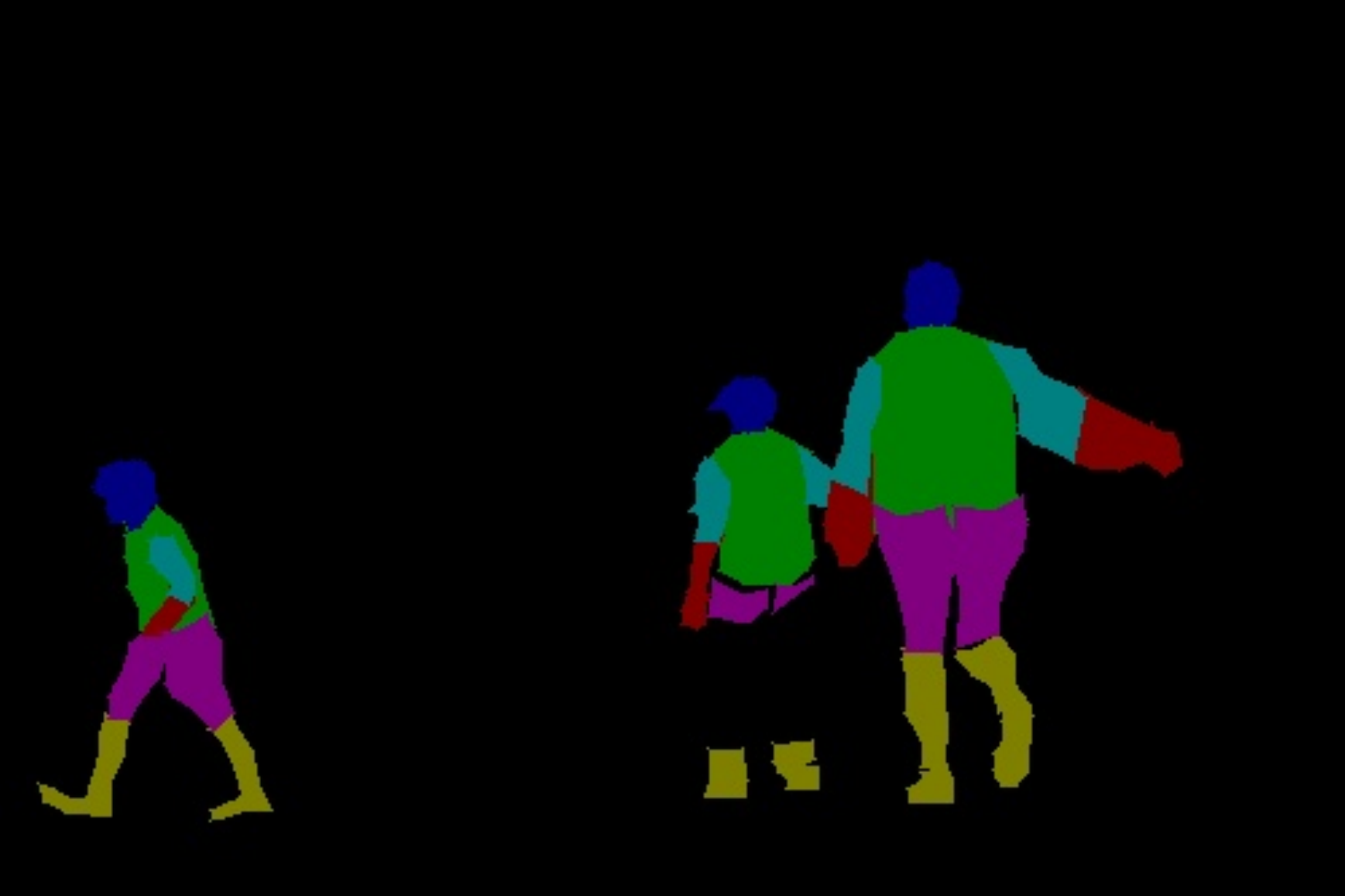} \\
			\includegraphics[height=1.5cm,width=2.0cm]{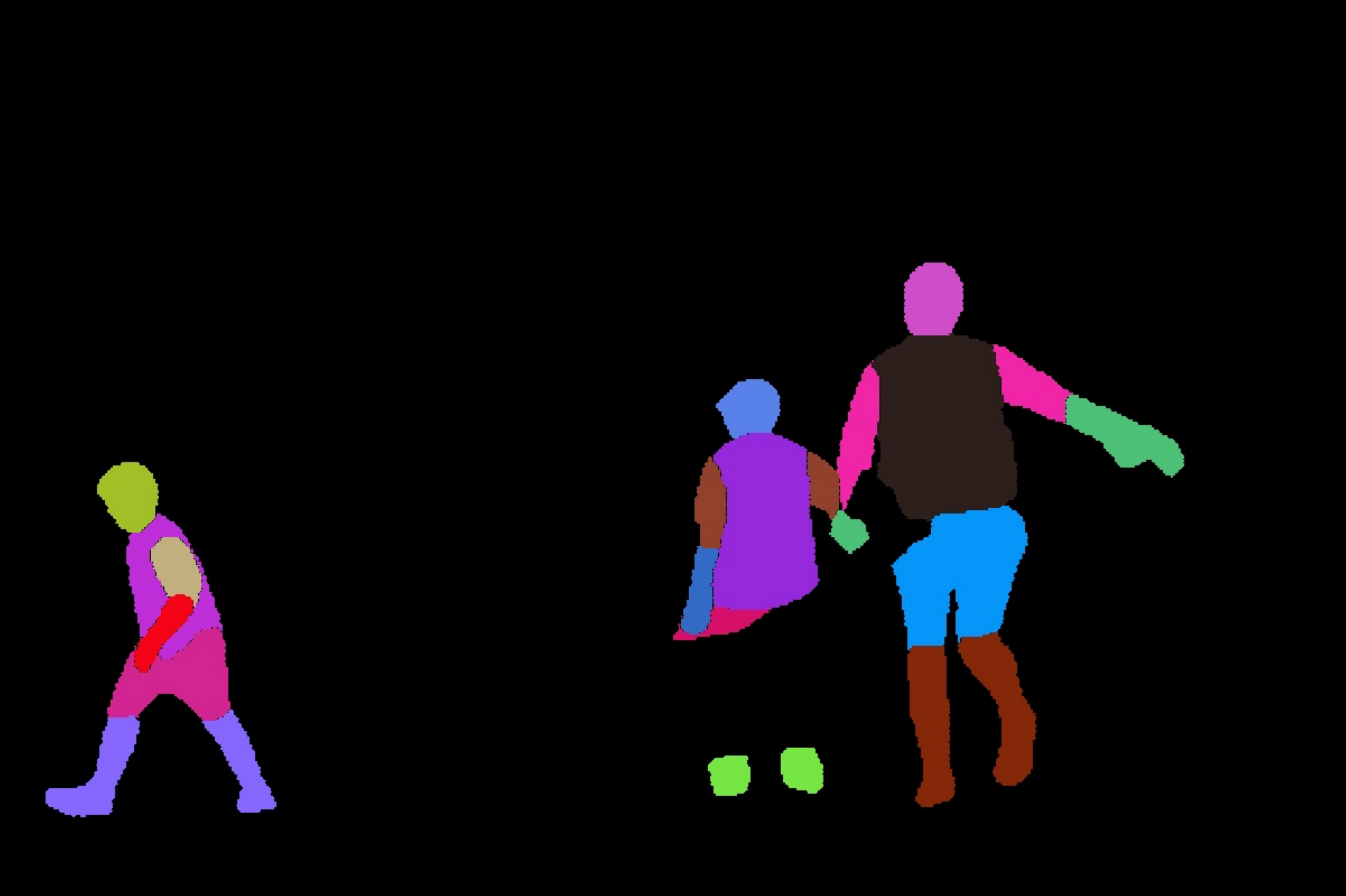} \\
			\includegraphics[height=1.5cm,width=2.0cm]{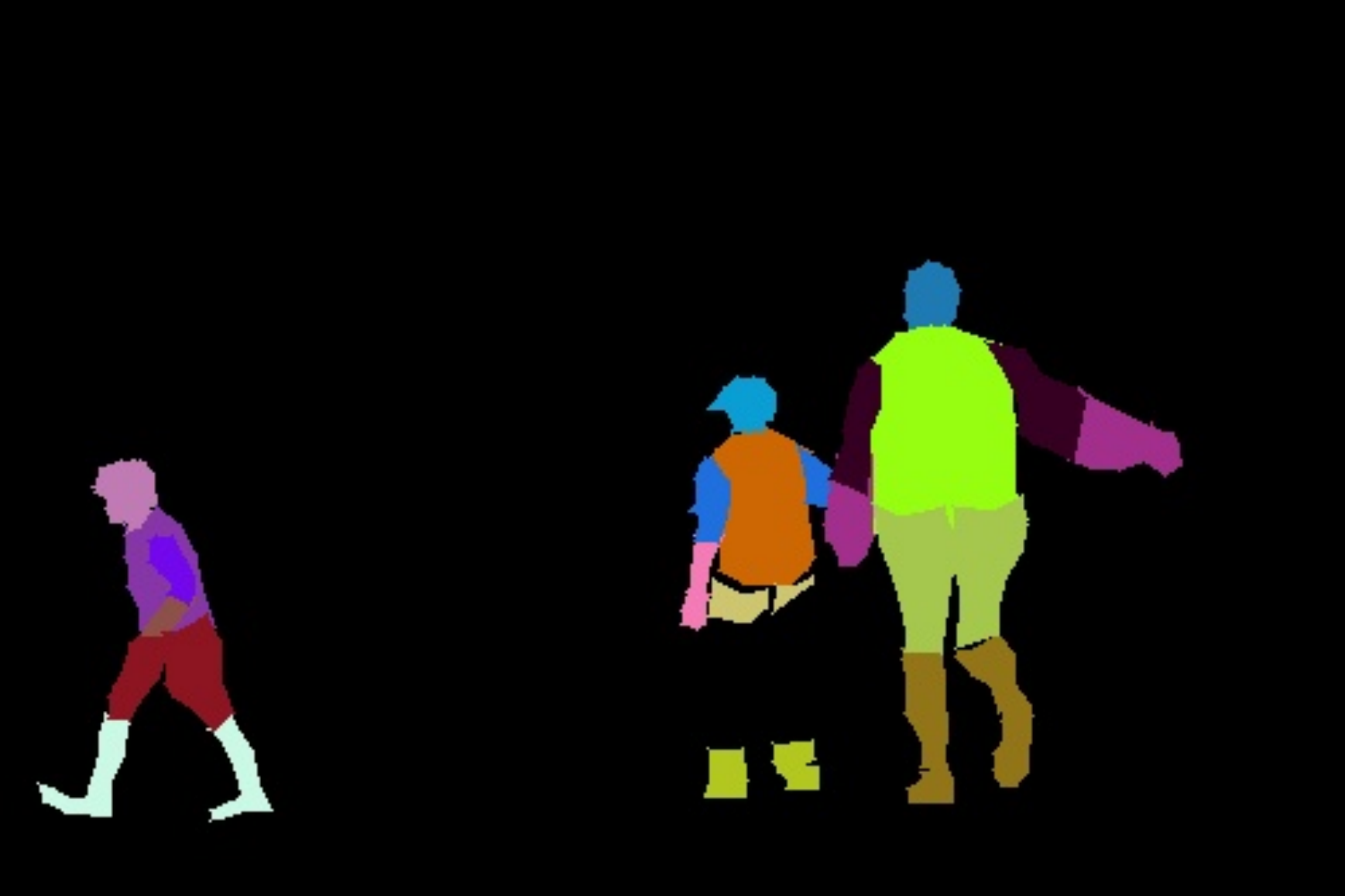} \\
			\includegraphics[height=1.5cm,width=2.0cm]{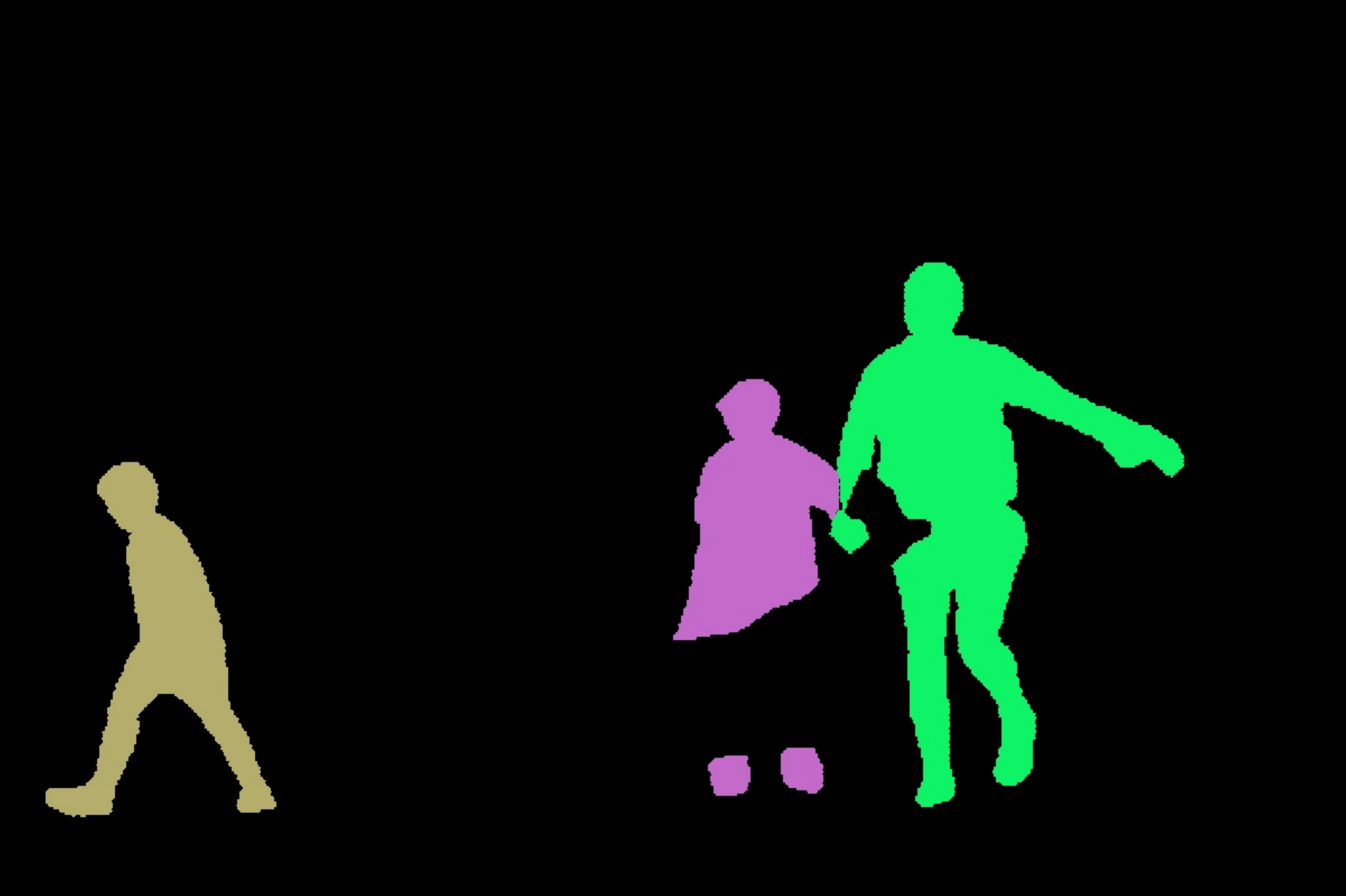} \\
			\includegraphics[height=1.5cm,width=2.0cm]{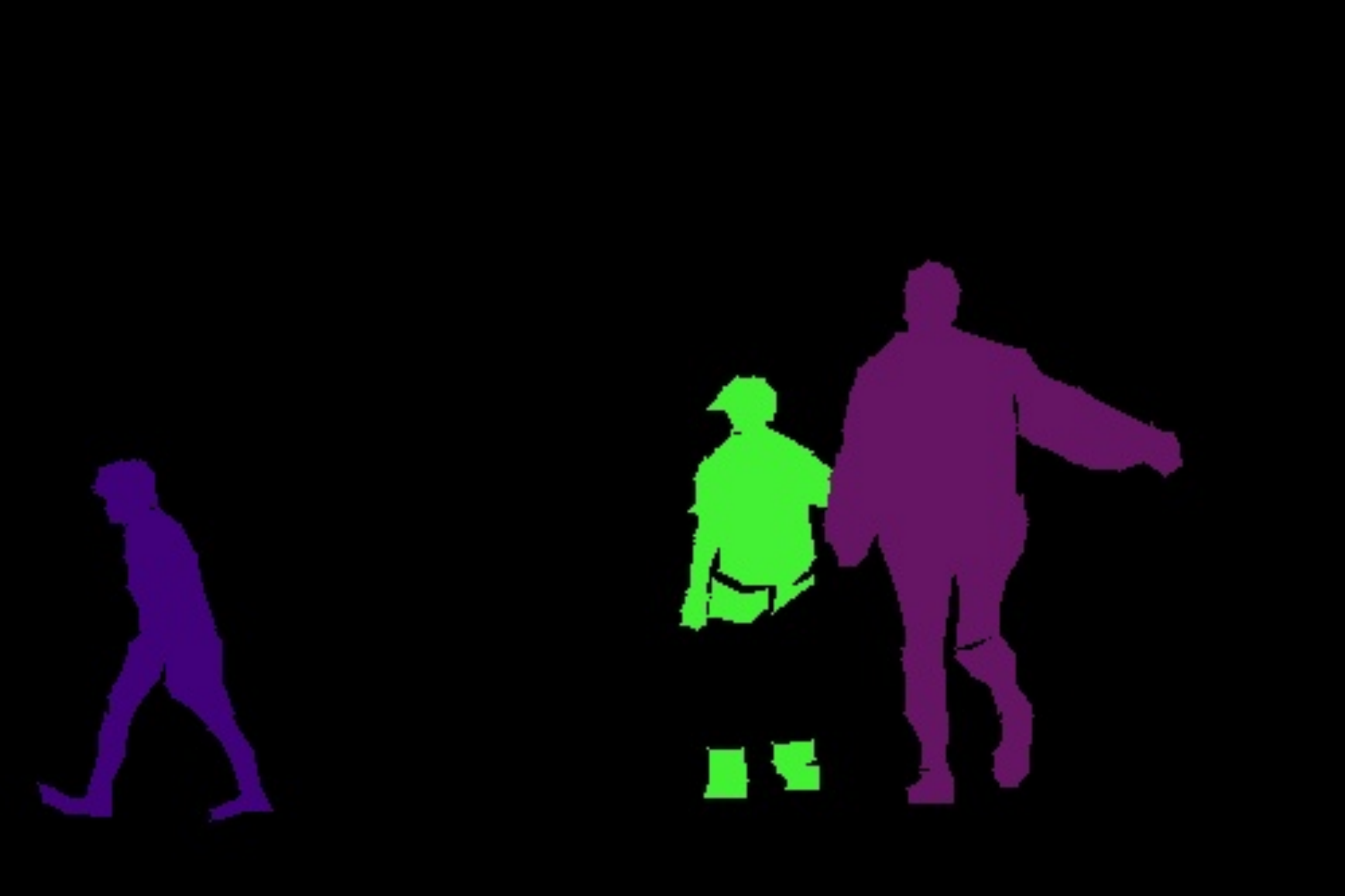} \\
		\end{minipage}
	}
	\subfigure
	{
	\begin{minipage}[b]{0.14286\linewidth}
		\centering
		\includegraphics[height=1.5cm,width=2.0cm]{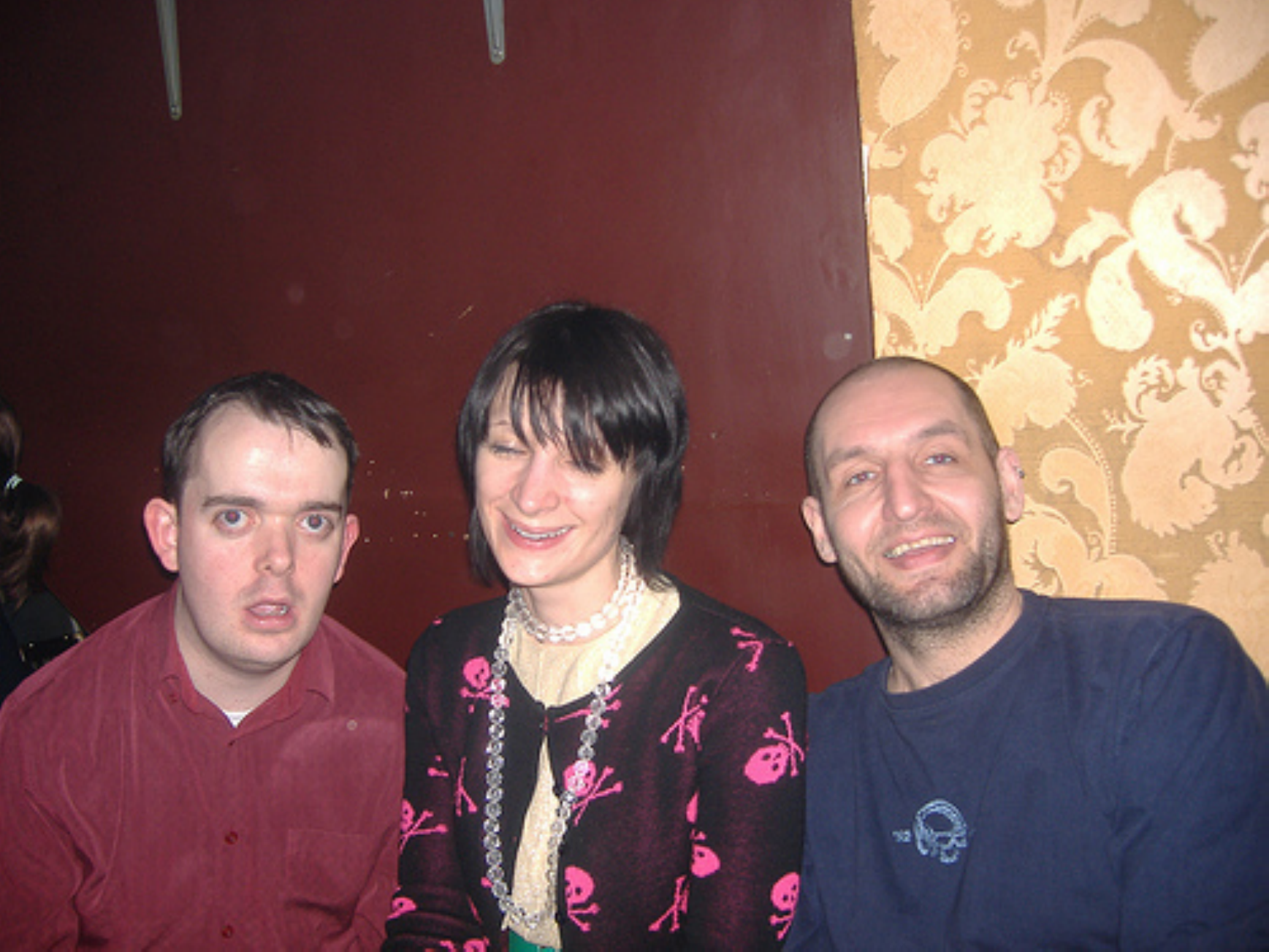} \\
		\includegraphics[height=1.5cm,width=2.0cm]{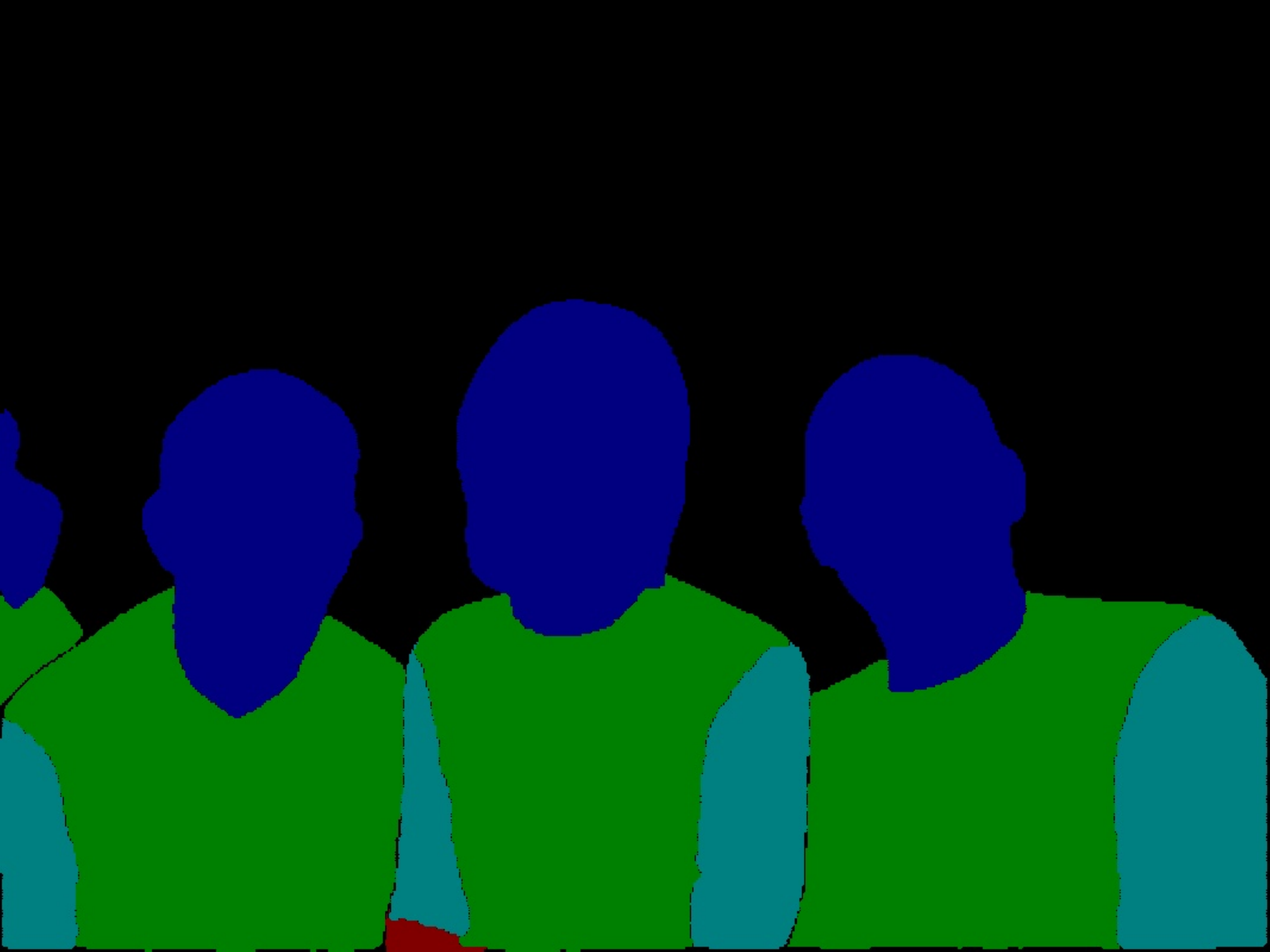} \\
		\includegraphics[height=1.5cm,width=2.0cm]{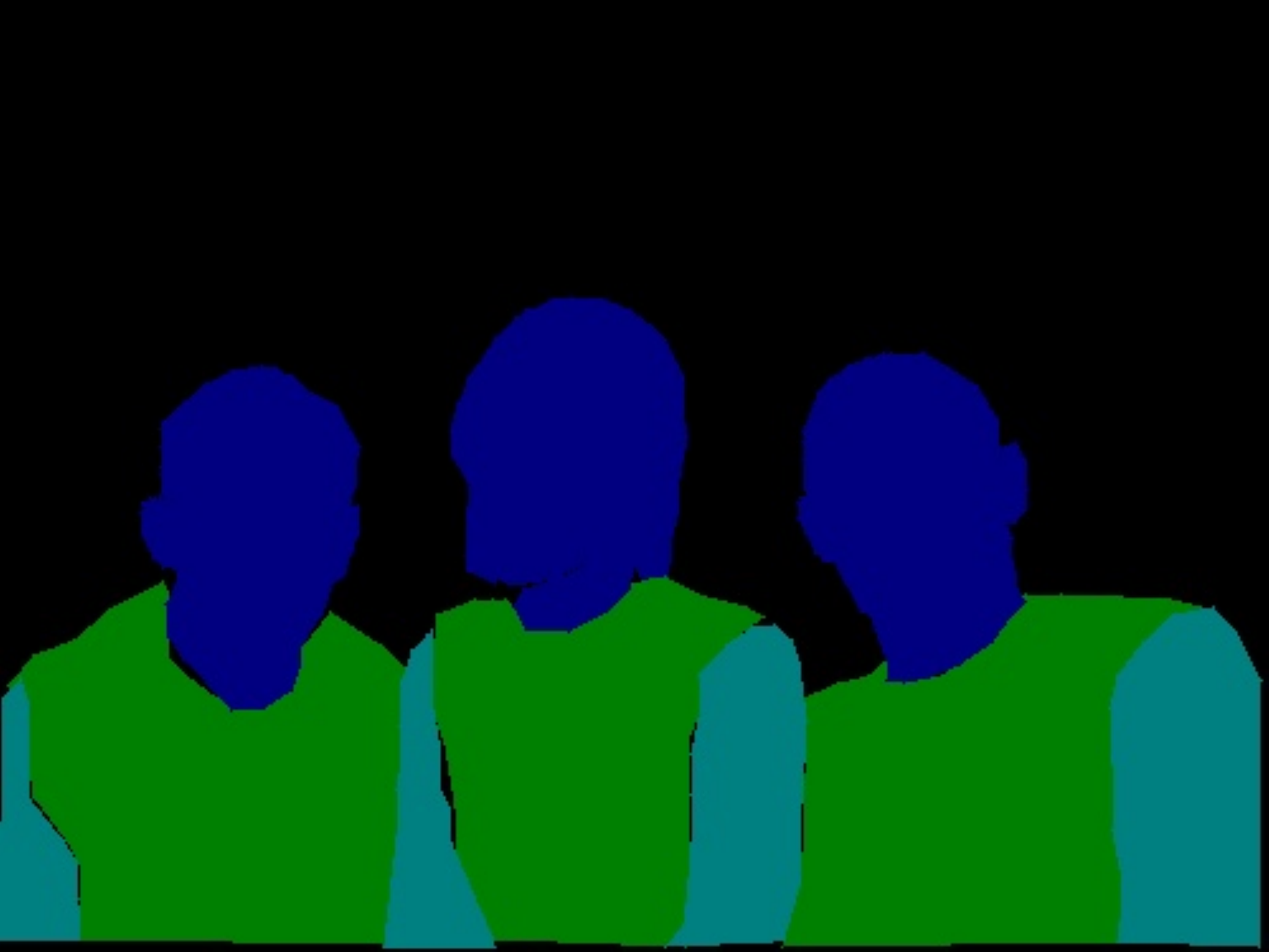} \\
		\includegraphics[height=1.5cm,width=2.0cm]{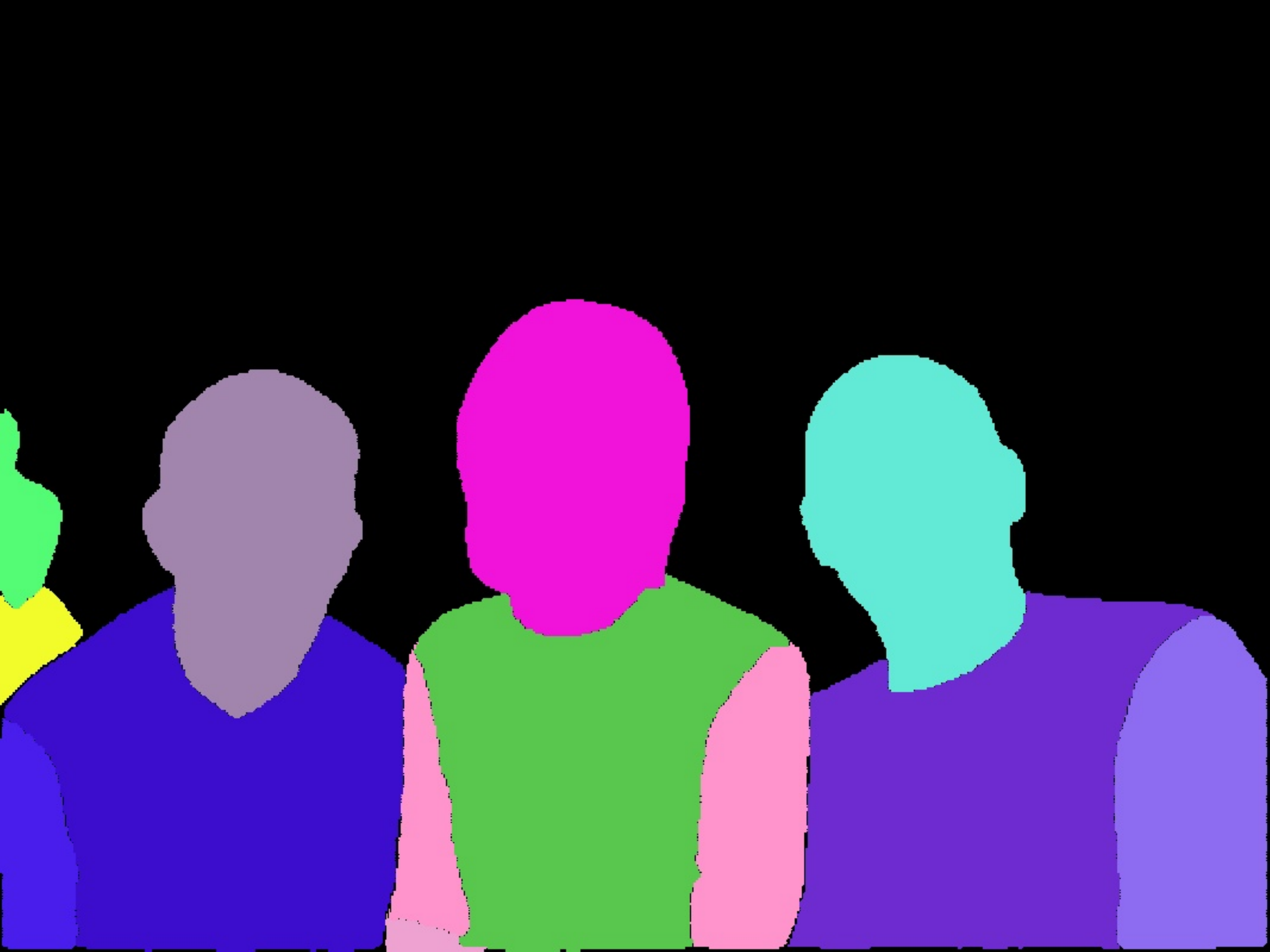} \\
		\includegraphics[height=1.5cm,width=2.0cm]{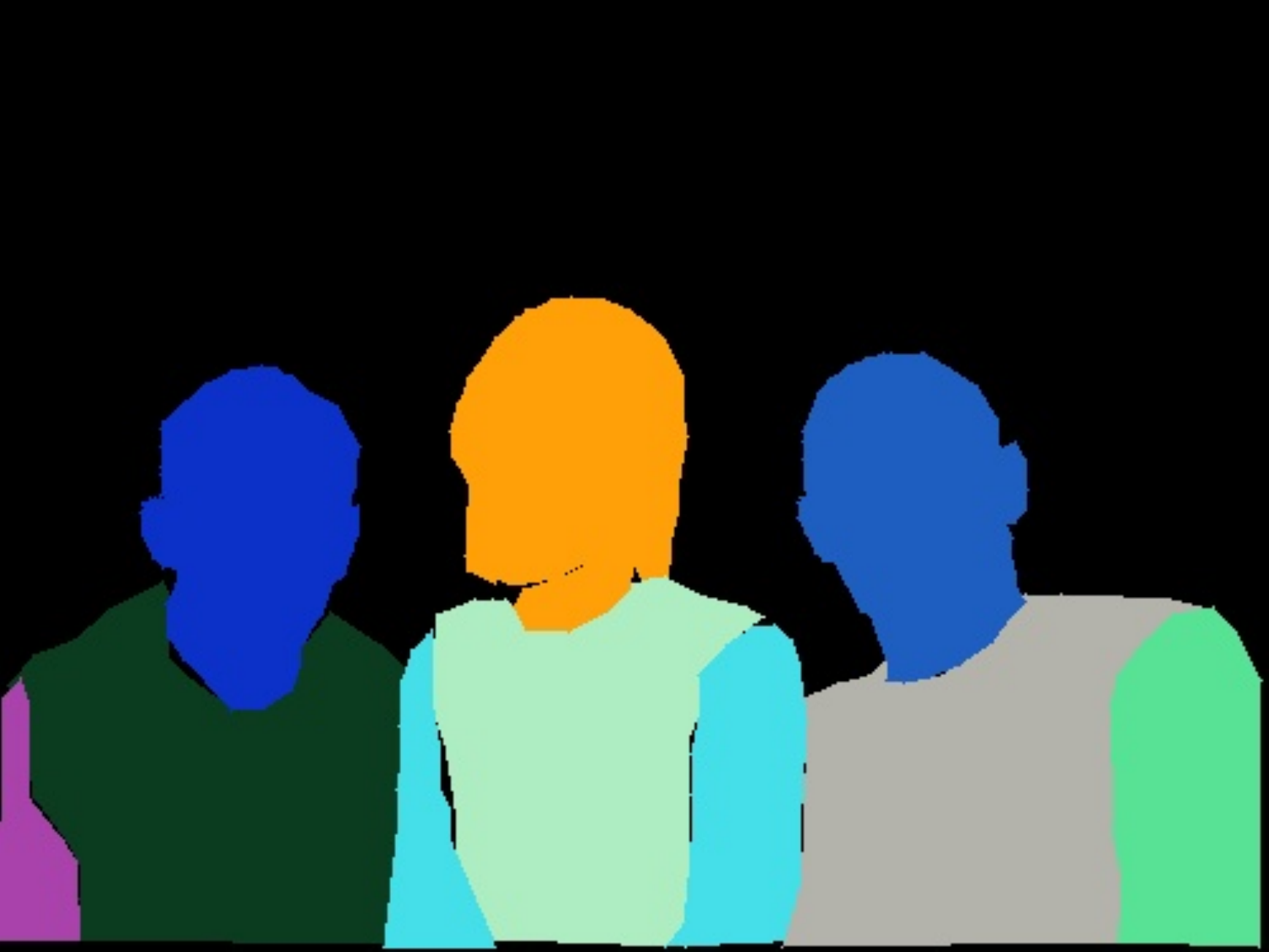} \\
		\includegraphics[height=1.5cm,width=2.0cm]{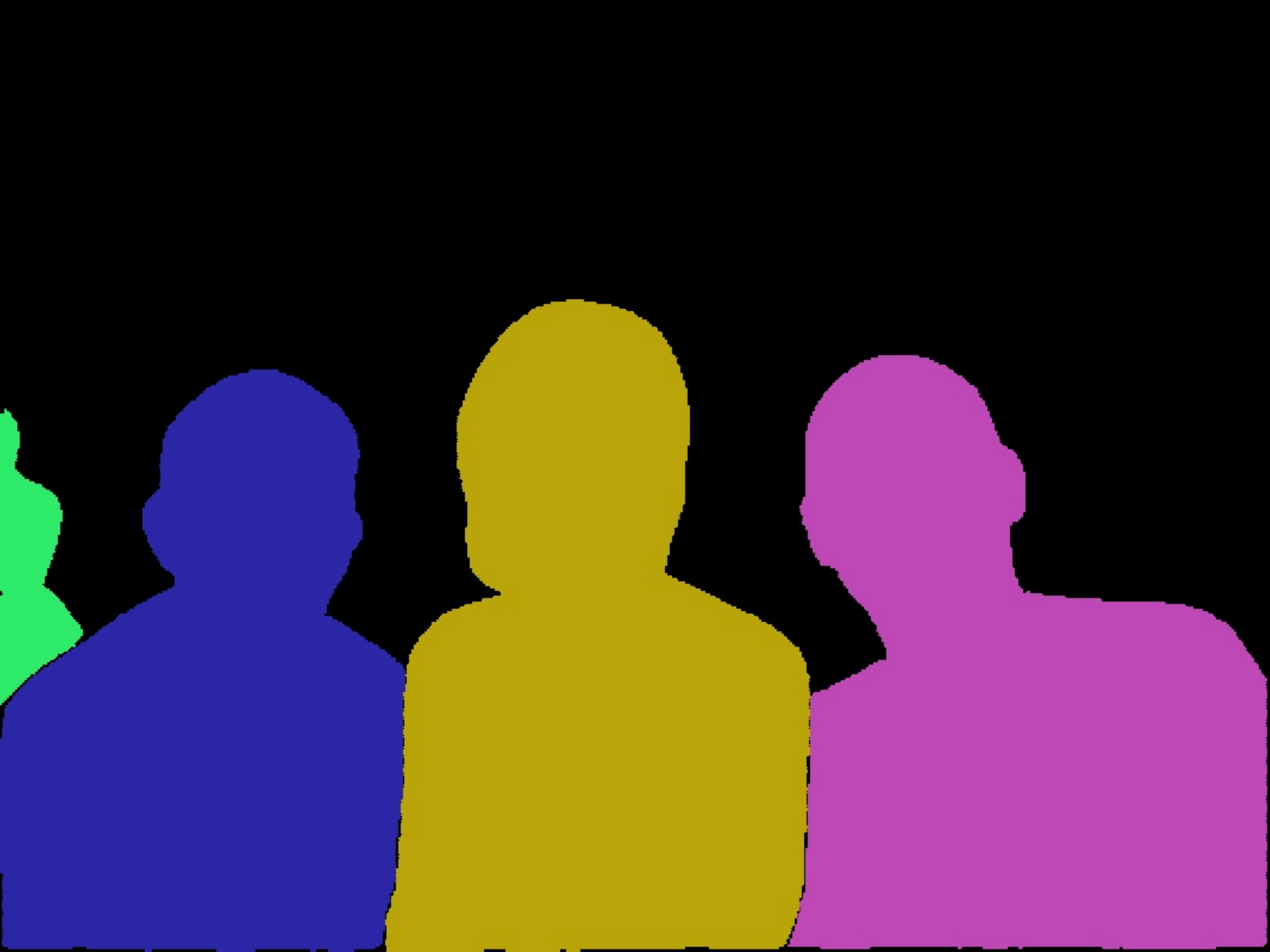} \\
		\includegraphics[height=1.5cm,width=2.0cm]{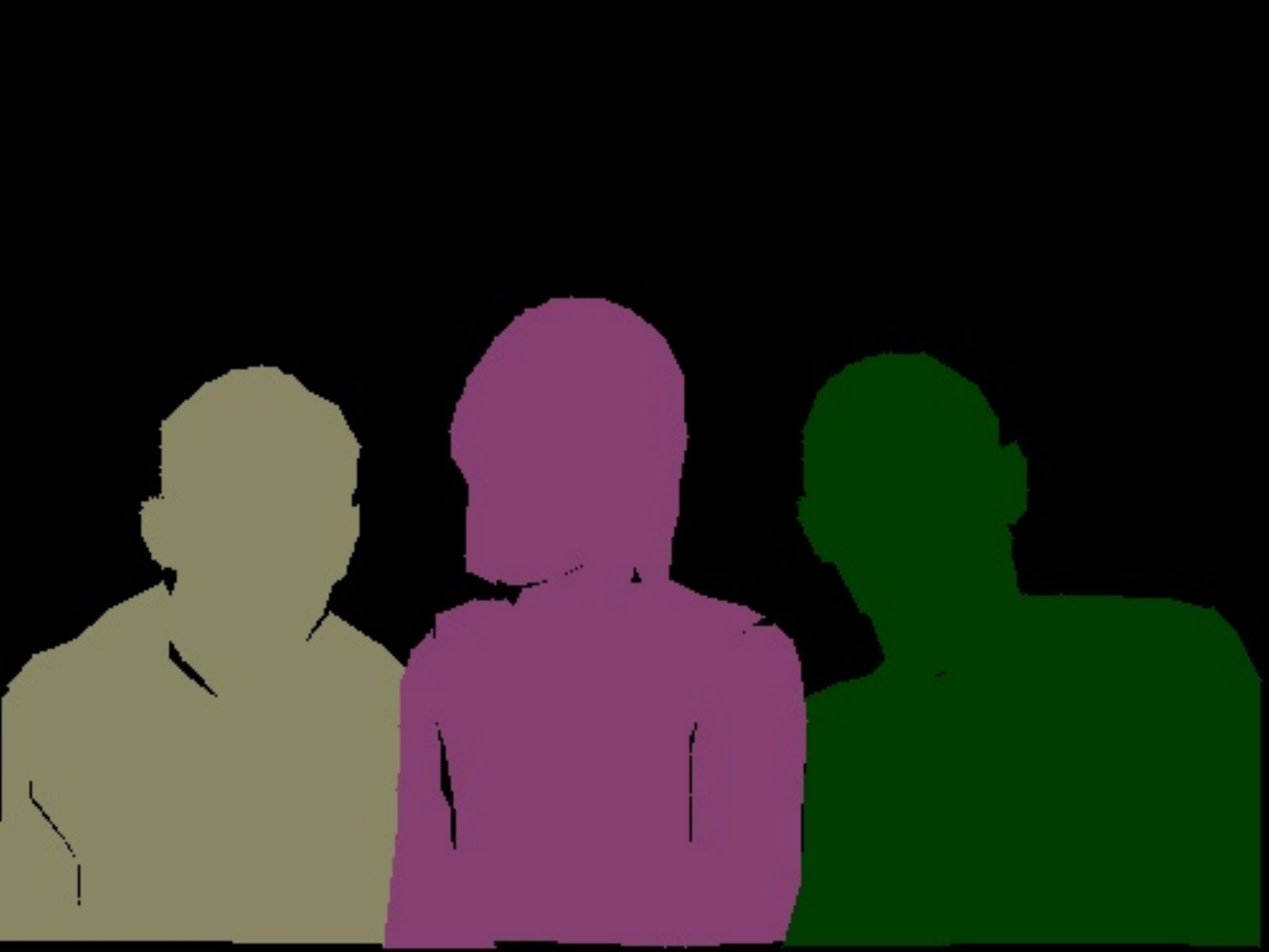} \\
	\end{minipage}
	}
	\subfigure
	{
		\begin{minipage}[b]{0.08071\linewidth}
			\centering
			\includegraphics[height=1.5cm,width=1.27cm]{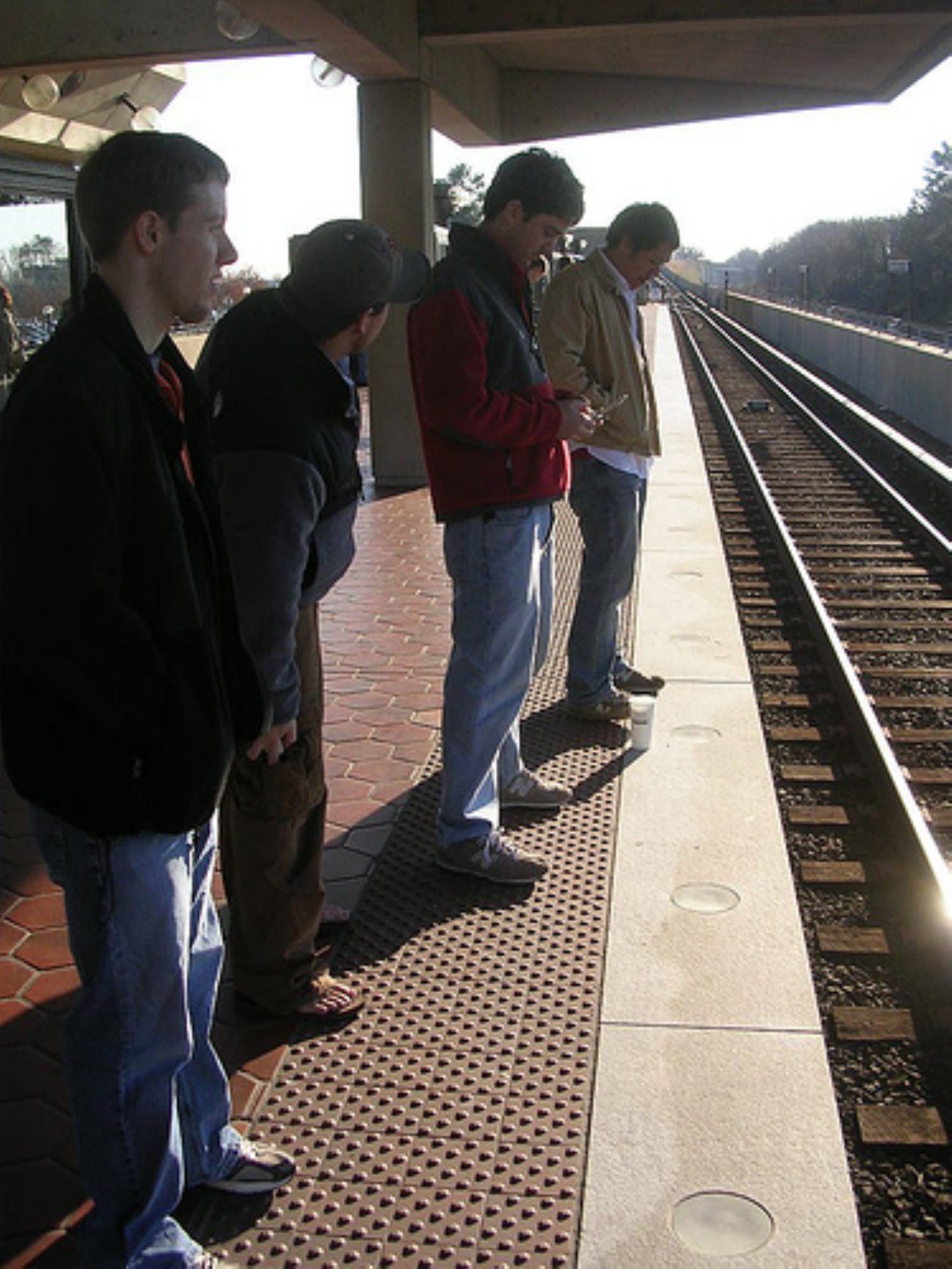} \\
			\includegraphics[height=1.5cm,width=1.27cm]{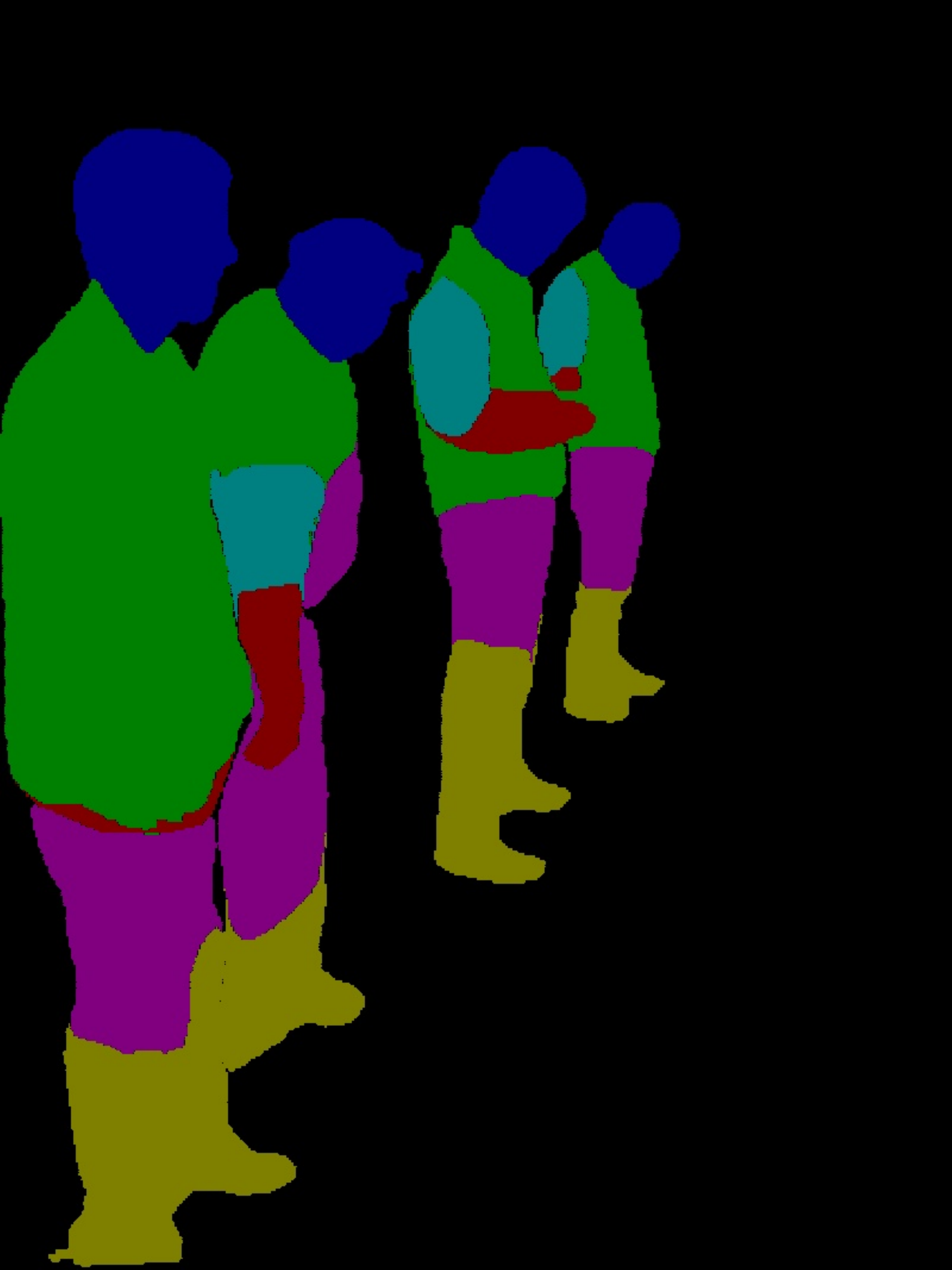} \\
			\includegraphics[height=1.5cm,width=1.27cm]{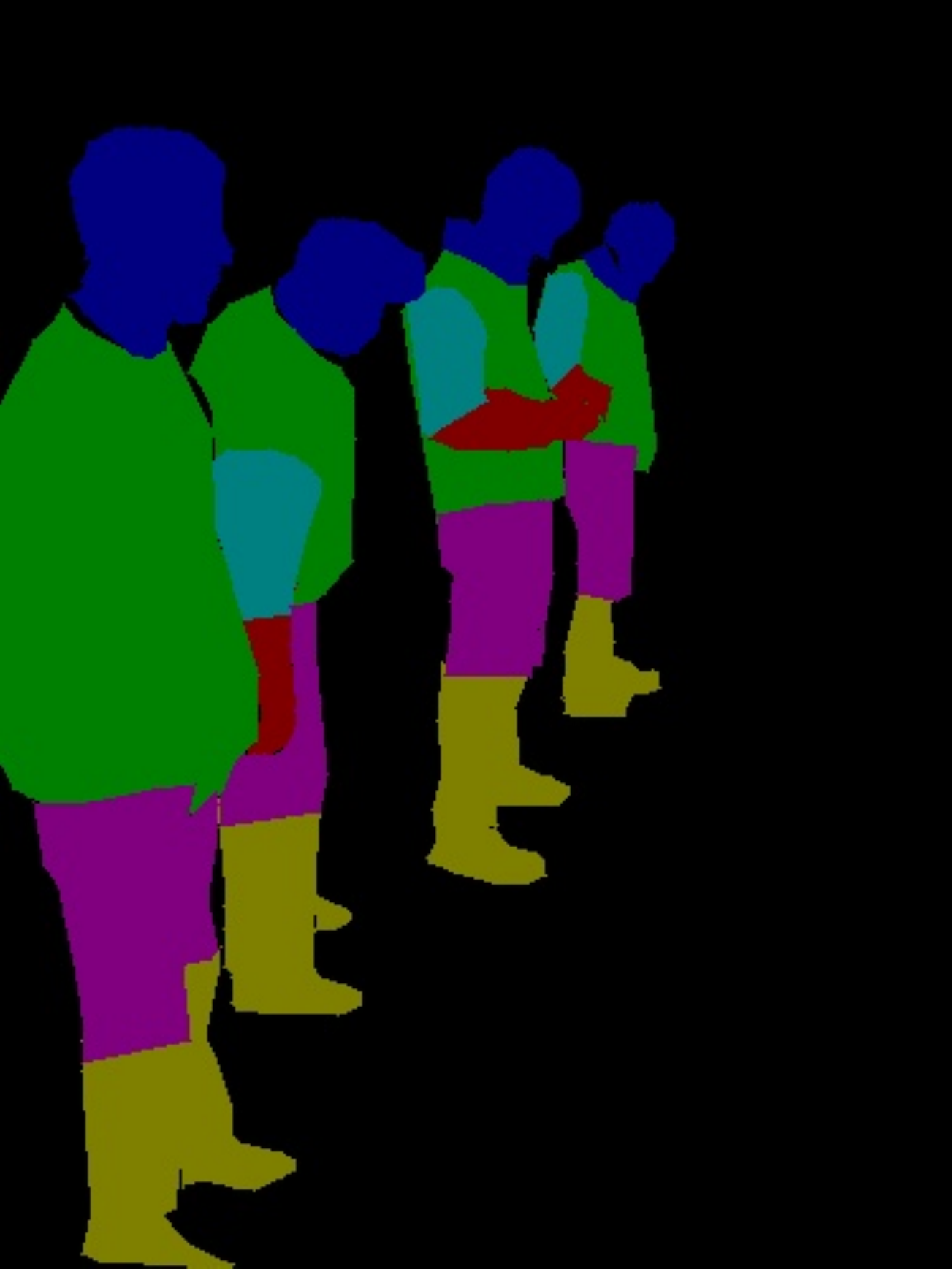} \\
			\includegraphics[height=1.5cm,width=1.27cm]{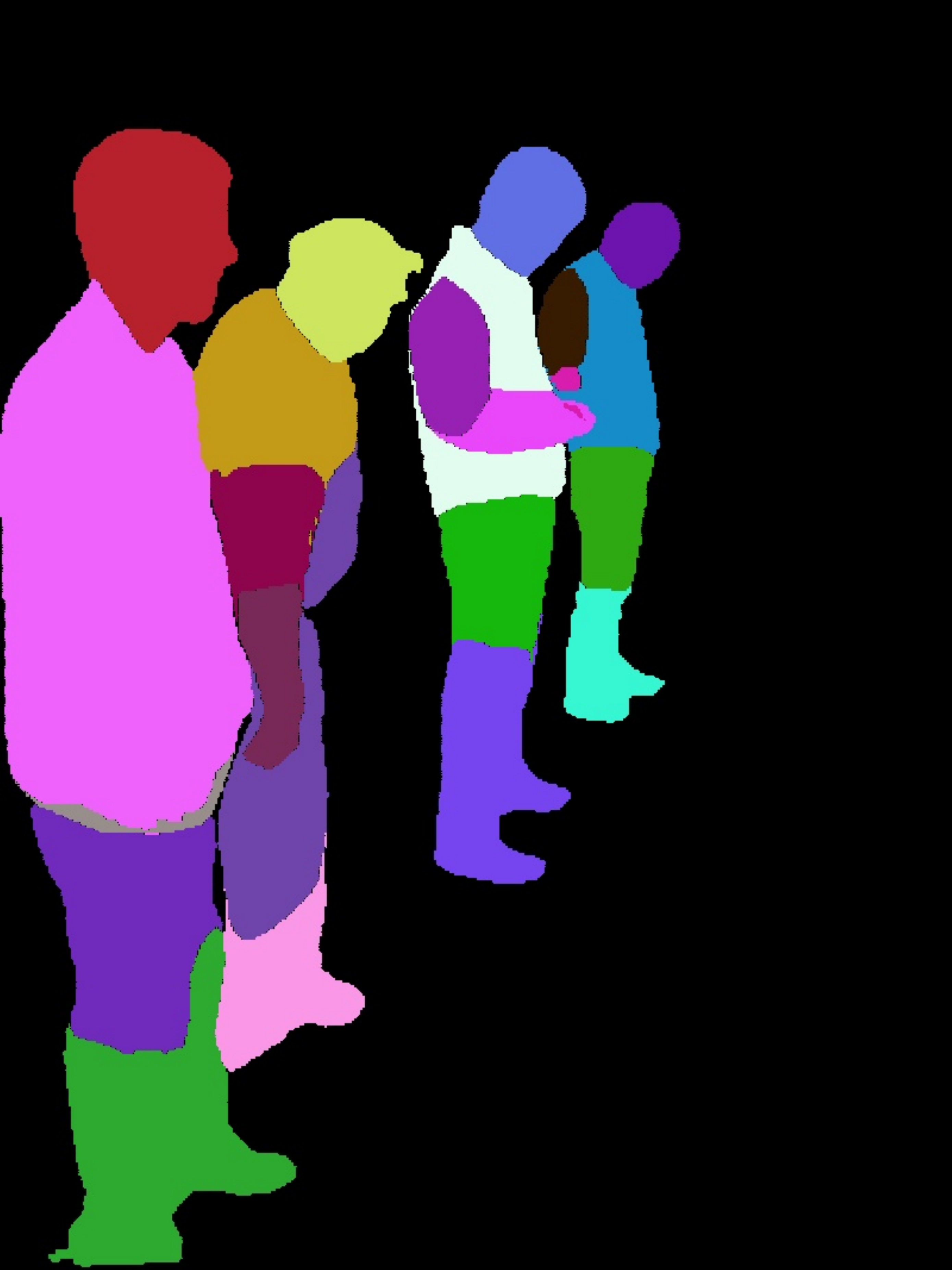} \\
			\includegraphics[height=1.5cm,width=1.27cm]{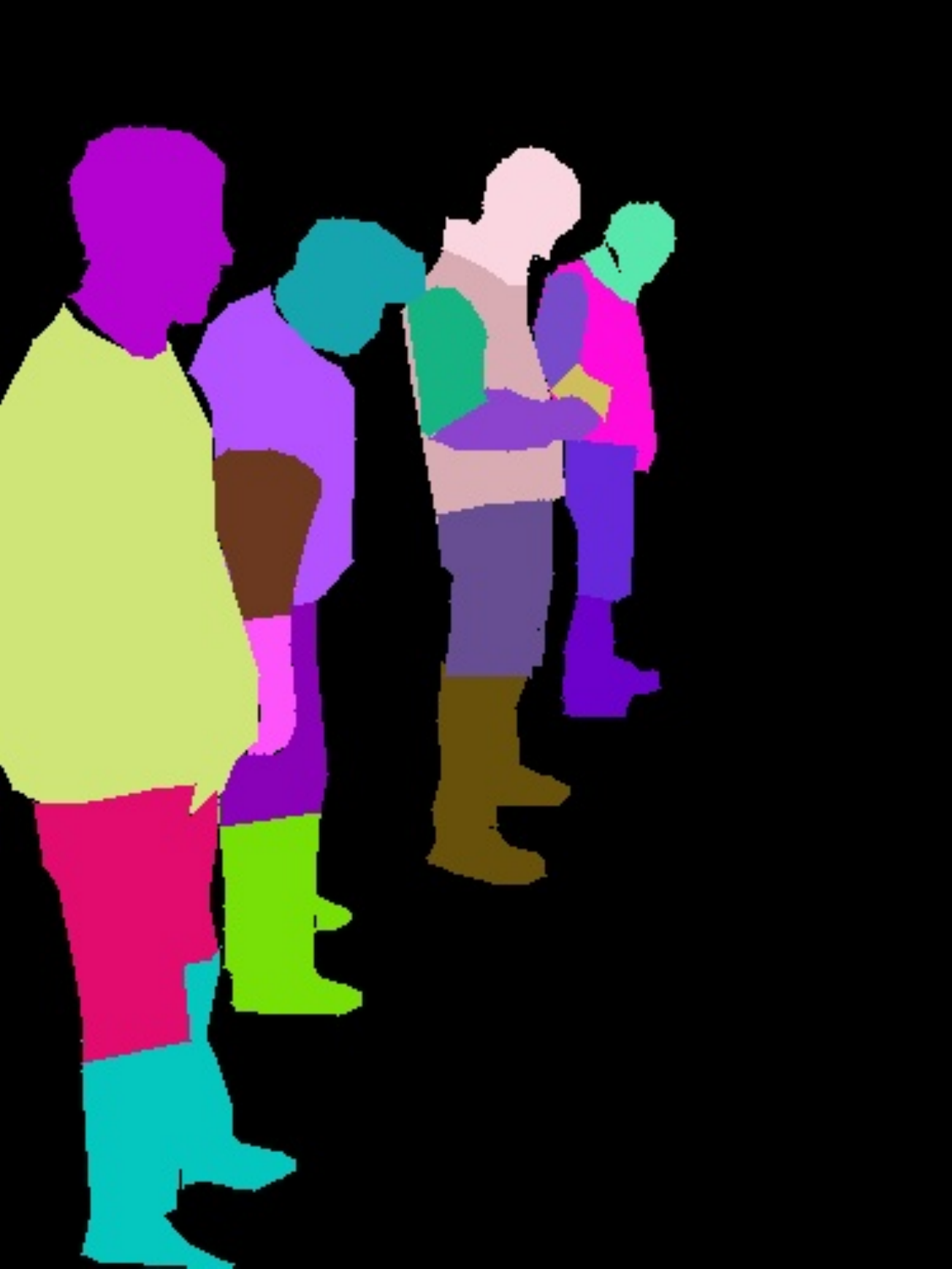} \\
			\includegraphics[height=1.5cm,width=1.27cm]{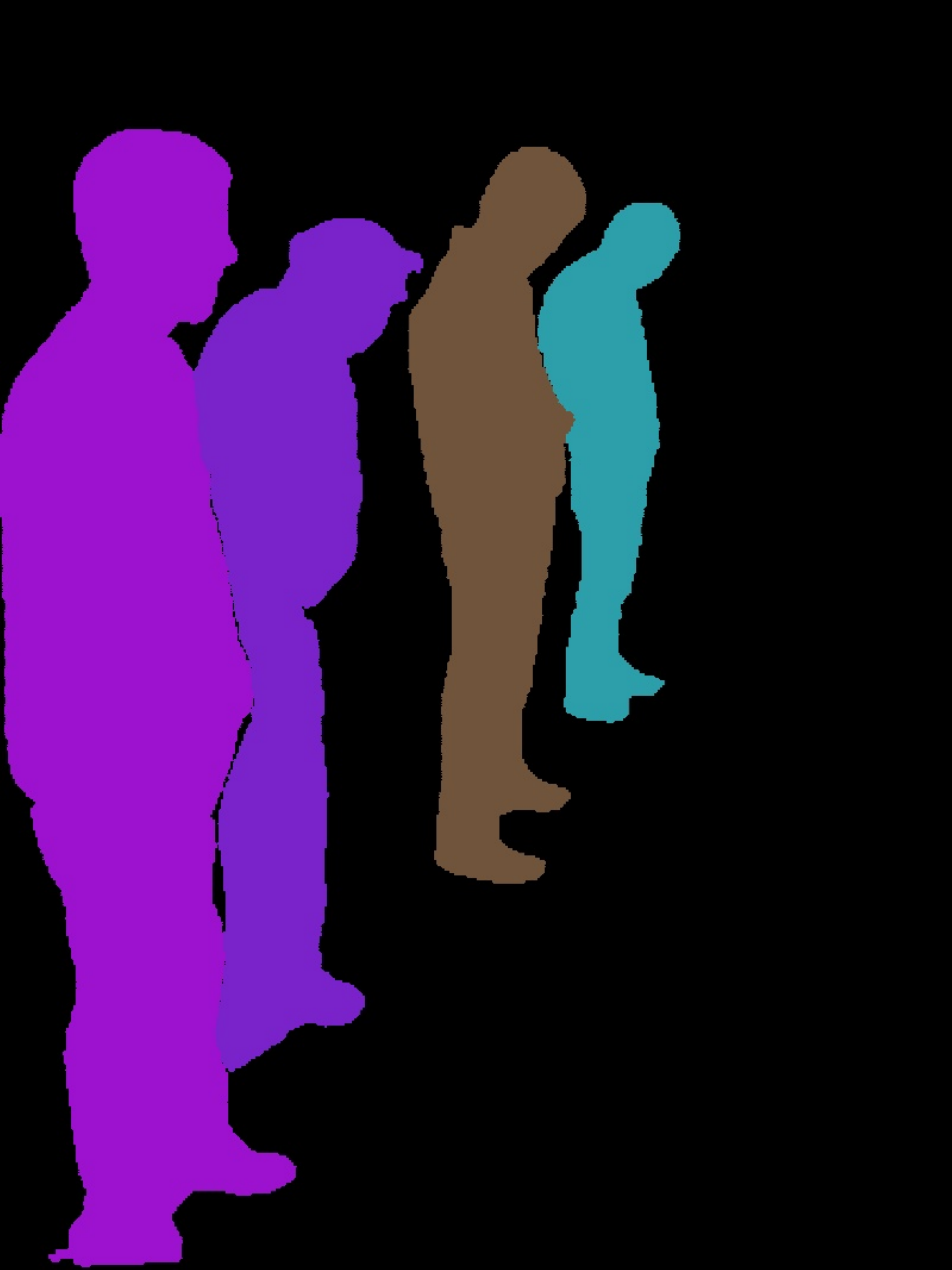} \\
			\includegraphics[height=1.5cm,width=1.27cm]{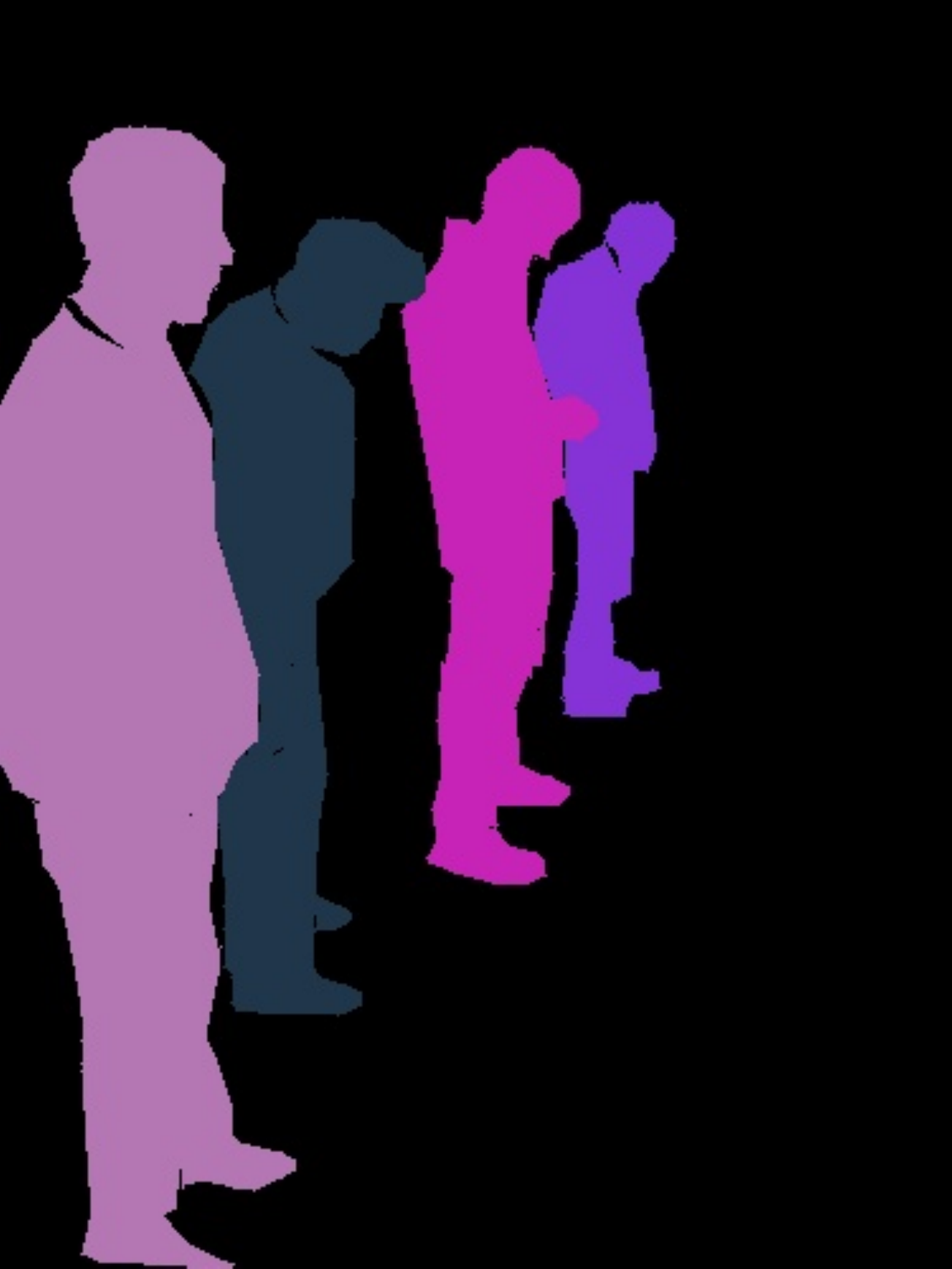} \\
		\end{minipage}
	}
	\subfigure
	{
		\begin{minipage}[b]{0.16071\linewidth}
			\centering
			\includegraphics[height=1.5cm,width=2.25cm]{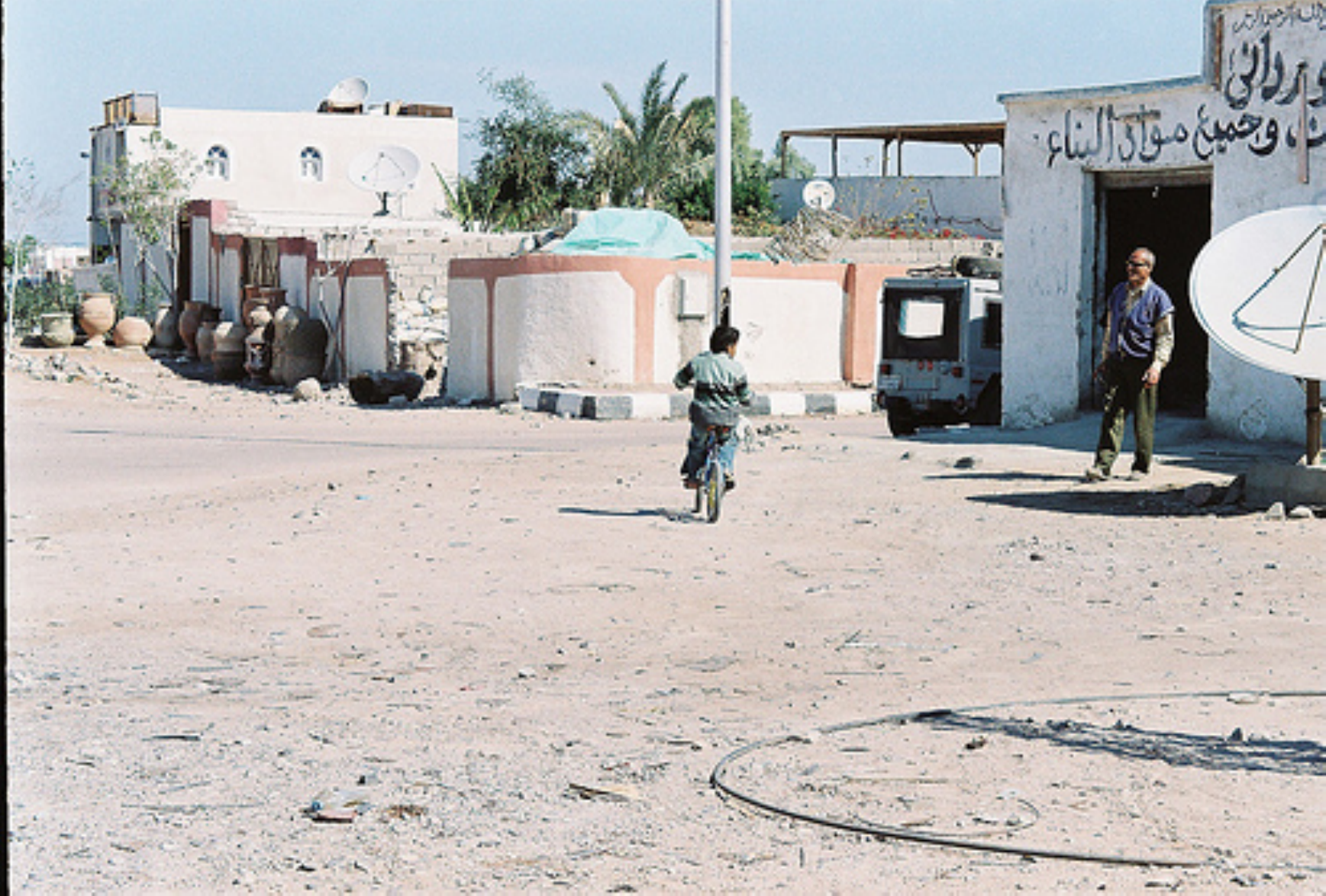} \\
			\includegraphics[height=1.5cm,width=2.25cm]{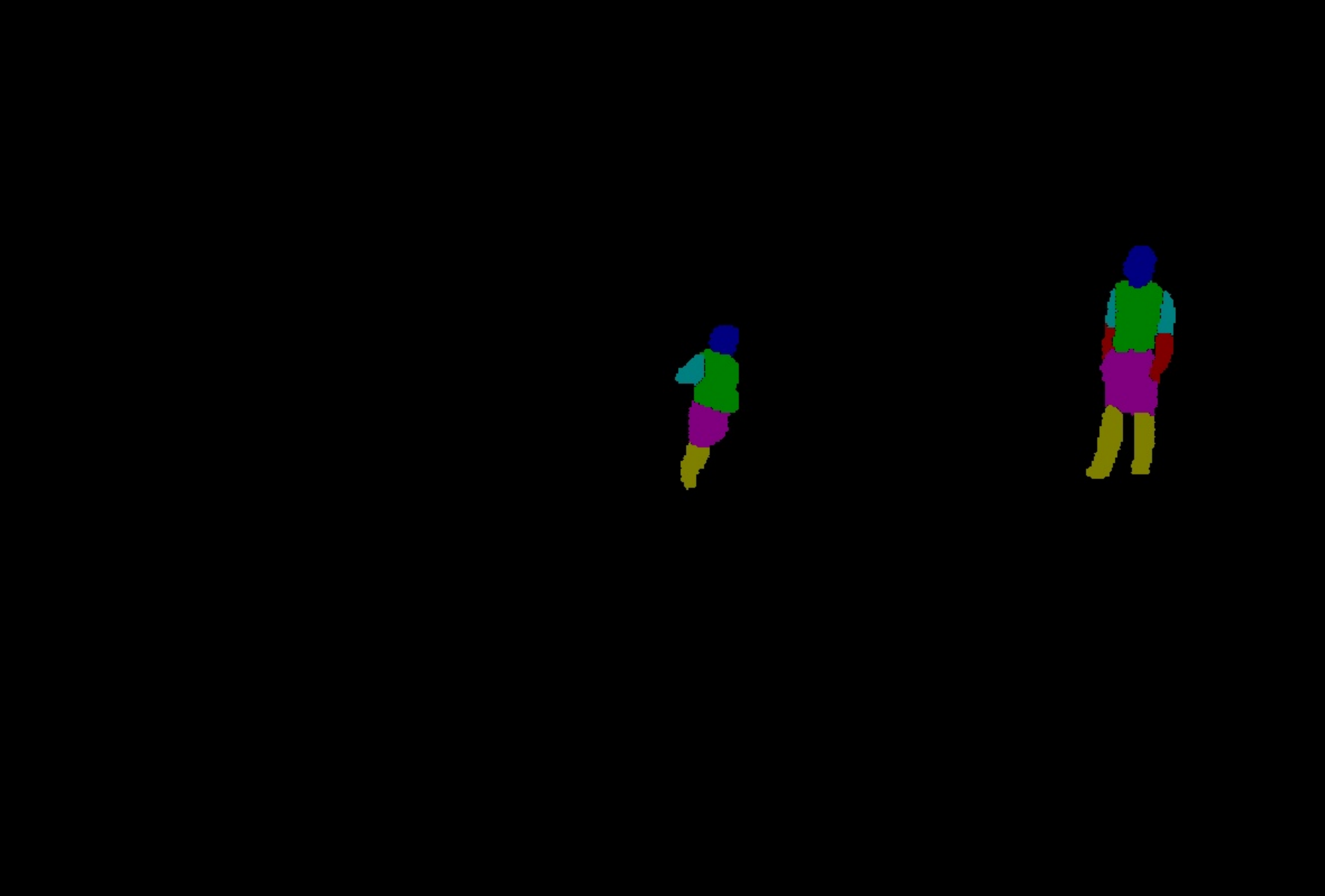} \\
			\includegraphics[height=1.5cm,width=2.25cm]{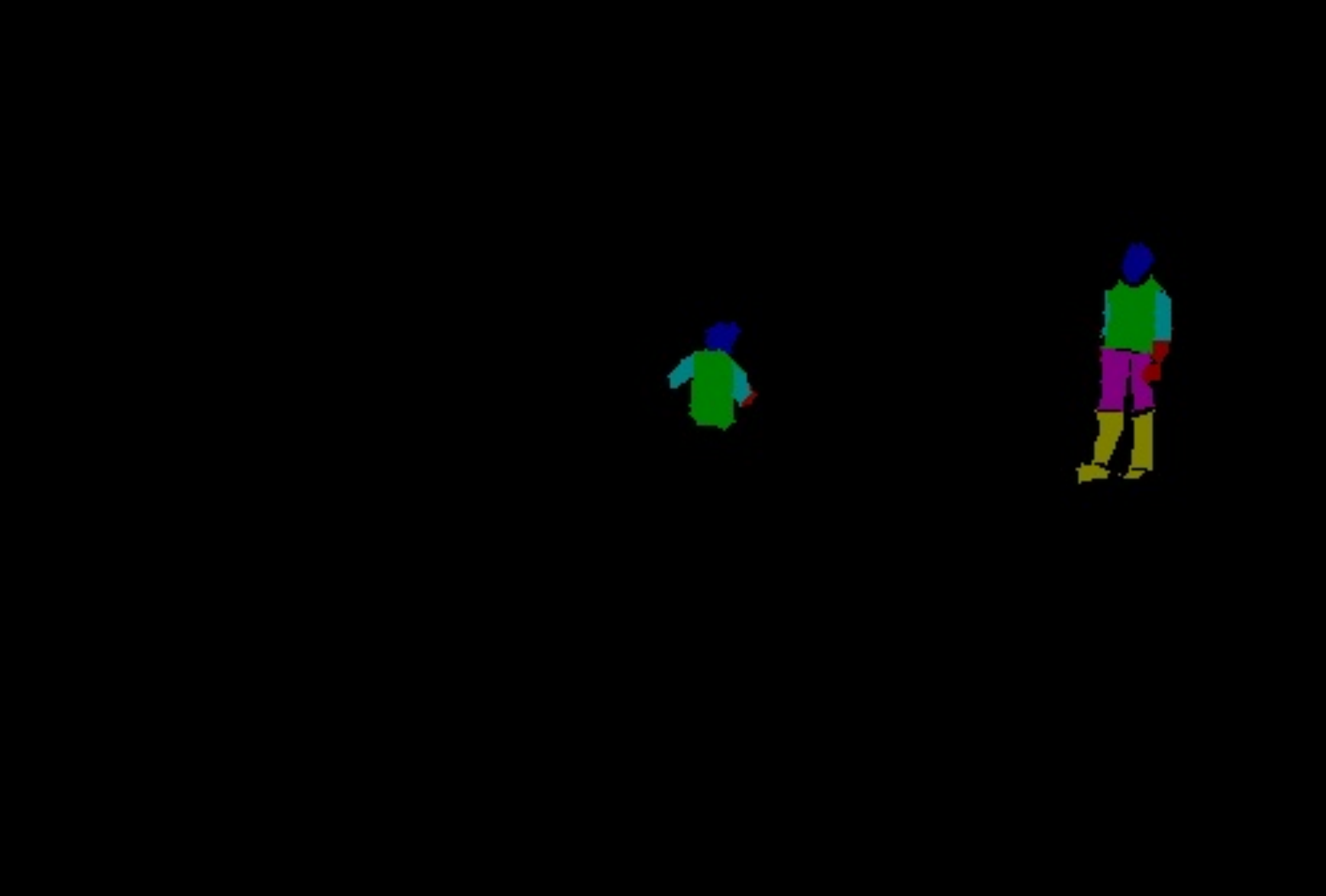} \\
			\includegraphics[height=1.5cm,width=2.25cm]{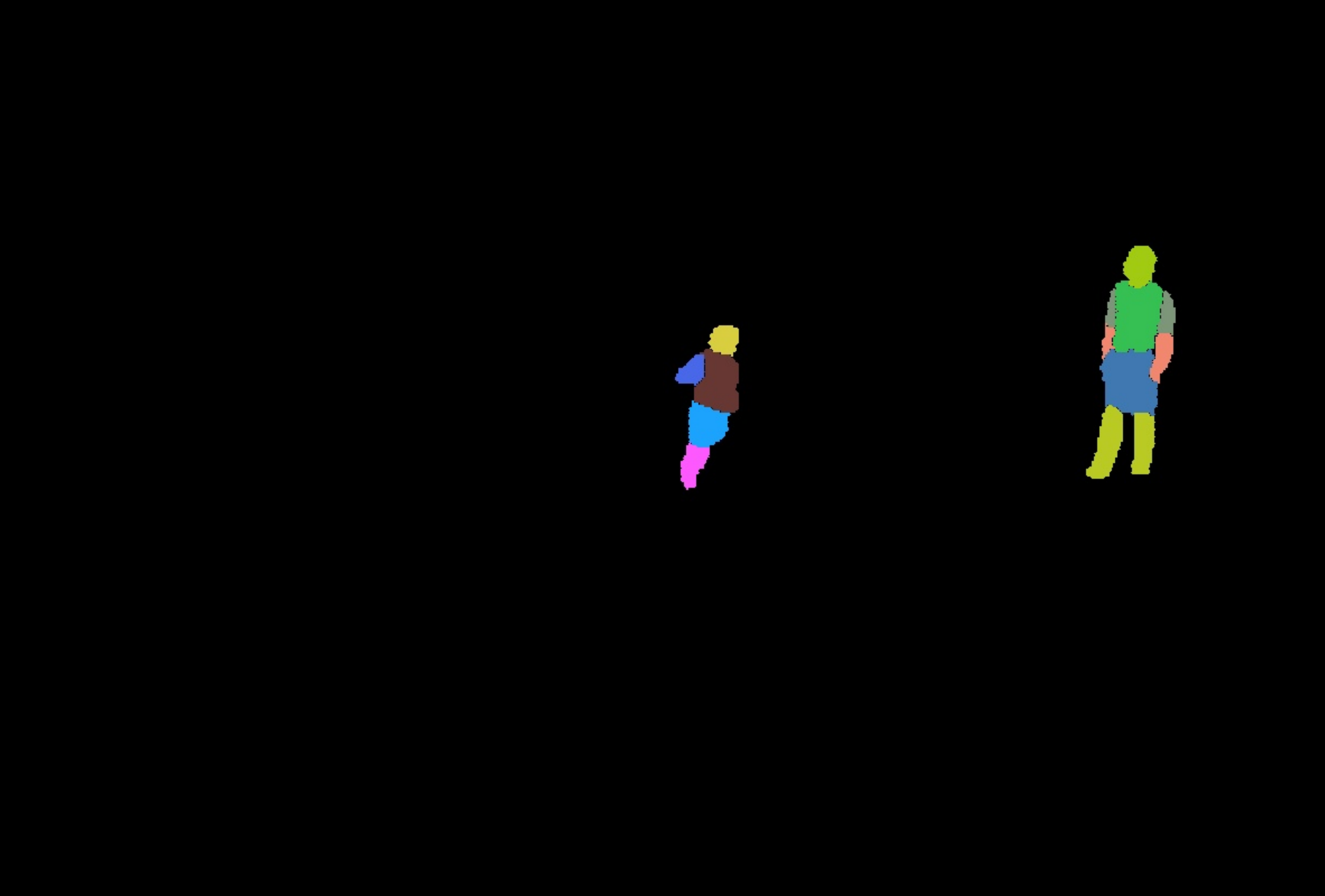} \\
			\includegraphics[height=1.5cm,width=2.25cm]{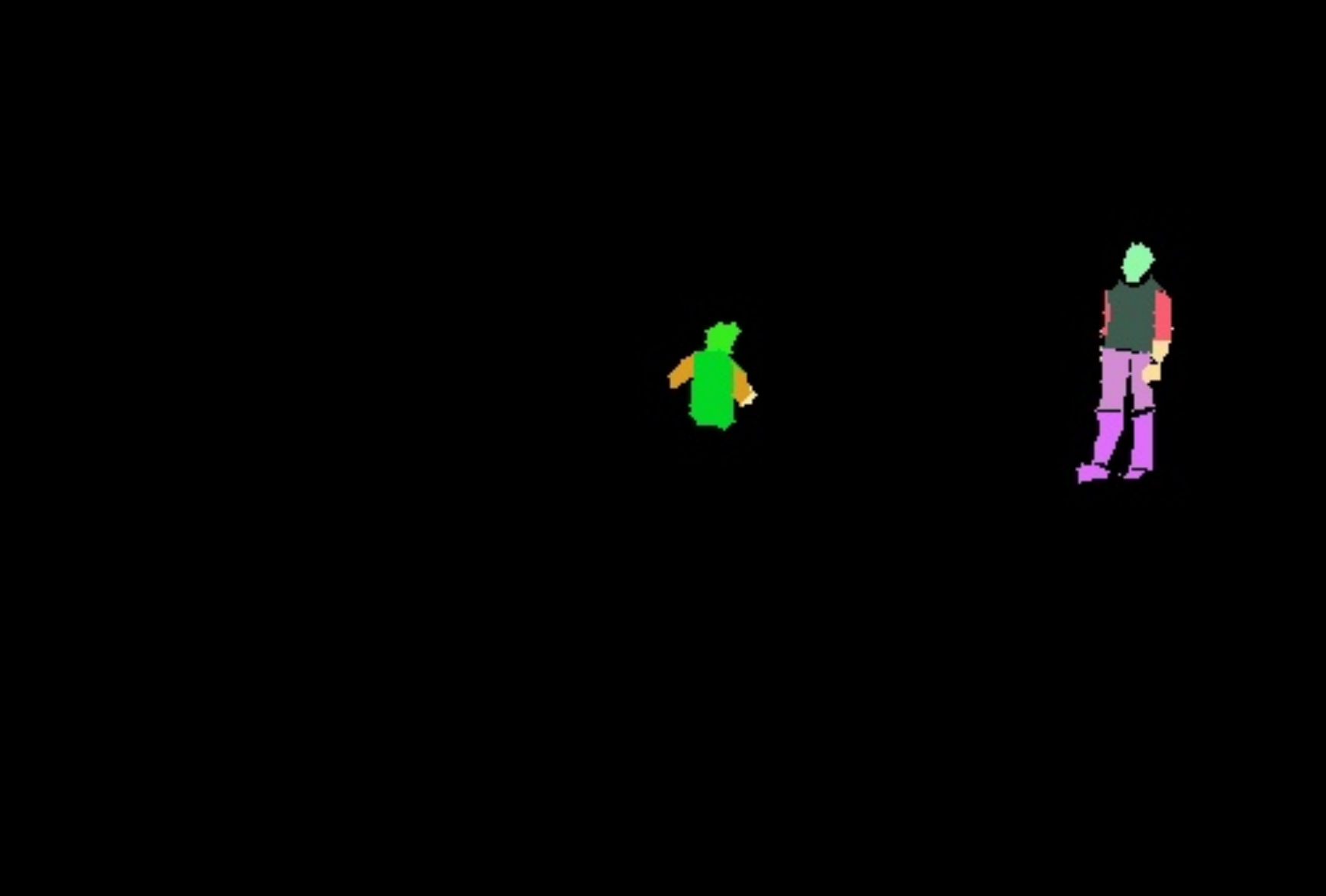} \\
			\includegraphics[height=1.5cm,width=2.25cm]{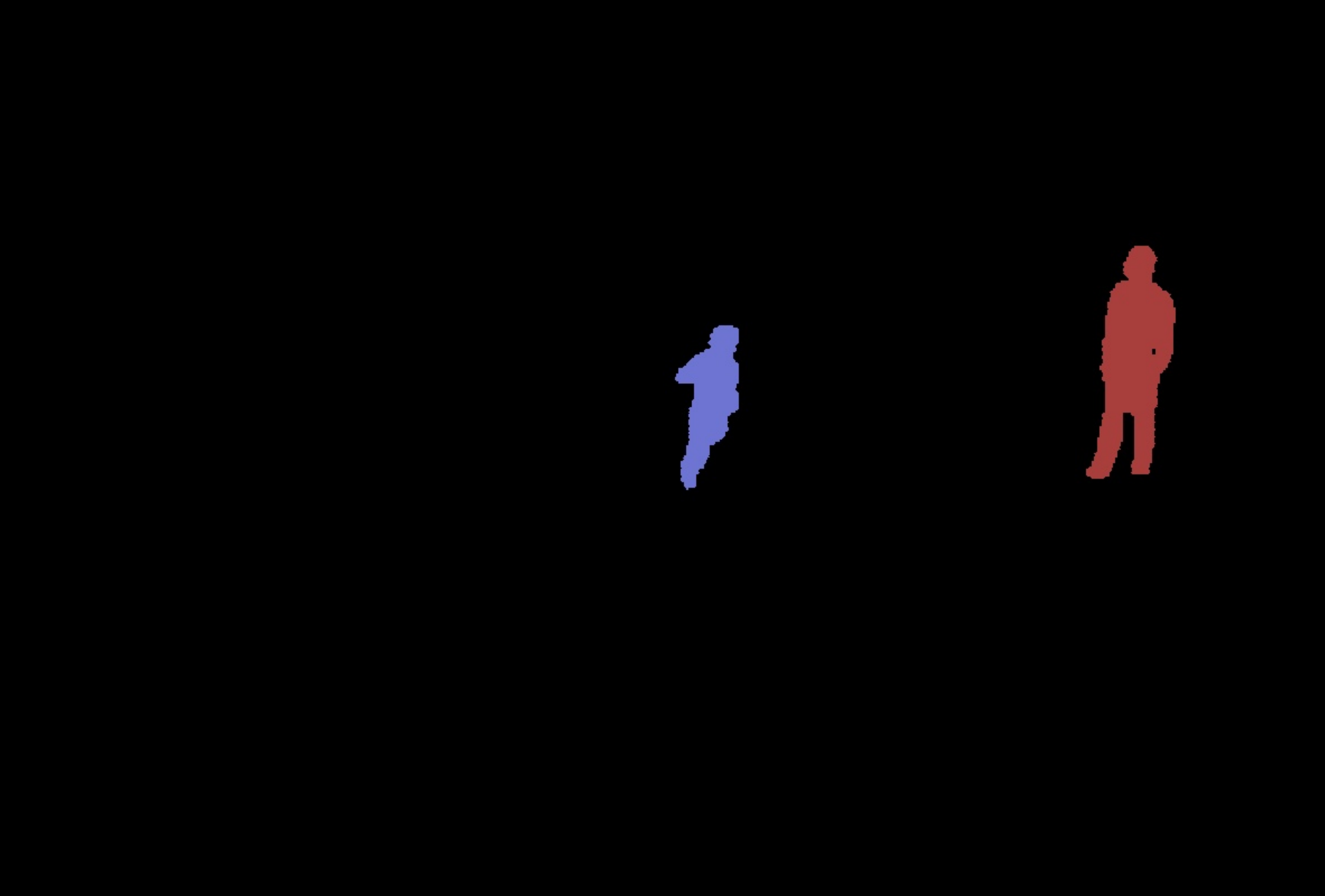} \\
			\includegraphics[height=1.5cm,width=2.25cm]{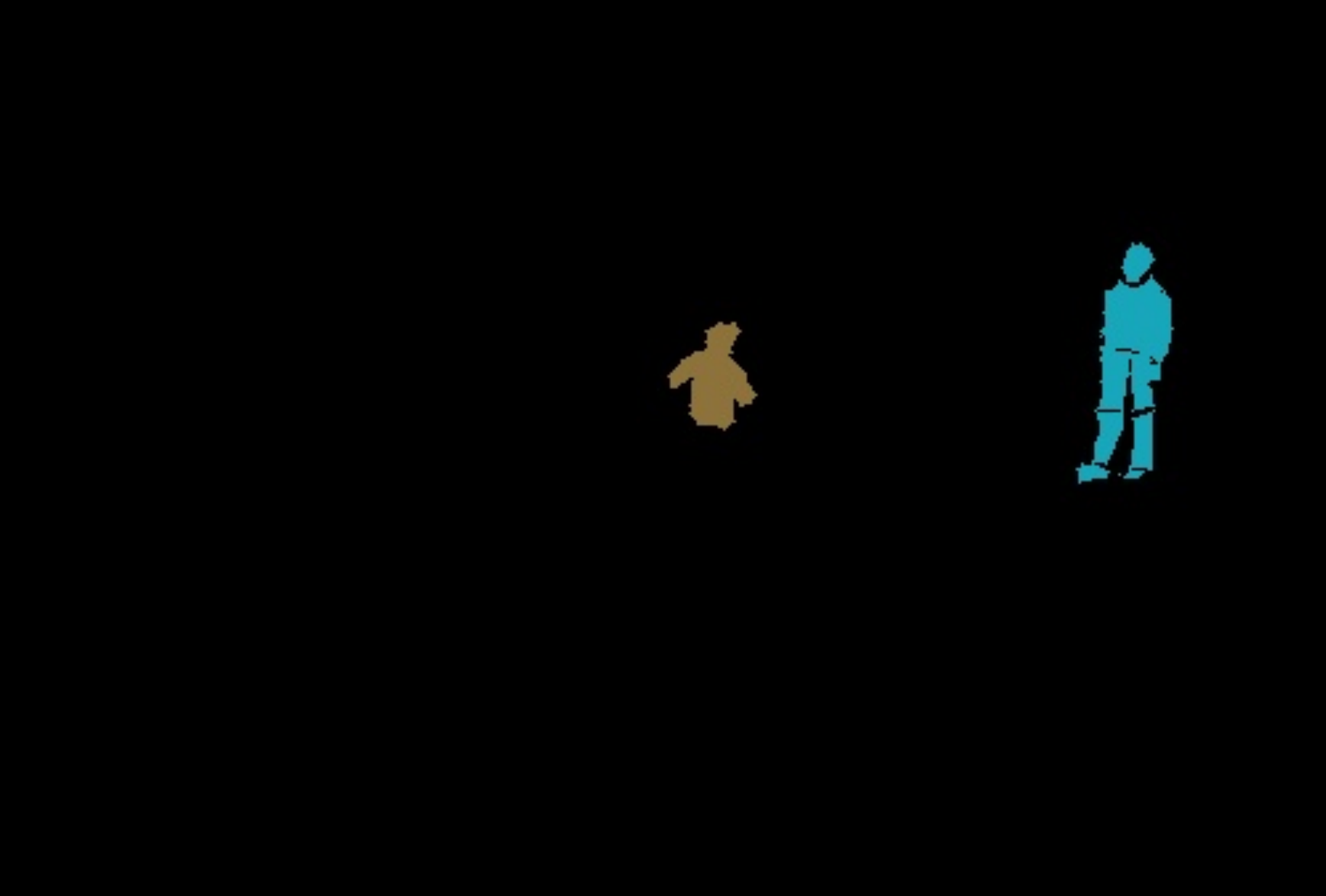} \\
		\end{minipage}
	}
	\subfigure
	{
		\begin{minipage}[b]{0.07143\linewidth}
			\centering
			\includegraphics[height=1.5cm,width=1cm]{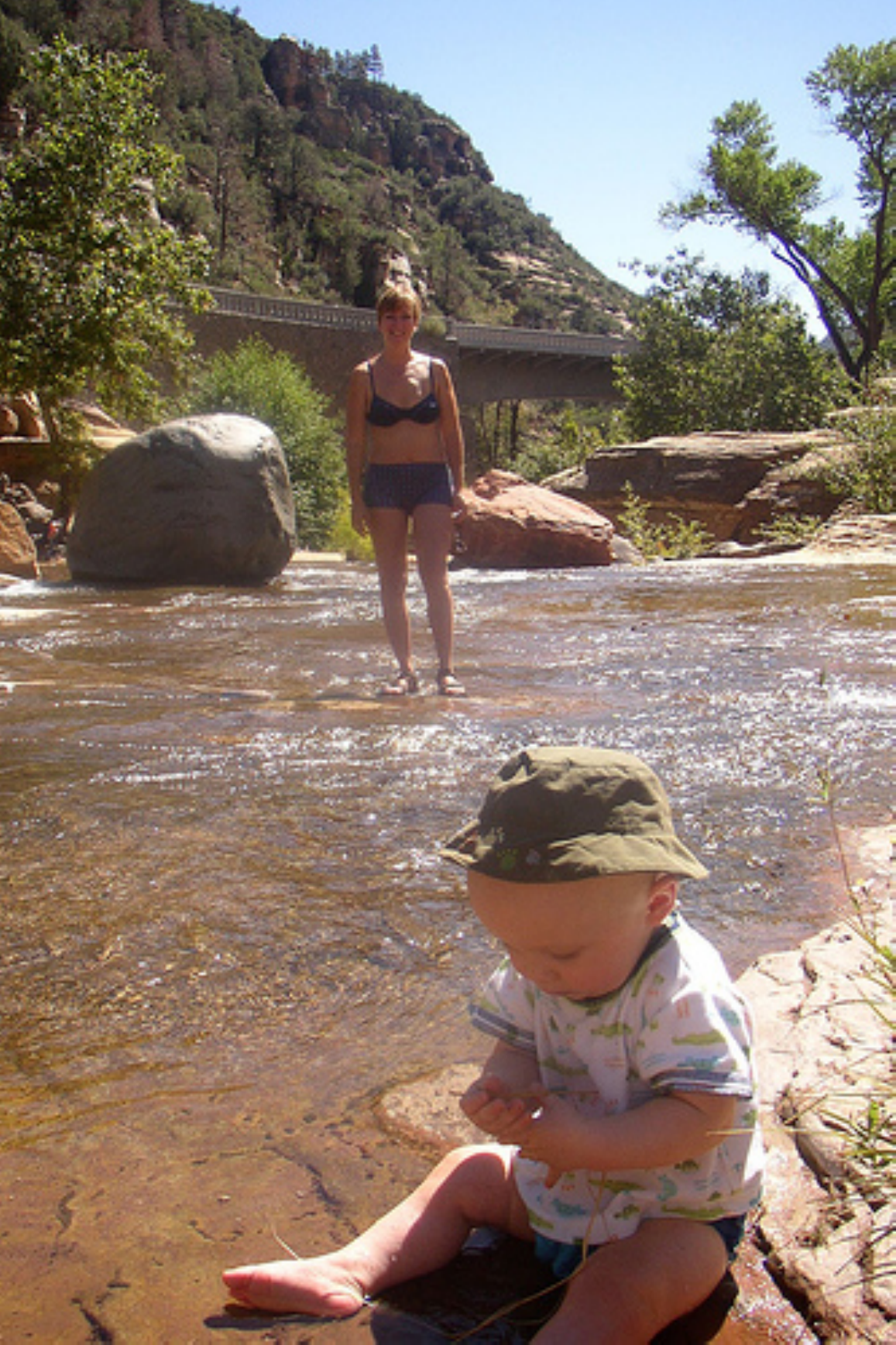} \\
			\includegraphics[height=1.5cm,width=1cm]{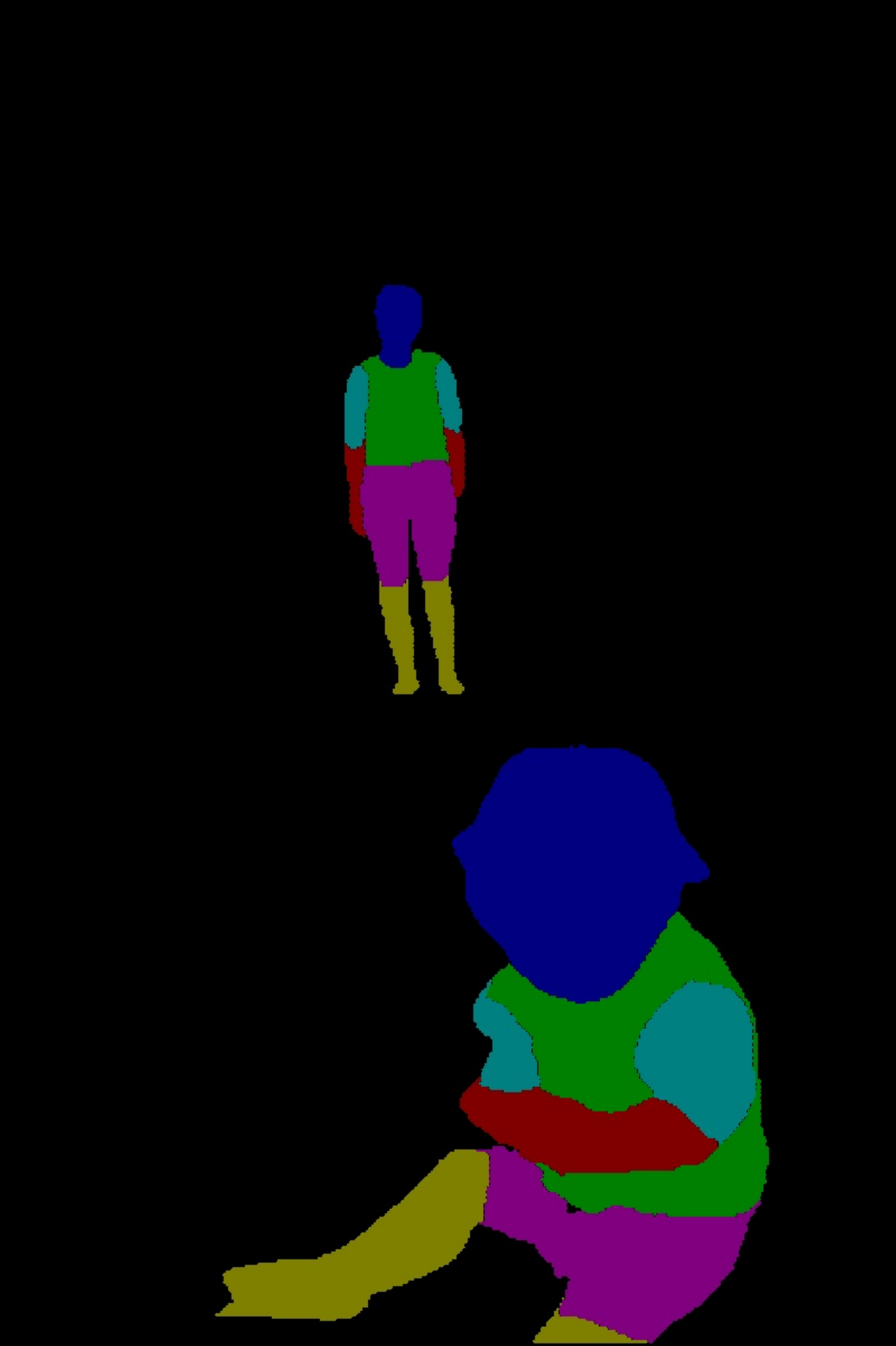} \\
			\includegraphics[height=1.5cm,width=1cm]{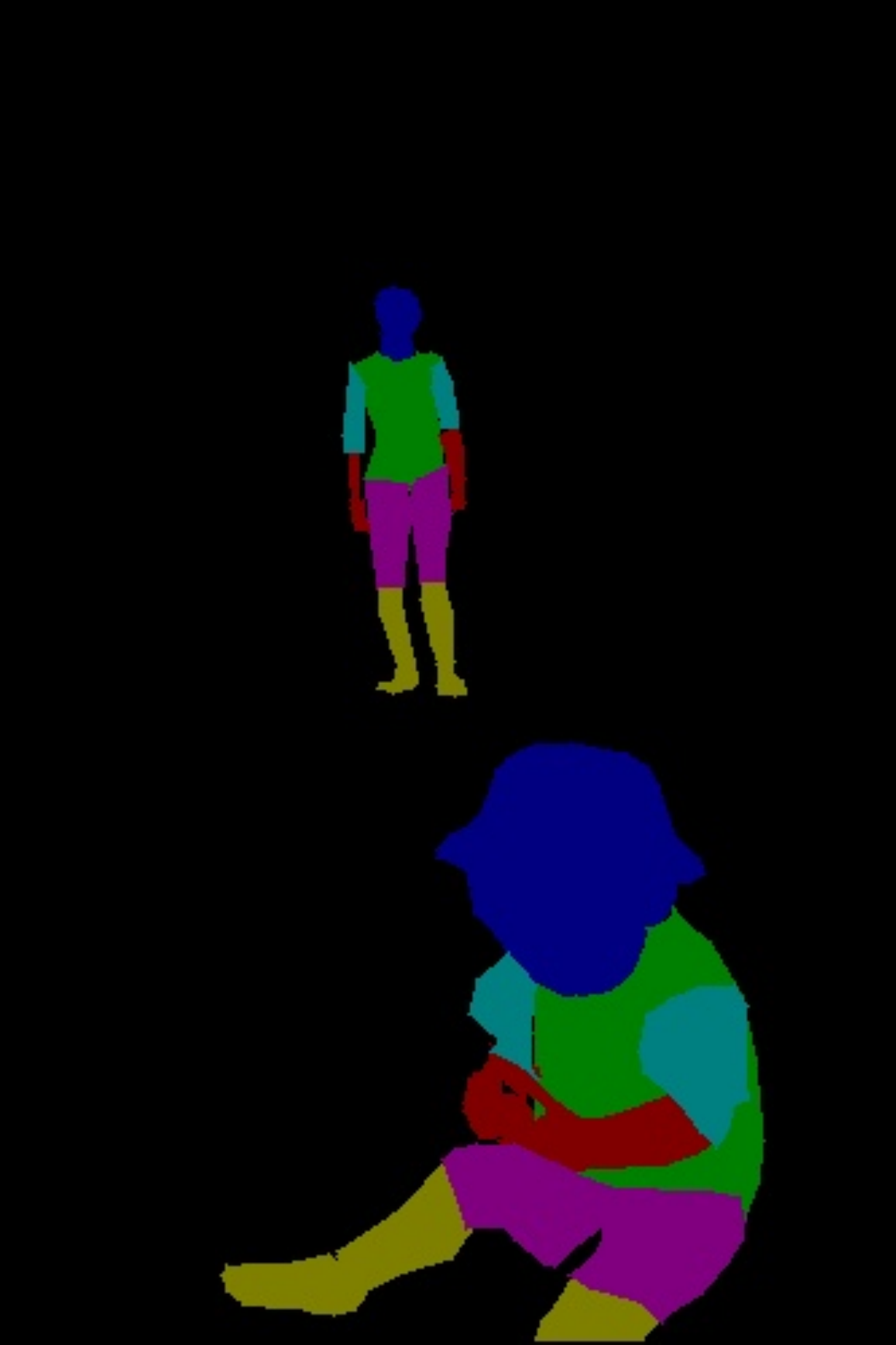} \\
			\includegraphics[height=1.5cm,width=1cm]{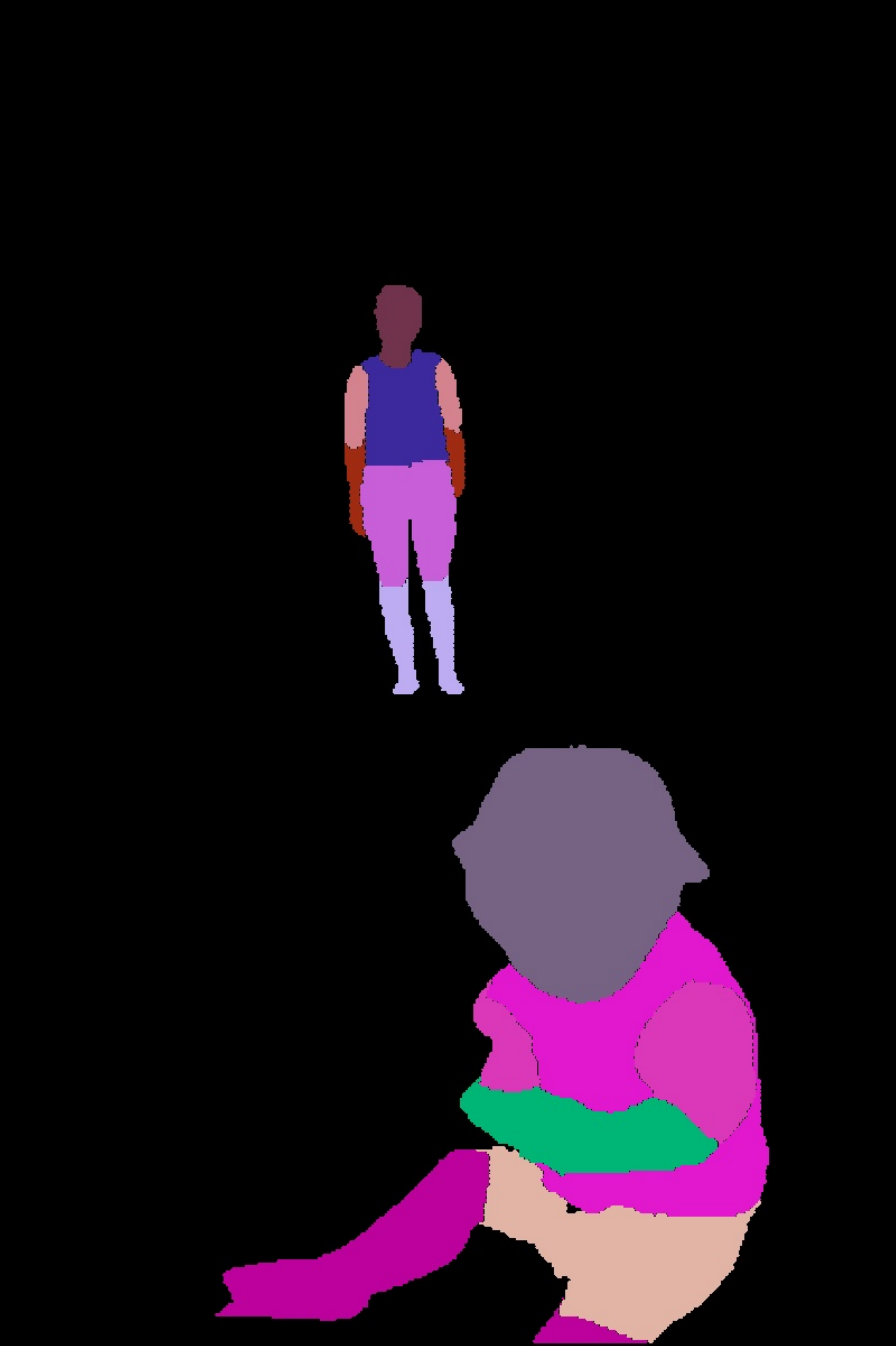} \\
			\includegraphics[height=1.5cm,width=1cm]{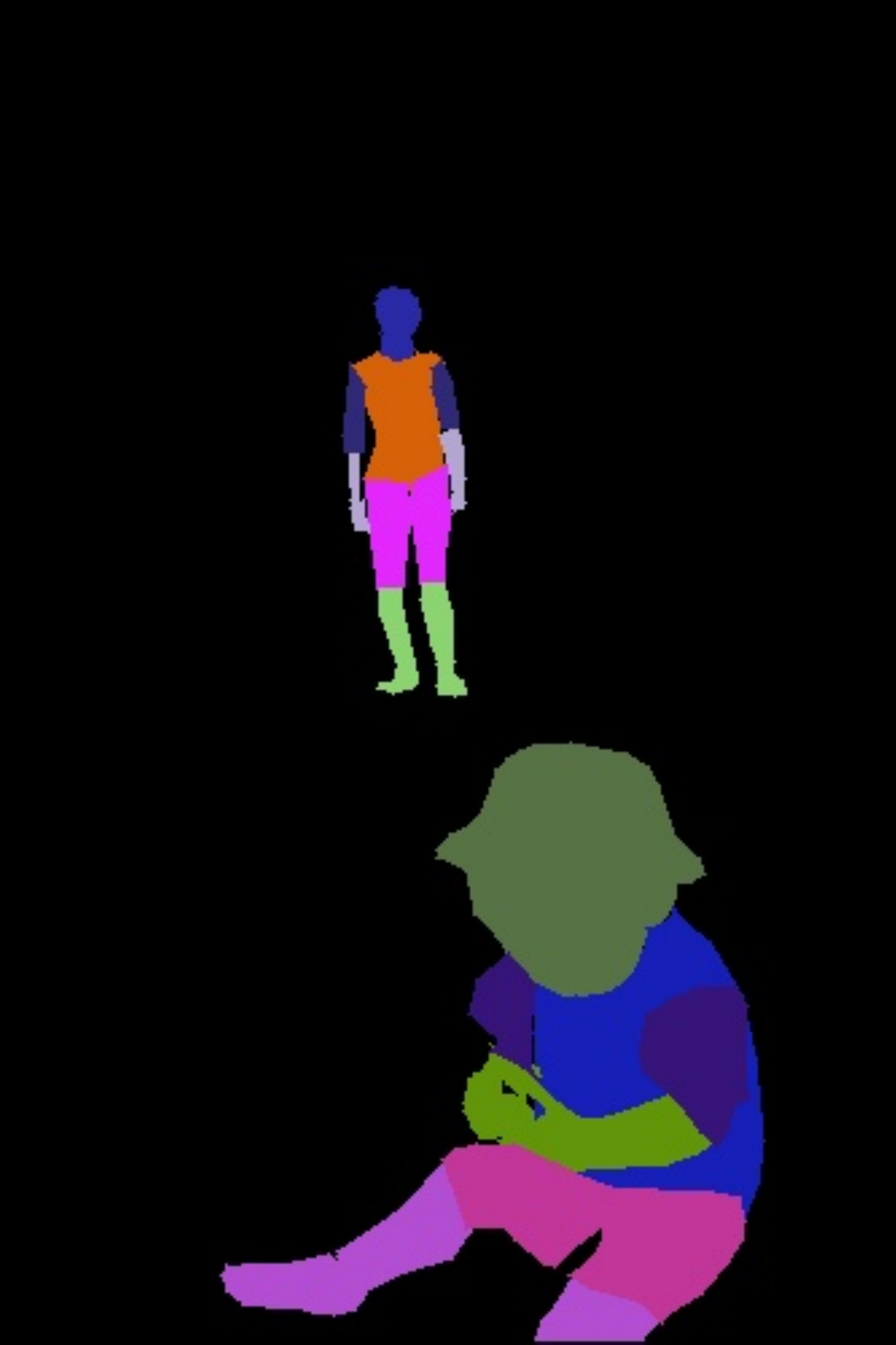} \\
			\includegraphics[height=1.5cm,width=1cm]{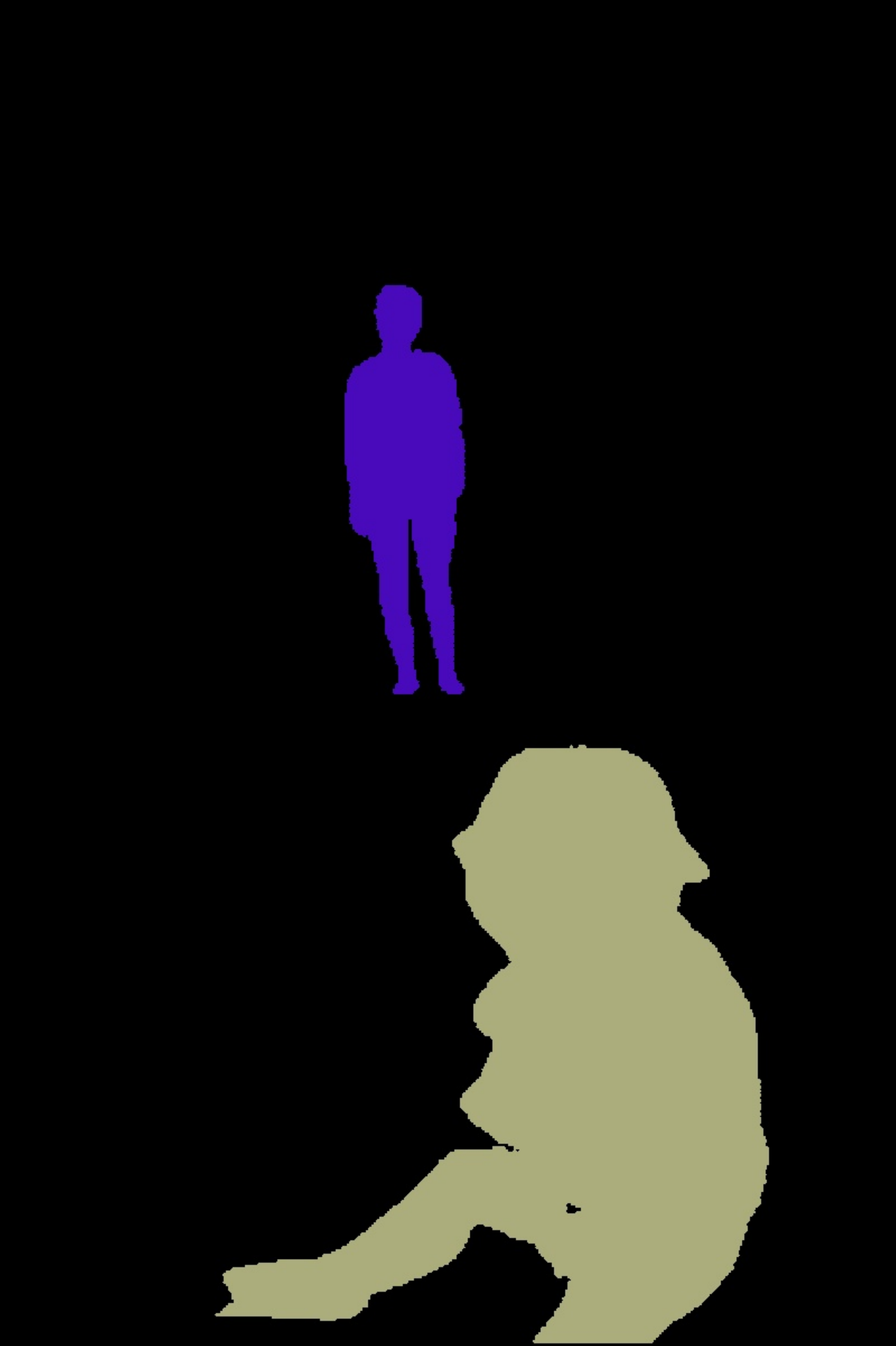} \\
			\includegraphics[height=1.5cm,width=1cm]{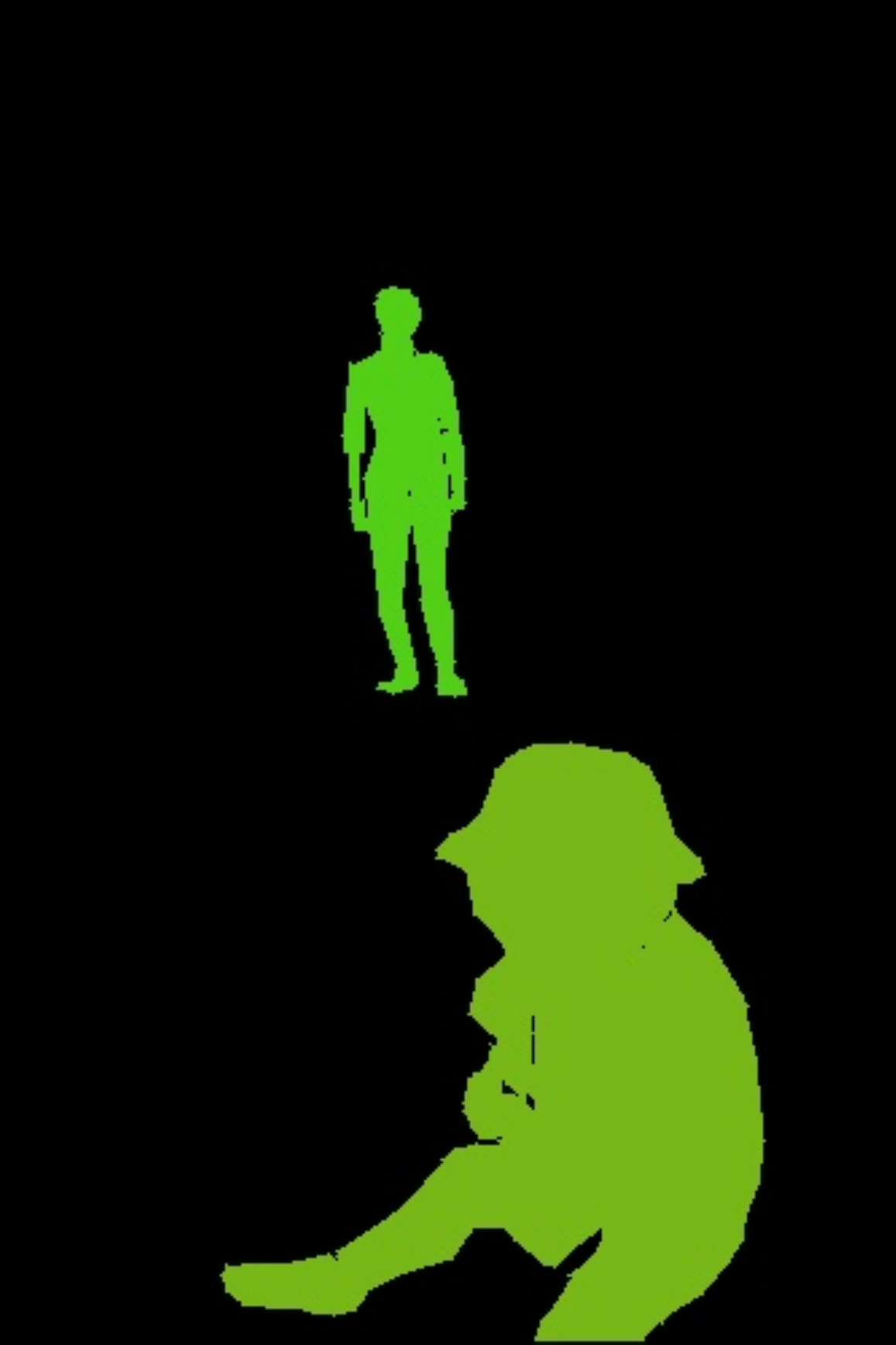} \\
		\end{minipage}
	}
	\caption{Visualization of NTHP on the PASCAL-Person-Part dataset. Pictures on the first line are original images, pictures on the second, fourth, and sixth lines are predictions, and those on the third, fifth, and last lines are ground truths.}
	\label{figure7}
\end{figure}

We show the results on the PASCAL-Person-Part dataset \cite{Chen2014} in Table \ref{tab:table7}. On this dataset, we use the metric $AP^r$ for a fair comparison. All previous methods use multi-scale and flip training. Besides, MNC \cite{Dai2016} was pre-trained on the Pascal VOC 2011/SBD dataset \cite{Hariharan2011}, and Holistic \cite{Li2017a} and PGN \cite{Gong2018} was pre-trained on the Pascal VOC dataset \cite{Everingham2010}. PGN \cite{Gong2018} further uses test-time augmentation (multi-scale and flip strategy). Unified \cite{Qin2020} employs the same setting as PGN \cite{Gong2018}. Unlike our experiments on the MHP v2.0 dataset \cite{Zhao2018}, we add flip operation in training due to the limited number of available images. From Table 7, we can see that with a shallower backbone ResNet50 and less training epochs and without any pre-training or test-time augmentation operation, our network outperform all state-of-the-art methods with 43.9 $AP^r_{vol}$ and 49.1, 40.0, 28.1 $AP^r$ with IoU thresholds of 0.5, 0.6, 0.7. We also show our result with pretraining on the MHP v2.0 dataset \cite{Zhao2018}. We obtain 47.1 $AP^r_{vol}$ and 53.9, 44.7, 31.9 $AP^r$ with IoU thresholds of 0.5, 0.6, 0.7, separately with a margin of 4, 5.8, 6.4, 6.2 points over the best previous entry. We also visualize good or bad results in Figure \ref{figure7}.

\section{Conclusions}

In this work, we design a straightforward and simple framework for multi-human parsing. Divergent from previous methods, we simultaneously conduct two types of instance segmentation for humans and parts using the same structure based on our newly proposed notion of ``indiscriminate objects with categories”. We also design a unified mask prediction module named UMPP which first generates a collection of prototypes shared among humans and parts, then makes a linear combination of them to get the binary masks. Besides, a simple post-processing strategy is developed to get the final results in a mutually beneficial way. We conduct extensive experiments on the challenging MHP v2.0 and PASCAL-Person-Part datasets and outperform all state-of-the-art methods.

\section*{Acknowledgements}
This work is supported by National Natural Science Foundation (NNSF) of China under Grant 62073237.
\bibliography{sci}

\end{document}